%% file: realease.tex
\documentclass[10pt,journal,compsoc]{IEEEtran}

\PassOptionsToPackage{numbers, compress}{natbib}

\usepackage[table,dvipsnames]{xcolor} 

\usepackage{amsmath,amsfonts,amssymb} 
\usepackage{nicefrac}                 
\usepackage{textcomp}                 
\usepackage{pifont}                   
\usepackage{bbding}                   
\usepackage{circledsteps}             

\usepackage{microtype}                
\usepackage{setspace}                 
\usepackage{enumitem}                 
\usepackage[normalem]{ulem}           
\useunder{\uline}{\ul}{}
\usepackage{graphicx}                 
\usepackage{subcaption}               
\usepackage{wrapfig}                  
\usepackage{stfloats}                 
\usepackage{tcolorbox}                

\usepackage{array}                    
\usepackage{booktabs}                 
\usepackage{multirow}                 
\usepackage{multicol}                 
\usepackage{makecell}                 
\usepackage{tabularx}                 

\usepackage{algorithm}                
\usepackage{algorithmicx}             
\usepackage{algpseudocode}            
\usepackage{verbatim}                 

\usepackage{natbib}                   

\usepackage{url}                      
\usepackage{hyperref}                 
\usepackage[accsupp]{axessibility}    
\usepackage{orcidlink}                

\usepackage{arydshln}                 

\usepackage{array} 
\newcolumntype{x}[1]{>{\centering\arraybackslash}p{#1pt}}

\newlength\savewidth

\definecolor{my_green}{RGB}{51,102,0}
\definecolor{my_red}{RGB}{204, 0, 0}

\usepackage{xspace}
\makeatletter
\DeclareRobustCommand\onedot{\futurelet\@let@token\@onedot}
\def\@onedot{\ifx\@let@token.\else.\null\fi\xspace}

\def\eg{\emph{e.g}\onedot} 
 
\def\cf{\emph{cf}\onedot} 
\def\etc{\emph{etc}\onedot}

\makeatother

\definecolor{myred}{RGB}{200,50,50}

\newcommand{{\MethodName}}{\bf{BEE}}

\definecolor{mycustomcolor}{HTML}{DDEEFF} 

\definecolor{crosscolor}{rgb}{0.969,0.580,0.114} %
\definecolor{checkcolor}{rgb}{0.485,0.640,0.204} %
\definecolor{brickred}{RGB}{203,65,84} %

\begin{document}



\title{Beyond the Eye: Efficient Multimodal Reasoning via Self-Regulated Implicit Visual Tools}

\author{
Xiuwei Chen,
Quanlin Chen,
Wentao Hu,
Zisheng Chen,
Kun Xiang,
Zehua Ma,
\\
Mingyang Zhang,
Jianhua Han, 
Hanhui Li,
Hang Xu, 
Xiaodan Liang\textsuperscript{\textdagger}
\IEEEcompsocitemizethanks{
\IEEEcompsocthanksitem  Corresponding author\textsuperscript{\textdagger}: Xiaodan Liang.
\IEEEcompsocthanksitem Xiuwei Chen, Quanlin Chen, Zisheng Chen, Kun Xiang, Hanhui Li, Zehua Ma, Mingyang Zhang, Xiaodan Liang are with Sun Yat-sen University, China. E-mail: \{chenxw83,chenzsh9,xiangk,mazh58\}@mail2.sysu.edu.cn, \{lihh77,liangxd9\}@mail.sysu.edu.cn.
\IEEEcompsocthanksitem Wentao Hu is with The Hong Kong Polytechnic University, China. E-mail: wayne-wt.hu@connect.polyu.hk.
\IEEEcompsocthanksitem Jianhua Han, Hang Xu are with Yinwang Intelligent Technology Co., Ltd., China. E-mail: \{hanjianhua2012,chromexbjxh\}@gmail.com.
}
}




\maketitle

\input{sec/0_abstract}
\input{sec/1_intro}
\input{sec/2_related}
\input{sec/3_method}
\input{sec/4_exp}
\input{sec/5_conclusions}

\bibliographystyle{IEEEtran}
\bibliography{refs-clean}

\input{sec/6_appendix}


\end{document}

%% file: sec/0_abstract.tex
\begin{abstract}
Recent multimodal large language models (MLLMs) have made remarkable progress on fine-grained perception tasks under the "Thinking with Images" (TwI) paradigm by iteratively performing various visual tool operations. However, this paradigm relies heavily on frequent external tool calls and repeated image re-encoding, which leads to substantial computational overhead and inference latency. Furthermore, existing models generally lack capability boundary awareness, making it difficult to adaptively determine when to rely on internal parametric knowledge and when to invoke external tools.
To address these issues, we propose \textbf{Beyond the Eye (BEE)}, a novel implicit visual tool paradigm centered on self-regulated capability. BEE directly incorporates visual tool invocation behaviors into the training objective and encourages the model to develop a self-regulated invocation mechanism. This design enables the model to adaptively balance internal knowledge and implicit tools, avoiding redundant tool usage while substantially reducing inference latency.
Specifically, BEE involves a two-stage training process: (1) Formalized Chain-of-Thought (CoT) Supervised Fine-tuning (SFT). We construct CoT trajectories with structured tool slots and mixed invocation states. This stage activates the model's implicit tool representations and adaptive switching capability. (2) Self-regulated Reward-Driven Alignment. To address redundant tool usage caused by ambiguous cognitive boundaries, we first introduce the Net Tool Gain (NTG) metric to quantify this phenomenon. Based on this observation, we further propose a self-regulated reward mechanism. This mechanism penalizes ineffective tool dependency and encourages the model to perform knowledge routing, ensuring that implicit tools are invoked only when the model's internal knowledge is insufficient.
BEE achieves state-of-the-art performance in fine-grained visual perception while remaining competitive in general reasoning tasks and achieving substantial gains in inference efficiency.
In particular, under the high-resolution setting of 8192, BEE achieves up to $\sim 86.2\%$ inference acceleration compared with the DeepEyes method.
BEE provides a promising direction for developing implicit tool MLLMs that combine strong cognitive ability with extremely low latency and further advances the development of self-regulated MLLMs.
The code is publicly available at \url{https://self-improvement-tool.github.io/bee.github.io/}.

\end{abstract}

%% file: sec/1_intro.tex
\section{Introduction} \label{sec:intro}
\input{table_figs/fig0_first}
\IEEEPARstart{M}ultimodal Large Language Models (MLLMs)~\cite{bai2025qwen2,bai2025qwen3,zhu2023minigpt,hurst2024gpt,liu2023visual,Li2023BLIP2BL,Dai2023InstructBLIPTG,comanici2025gemini} integrate multimodal information into language intelligence and have attracted considerable attention for their ability to understand and generate multimodal content. In recent years, extending the Chain-of-Thought (CoT) paradigm~\cite{Li2025FromS1,Wang2025MultimodalCR,Zhang2024ImproveVL,Dong2024InsightVEL,Zhang2023MultimodalCR,zhou2025r1zerosahamomentvisual,R1VL,R1Onevision} to multimodal settings enables models to decompose complex problems into multiple reasoning steps, leading to substantial improvements in task performance. However, the conventional ``Think about Images'' paradigm~\cite{bai2025qwen2,bai2025qwen3,liu2023visual} typically relies on a one-time visual encoding, after which reasoning is carried out entirely in the textual space. Due to the substantial gap between the continuous visual world and discrete symbolic semantics, this one-shot feature extraction strategy struggles to capture dynamic visual details, leading to performance bottlenecks on complex visual tasks that require deep and iterative reasoning.

This limitation drives the transition from reasoning about images to a new paradigm, namely ``Think with Images'' (TwI)~\cite{Fan2025GRITTM,feng2025retool, zhang2025thyme, lai2025mini, zheng2025deepeyes,zhang2025chain,hu2024visual}, where models reason by directly interacting with visual representations. In this paradigm, visual information is no longer treated as a static input but becomes a dynamic cognitive workspace that can be actively manipulated. Models can query, transform, and even generate new visual representations, which serve as important intermediate steps during reasoning and enable a deeper understanding of visual cues.  For instance, Thyme~\cite{zhang2025thyme} and DeepEyes~\cite{zheng2025deepeyes} generate executable code to invoke external visual tools, such as local zooming, and then re-encode the generated images into the reasoning process, thereby supporting fine-grained perception.

Nevertheless, this heavy reliance on external tools inevitably introduces frequent {I/O} operations and repeated image re-encoding, resulting in considerable computational latency. To alleviate this issue, recent latent reasoning approaches~\cite{Chen2025ReasoningBL,Yu2026TheLS}, such as Monet~\cite{wang2026monet} and EVA~\cite{chen2026latent}, attempt to replace explicit tool invocation with continuous implicit visual representations, significantly improving inference efficiency. Unfortunately, both conventional TwI methods and emerging latent reasoning models generally lack capability boundary awareness. They are unable to adaptively estimate task difficulty or dynamically determine whether to rely on internal parametric knowledge or invoke external tools and implicit visual reasoning. As a result, redundant tool usage and unnecessary computational overhead remain unavoidable.


To address these limitations, we propose \textbf{Beyond the Eye (BEE)}, a novel implicit visual tool (IvT) paradigm centered on self-regulated capability. Rather than calling external tools, BEE folds visual tool invocation directly into the training objective and learns an explicit capability boundary, so that it adaptively balances internal knowledge against implicit tool use, avoiding redundant invocations and reducing latency.
BEE consists of two training stages: 1) \textit{formalized CoT supervised fine-tuning} and 2) \textit{self-regulated reward-driven alignment}. The Stage 1 training of \textit{formalized CoT supervised fine-tuning} aims to activate the model's implicit tool representations, adaptive switching mechanism, and basic reasoning ability. During this stage, the model predicts both the type and output of tool behaviors required for reasoning and dynamically generates them in the form of CoT trajectories equipped with implicit tool slots. To suppress redundant tool usage and improve inference efficiency, we adopt carefully designed formalized CoT labels as supervision signals. Specifically, the label set contains a mixture of reasoning trajectories with and without implicit tool slots, which explicitly guides the model to learn when to invoke tools and when to rely on internal parametric knowledge.

However, supervised fine-tuning alone is often insufficient for establishing a clear cognitive boundary. To quantitatively evaluate this issue, we introduce the Net Tool Gain (\textbf{NTG}) metric, which measures the actual marginal contribution of tool invocation to task success after excluding the intrinsic capability of the base model. The NTG analysis reveals an important observation: although the model optimized via the first training stage and a simple DAPO~\cite{Yu2025DAPOAO} algorithm possesses the capability of tool usage, it still suffers from substantial redundant tool invocation, as illustrated in Figure~\ref{fig:fig3_motivation}.

To fundamentally overcome this limitation, we introduce the Second training stage, namely \textit{self-regulated reward-driven alignment}. This stage aims to mitigate the model's overfitting to pre-constructed CoT labels and 
further improve its self-regulated capabilities, thereby enhancing complex reasoning and generalization.
Based on this observation, we propose a \textbf{self-regulated} reward mechanism. By penalizing ineffective tool dependency, this mechanism encourages the model to perform knowledge routing and ensures that implicit tools are invoked only when the model's internal knowledge is insufficient.
Compared with conventional supervised fine-tuning methods that rely on fixed CoT labels, BEE adaptively switches tools during CoT generation and establishes a dynamic implicit tool switching mechanism. Moreover, driven by self-regulated rewards, BEE can assess its own cognitive boundaries in a self-regulated manner and actively suppress redundant tool usage whenever internal parametric knowledge is sufficient to solve the problem.

We conduct extensive experiments on multiple public perception tasks, including HRBench~\cite{wang2025divide} and MME-RealWorld~\cite{zhang2024mme}, \etc. 
Experimental results show that BEE consistently outperforms existing explicit tool-based methods while substantially reducing inference latency, achieving up to \textbf{86.2\%} latency reduction under the 8K resolution setting (\eg, Figure~\ref{fig:fig0_first} (right)). Notably, BEE-8B and BEE-4B even \textbf{surpass} the commercial model Kimi-K2.6~\cite{team2026kimi} in average accuracy on perception tasks (\eg, Figure~\ref{fig:fig0_first} (left)). These results demonstrate the effectiveness of the self-regulation-driven visual tool internalization paradigm in improving both reasoning accuracy and inference efficiency of MLLMs.
We further conduct extensive ablation studies and in-depth analyses to investigate different implicit tool representations, the effect of decoupling perception and reasoning, and deployment feasibility in real-world scenarios, \etc. These analyses not only validate the effectiveness of each core component but also provide insights into how the BEE paradigm improves model performance while preserving reasoning interpretability.

To summarize, the main contributions of this paper are:

\begin{itemize}
    \item We propose \textbf{BEE}, a novel self-regulation-driven implicit visual tool (IvT) paradigm. BEE aims to avoid redundant tool usage and substantially reduce inference latency, enabling adaptive and generalizable reasoning across diverse tasks and complex scenarios.
    \item We construct formalized CoT labels for supervised fine-tuning to effectively activate the model's implicit tool representations, adaptive switching mechanism, and fundamental multimodal reasoning ability. In addition, we introduce the Net Tool Gain (\textbf{NTG}) metric to quantify the actual marginal contribution of tool invocation to task success.
    \item We propose a \textbf{Self-regulated} reward mechanism. This mechanism enables the model to optimize its reasoning path in a self-regulated manner and fundamentally suppress redundant tool usage.
    \item Experimental results demonstrate that BEE outperforms existing tool-based methods on multiple perception benchmarks. The proposed method simultaneously improves reasoning ability and reduces inference latency, providing valuable insights for designing efficient and self-regulated visual reasoning paradigms.
\end{itemize}

%% file: table_figs/fig0_first.tex
\begin{figure*}[ht]
    \centering
    \includegraphics[width=0.95\linewidth]{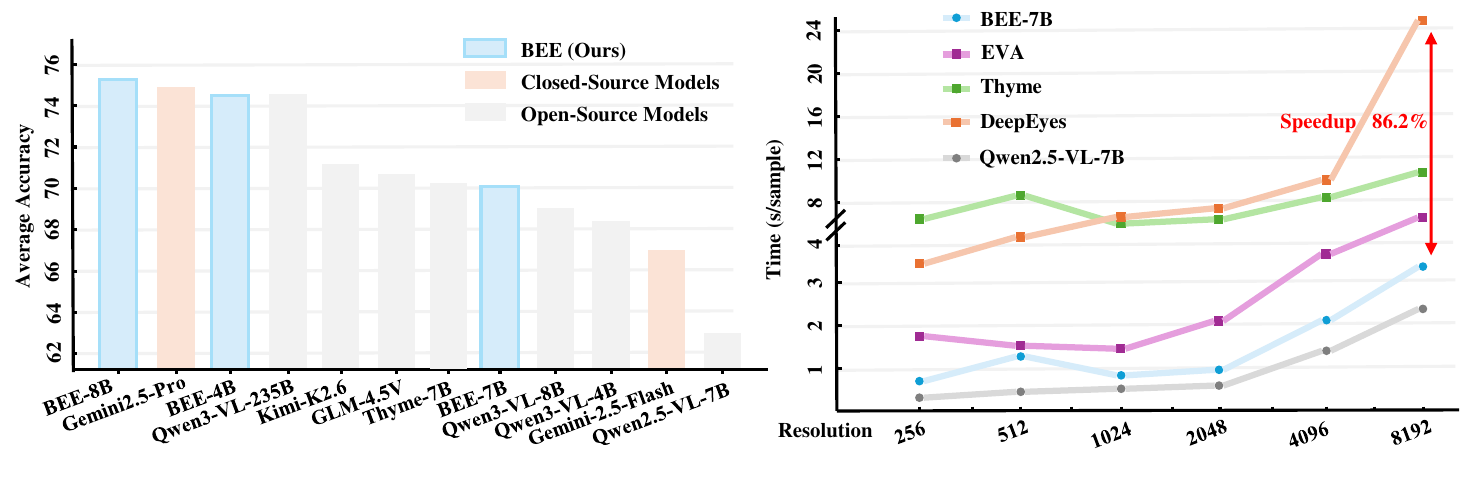}
    \caption{
        (\textbf{Left}) \textbf{Average scores across multimodal perception benchmarks}, including V* Bench, HRBench, MME-RealWorld, MME-RealWorld-Lite. BEE demonstrates superior performance compared with much larger open-source models (\eg, Qwen3-VL-235B) and closed-source models (\eg, Gemini-2.5-Pro).
        (\textbf{Right}) \textbf{Inference time comparison (s/sample) across different resolutions}. BEE exhibits higher inference efficiency than TwI and latent methods (\eg, DeepEyes, Thyme, and EVA) under all resolution settings.
    }
    \label{fig:fig0_first}
\end{figure*}

%% file: sec/2_related.tex
\section{Related Work}\label{relatedwork}

\subsection{Thinking about Images}
Multimodal Large Language Models (MLLMs)~\cite{bai2025qwen2,bai2025qwen3,zhu2023minigpt,hurst2024gpt,liu2023visual,Li2023BLIP2BL,Dai2023InstructBLIPTG,comanici2025gemini,jaech2024openai,Chen2023InternVS,Li2024LLaVAOneVisionEV,Zhang2024MMLLMsRA,Lin2024MoELLaVAMO,Jin2024EfficientML} have witnessed rapid advancements in recent years, leading to significant breakthroughs in visual understanding and cross-modal reasoning within the field of artificial intelligence.
The introduction of OpenAI o1~\cite{jaech2024openai} establishes the modern paradigm of inference-time scaling, where native long-chain reasoning substantially enhances the reasoning capabilities of MLLMs.
Recent advances in multimodal reasoning mainly follow two directions: supervised fine-tuning (SFT) and reinforcement learning (RL). Early SFT-based methods~\cite{bai2025qwen2,bai2025qwen3,liu2023visual,LLaVACoT,AtomThink,LLaMABerry,Zhang2023MultimodalCR,Qiao2024WeMathDY,Peng2025LMMR1E3} construct large-scale reasoning data to teach explicit chain-of-thought processes. For example, LLaVA-CoT~\cite{LLaVACoT} and AtomThink~\cite{AtomThink} introduces multimodal reasoning annotations, while LLaMA-Berry~\cite{LLaMABerry} further leverages Collective Monte Carlo Tree Search (CoMCTS) to generate higher-quality reasoning trajectories. 
Inspired by the success of DeepSeek-R1~\cite{guo2025deepseek}, RL has recently emerged as a powerful paradigm~\cite{zhou2025r1zerosahamomentvisual,R1VL,R1Onevision,Yue2025DoesRL,Liu2025VisualRFTVR,Hu2025OpenReasonerZeroAO,Wang2025VLRethinkerIS,Feng2025GroupinGroupPO,Zuo2025TTRLTR} for multimodal reasoning. 
Building on this paradigm, open-source reinforcement learning methods such as MMR1~\cite{MMR1}, Vision-R1~\cite{VisionR1}, R1-VL~\cite{R1VL}, and MM-EUREKA~\cite{meng2025mm} further improve self-verification through rule-based reward mechanisms. 
Vision-R1~\cite{VisionR1} adopts GRPO together with hard format rewards and cold-start initialization data to enhance emergent reasoning. 
R1-VL~\cite{R1VL} further introduces Step-wise Group Relative Policy Optimization (StepGRPO), an online reinforcement learning framework that uses dense step-wise rewards to improve reasoning performance. 
MM-EUREKA~\cite{meng2025mm} shows that rule-based reinforcement learning can induce emergent reasoning behaviors without supervised fine-tuning, leading to improved data efficiency. 
Combining the advantages of supervised fine-tuning and reinforcement learning, methods such as R1-OneVision~\cite{R1Onevision} and Mimo-VL~\cite{MiMoVL} adopt multi-stage post-training to iteratively refine policy models while preserving both strong reasoning ability and general-purpose performance.
Despite the rapid progress, existing methods often exhibit a strong preference for textual reasoning while underutilizing visual evidence, leading to spurious correlations and hallucinated reasoning chains. Moreover, this issue tends to worsen as reasoning trajectories become longer. To address this limitation, the {Thinking with Images} paradigm has recently been proposed to enhance visual grounding by enabling reasoning directly over visual representations.


\subsection{Thinking with Images}

Recent progress in tool-augmented large models~\cite{team2025kimi,geng2025webwatcher,mai2025agent,tao2025webshaper,xue2025simpletir,Hou2025CodeVCW,Zhou2025ReinforcedVP,Zhao2025PyVisionAV,Wu2025VToolR1VL,Chng2025SenseNovaMARSEM} has introduced a complementary direction for improving reasoning by enabling models to interact with external computational environments. Early studies mainly focused on integrating symbolic solvers, code execution engines, and structured verification modules into the reasoning process, enabling models to delegate algebraic manipulation, equation solving, and geometric computation to specialized tools. Sketchpad~\cite{hu2024visual} introduces a latent canvas for intermediate visual reasoning, while DeepEyes~\cite{zheng2025deepeyes} equips models with cropping operations to inspect local regions. Chain-of-Focus~\cite{zhang2025chain} formulates multimodal reasoning as an adaptive zooming process through dedicated control tokens, and Pixel-Reasoner~\cite{su2025pixel} further incorporates a curiosity-driven exploration strategy. Active-o3~\cite{zhu2025active} extends zooming to a broader active perception setting and trains the policy with GRPO, whereas MGPO~\cite{huang2025high} improves visual grounding through multi-round grounding optimization using only answer-level supervision.
Subsequent work has expanded both the scope of interactions and the diversity of tools. SyncLoop~\cite{Chen2025C2EvoCM} proposes a self-improving framework that progressively aligns task difficulty with model capability. Thyme~\cite{zhang2025thyme} introduces a Python-based sandbox for flexible tool use, while Mini-o3~\cite{lai2025mini} increases both the number and diversity of interactions. DeepEyes-v2~\cite{hong2025deepeyesv2} and R1V4~\cite{zhang2025skywork} further extend the available tools by incorporating external resources such as web search.
Despite these advances, existing approaches still rely heavily on external agents and discrete tool invocations, which introduce considerable inference overhead and limit reasoning efficiency.



\subsection{Implicit Visual Tools}
Conventional Chain-of-Thought (CoT) reasoning is typically expressed in human-interpretable language. In contrast, Coconut~\cite{hao2024training} introduces the concept of continuous chains of thought, replacing discrete linguistic forms with latent representations that are argued to contain richer reasoning information. LatentSeek~\cite{li2025seek} further extends this idea to test-time scaling by iteratively refining latent trajectories based on intermediate feedback.
Recent studies~\cite{chen2025mint,lvr,chen2026latent,wang2026monet,bigverdi2024perception,Qin2025ChainofVisualThoughtTV,Zhang2025LatentSS,Wei2025OpenVR,Gu2025ThinkMorphEP,Liu2025ReasoningWT,Dong2025InterleavedLV,Pham2025MultimodalCO,Chen2025ReasoningIT,Cao2025SpatialDreamerIS,Wang2026ForestBT,Ray2025MullTokensML,Zhang2025DeepSketcherIV,Li2025LatentIV,Li2026ImaginationHV,Jeon2026VisionalignedLR} have brought latent reasoning into multimodal settings. AURORA~\cite{bigverdi2024perception} introduces dedicated perception tokens to support deeper visual understanding, while others~\cite{mirage,Dong2025InterleavedLV,Qin2025ChainofVisualThoughtTV,Chen2025ReasoningIT,Gu2025ThinkMorphEP}, interleave textual and latent tokens to enable implicit multimodal reasoning. MINT-CoT~\cite{chen2025mint} further aligns interleaved visual tokens with intermediate reasoning steps, and LVR~\cite{lvr} explores reasoning over latent visual representations to reduce reliance on textual CoT.
Beyond architectural designs, several studies~\cite{Wang2026ForestBT,Li2026ImaginationHV,Jeon2026VisionalignedLR} investigate how to optimize latent reasoning. EVA~\cite{chen2026latent} proposes D-GSPO to separately optimize discrete textual outputs and continuous latent reasoning states, whereas Monet~\cite{wang2026monet} introduces VLPO to address the optimization challenges associated with continuous latent tokens.
CapImagine~\cite{Li2026ImaginationHV} further proposes a text-space imagination-based method.
Despite their promising efficiency and reasoning performance, existing latent reasoning methods generally rely on predefined reasoning budgets and lack awareness of their reasoning boundaries. Consequently, they cannot adaptively determine when additional latent tokens are needed, which may lead to unnecessary computation on simpler ones.



%% file: sec/3_method.tex
\section{Method} \label{sec:3}
In this section, we first present an overview of the reasoning pipeline of BEE (Section~\ref{sec:3.1}). We then describe the two-stage training strategy for the implicit visual tool paradigm (Section~\ref{sec:3.2}), followed by the self-regulated reward (MC-Reward), which enables the model to achieve self-regulated tool invocation (Section~\ref{sec:3.3}).

\input{table_figs/fig0_main}

\subsection{Overall Pipeline}
\label{sec:3.1}
{\MethodName} is a multimodal large language model with self-regulated decision-making capabilities. Its core idea is to assess task difficulty according to its own capability boundary and to trigger an implicit tool-augmented reasoning process when needed. Given a question $Q$ and an input image $I \in \mathbb{R}^{H \times W \times C}$, the inference procedure of {\MethodName} proceeds as follows. The model (parameterized as ${\pi}_\theta$) first produces an initial reasoning plan $R$ to decide whether additional visual operations, such as local zooming, viewpoint rotation, or auxiliary line drawing, are necessary to solve the task. 
This self-regulated decision process can be described using an implicit routing variable $g \in \{0, 1\}$. When $g=1$, indicating that the task exceeds the model's direct reasoning capability and requires implicit tool usage, {\MethodName} does not rely on external API-based tool invocation.
Instead, it activates an internal mechanism (tool slot) through a control token such as $<$tool\_start$>$ and predicts, in an autoregressive manner, image-level semantic feature tokens that correspond to the intended visual operations, denoted as $\mathcal{A} \in \mathbb{R}^{K \times d}$, where $K$ is the sequence length and $d$ is the hidden feature dimension. These tokens are extracted offline by a pretrained semantic tokenizer and are stored within the model, serving as implicit representations of simulated visual feedback. The model then integrates these implicit tokens into its chain-of-thought reasoning process to support reasoning in complex scenarios. When the model judges that its capability is sufficient, it skips the tool invocation stage and directly produces the final answer. This dynamic routing mechanism reduces latency from external tool calls and reduces cascading errors in simpler cases, achieving a better trade-off between efficiency and accuracy.

The overall framework of BEE is illustrated in Figure~\ref{fig:fig0_main}. Specifically, BEE consists of two training stages: 1) Formalized CoT Supervised Fine-Tuning, which constructs formalized CoT labels to train the multimodal large language model. This stage equips the model with implicit tool representations, adaptive tool switching, and fundamental reasoning capabilities, while also laying the foundation for low-latency inference.
2) Self-regulated Reward-Driven Alignment, which focuses on refining the model's tool usage behavior. We first introduce the Net Tool Gain metric to quantitatively analyze the model obtained after supervised fine-tuning and observe that it tends to overuse tools due to ambiguous cognitive boundaries. To alleviate this issue, we propose a self-regulated reward mechanism that penalizes ineffective tool dependency and encourages the model to perform knowledge routing. Consequently, implicit tools are activated only when the model's internal knowledge is insufficient, leading to more efficient and adaptive reasoning.

{\MethodName} supports self-regulated optimization of its reasoning process. Compared with conventional explicit tool invocation methods based on external APIs, this implicit framework has three main advantages. 
\begin{itemize}
    \item \textbf{End-to-end joint optimization}, where the generation of implicit tool features and the reasoning process are optimized within a unified framework, reducing the separation between modules in traditional pipelines and easing cascading error accumulation. 
    \item \textbf{Improved inference efficiency}, since the method removes system-level latency caused by external interaction and repeated image encoding. The model performs tool-augmented reasoning in a single forward pass, improving throughput and reducing time per output token, which leads to more efficient inference. 
    \item \textbf{Capability-aware adaptive triggering}, where the self-regulated mechanism enables the model to activate implicit tools only when the task exceeds its internal capability boundary, while avoiding unnecessary tool usage when the model is already sufficient. This design enables more effective routing and more efficient use of computational resources.
\end{itemize}

\input{table_figs/fig3_data2}
\subsection{Self-Regulation-Driven IvT Framework}
\label{sec:3.2}

\subsubsection{Stage 1: Formalized CoT Supervised Fine-tuning} The BEE-320K dataset is constructed through the following steps.

\begin{enumerate}
  \item \textit{\textbf{Source Data Collection}}. 
  Inspired by the way humans adapt their strategies and flexibly use different skills when solving complex problems, BEE is designed to mimic this cognitive process during data collection. To this end, we integrate heterogeneous data from Refocus~\cite{fu2025refocus}, Thyme~\cite{zhang2025thyme}, mini-o3~\cite{lai2025mini}, and Visual-CoT~\cite{shao2024visual} to build a dataset with diverse task types and image distributions. This design provides broad domain coverage and sufficient sample diversity for training.
  \item \textit{\textbf{Trajectory Construction and Polishing}}. 
  The training data is constructed as follows. We first use the code sandbox environment in Thyme to generate reasoning trajectories with explicit tool invocation. Each trajectory consists of an initial plan ($R$), code-based tool operations ($T$), subsequent reasoning ($R_t$), and the final answer ($A$). During execution, we discard samples with non-executable code and use Qwen3-VL-30B to verify whether the outputs of the executed code are consistent with the intended objective reflected in the model's analysis, such as improving image readability through rotation. Samples that fail this verification are removed. We then transform explicit tool invocation into implicit representations. Using a pretrained tool semantic tokenizer (which is described in {\color{blue} \bf APPENDIX \ref{app:appendix_a3}}), we encode the auxiliary images into semantic tokens (Tool\_pad) and replace the code segment $T$ in the original reasoning trajectory with these tokens $\mathcal{A}$. To maintain textual fluency, we further use Qwen3-VL-30B to rewrite the trajectory while preserving the original reasoning logic, ensuring smooth transitions between the semantic tokens and the surrounding context. The resulting dataset contains implicit tool-based reasoning chains and covers six visual operations, including zooming, cropping, and rotation, \etc. The diversity of tasks allows BEE to learn a broad set of fundamental implicit tool-use skills. In addition, to equip BEE with an initial capability for adaptive tool selection, namely the ability to solve problems without tools when they are unnecessary, we collect additional samples from Thyme that do not require tool intervention and construct a basic training subset.
  \item \textit{\textbf{Implicit Token Understanding}}. 
  Our preliminary experiments show that implicit tool-based reasoning chains do not provide substantial improvements over the no-tool baseline on several evaluation metrics. We find that the main reason is the model's limited ability to distinguish and understand the newly introduced visual tokens. To address this representation bottleneck, we construct an auxiliary dataset to strengthen semantic alignment. In this dataset, implicit tool tokens are provided as input, and the model is trained to predict their corresponding visual descriptions through a caption prediction task. Further details can be found in {\color{blue} \bf APPENDIX \ref{app:detail_bee_data}}.
\end{enumerate}
Combining the above components, we construct the final BEE-SFT-320K dataset. It mainly contains three categories of data. The proportion of each category and the distribution of reasoning trajectory lengths are shown in Figure~\ref{fig:fig3_data2}.
More details of the data are provided in {\color{blue} \bf APPENDIX \ref{app:detail_bee_data}}, and the format of the final SFT dataset is described in {\color{blue} \bf APPENDIX \ref{app:sft_rl_example}}.



\subsubsection{Net Tool Gain (NTG) Metric} 
\input{table_figs/fig3_motivation}
To understand how the model uses external tools, we first examine the distribution of tool invocations made by the BEE-Static model (\eg, trained via SFT combined with a simple DAPO~\cite{Yu2025DAPOAO} algorithm) and the baseline model on correctly and incorrectly predicted samples from the MME-Real-Lite benchmark, as shown in Figure~\ref{fig:fig3_motivation}. The initial observations suggest that the model tends to over-rely on external tools. To quantify the effectiveness of tool usage, we perform inference twice on the same test set, once without tool access (Base) and once with tool access enabled (BEE-Static). Let $\mathcal{D}$ denote the test set. We define $S_{base}$ as the set of samples correctly predicted under the base mode, $S_{bee}$ as the set of samples correctly predicted under the tool-enabled mode, and $C_{tool}$ as the set of samples for which the model invokes external tools during inference. Based on these definitions, we partition $C_{tool}$ into three mutually exclusive subsets.
\begin{itemize}
    \item \textbf{Effective Tool Gains}. Effective Tool Gains are defined as $G = (S_{bee} \setminus S_{base}) \cap C_{tool}$, which consist of samples that are incorrectly predicted by the base model but become correct after tool invocation. 
    \item \textbf{Redundant Tool Calls}. Redundant Tool Calls are defined as $R = (S_{base} \cap S_{bee}) \cap C_{tool}$, which include samples that can already be answered correctly using the model's internal knowledge but still trigger unnecessary tool usage. 
    \item \textbf{Tool-Induced Hallucinations}. Tool-Induced Hallucinations are defined as $L = (S_{base} \setminus S_{bee}) \cap C_{tool}$, which correspond to samples that are correctly answered by the base model but become incorrect after being misled by external tool outputs.
\end{itemize}

To measure the actual marginal benefit of tool invocation, we introduce the Net Tool Gain (\textbf{NTG}) metric (\eg, Equation~\ref{eq:eq_ntg}).
\begin{equation}
\label{eq:eq_ntg}
    NTG = \frac{|G| - |L|}{|C_{tool}|}.
\end{equation}
This metric measures the average net performance gain brought by tool usage. Substituting the empirical results into the above equation, we find that the NTG of BEE-Static is only \textbf{6.2\%}. Together with the distribution shown in Figure~\ref{fig:fig3_motivation} and the extremely low NTG value, these results indicate that although the BEE-Static learns to invoke tools, it still lacks a clear understanding of its own capabilities and limitations. As a result, it produces a large number of costly and redundant tool calls. This observation directly motivates the introduction of the self-regulated reward mechanism, MC-Reward, in the next stage to encourage the model to invoke tools only when necessary.

\subsubsection{Stage 2: Self-regulated Reward-driven Alignment} 
The policy model $\pi$ samples $G$ output trajectories, denoted as ${o_1, o_2, \dots, o_G}$. To improve both the robustness and reasoning ability of the model, we design a composite reward function that jointly optimizes answer correctness, output format compliance, and the logical consistency of the reasoning process, while regulating tool invocation through the self-regulated reward (MC-Reward). For each trajectory $o_i$, the overall reward is defined as Equation~\ref{eq:eq_all}.
\begin{equation}
\label{eq:eq_all}
R(o_i)=r_{acc}+r_{cons}+r_{fmt}+r_{mc}.
\end{equation}
\begin{itemize}
    \item \textbf{Answer Accuracy ($r_{\text{acc}}$)}. The accuracy reward $r_{acc}$ evaluates whether the final answer is correct. We adopt a hierarchical evaluation strategy. Rule-based methods, including exact string matching, option matching, and regular expression matching, are first applied for efficient verification. A reward of 1 is assigned if a match is found. For complex responses that cannot be reliably verified by these rules, such as long or ambiguous answers, we use a strong vision-language model, such as Qwen3-VL-4B, as the judge model to assess the semantic equivalence between the predicted answer and the reference answer. The reward is set to 1 if the two answers are semantically equivalent and 0 otherwise.
    \item \textbf{Reasoning-answer Consistency ($r_{\text{cons}}$)}. The consistency reward $r_{cons}$ measures whether the reasoning process is semantically consistent with the final answer. We also use the judge model, such as Qwen3-VL-4B, to examine the relationship between the generated reasoning trajectory and the predicted answer. The reward is assigned as 1 when the reasoning process supports the predicted answer and the two are semantically consistent. Otherwise, the reward is set to 0.
    \item \textbf{Format Adherence ($r_{\text{fmt}}$)}. The format reward $r_{fmt}$ ensures that the output trajectory follows a predefined structured template, such as specific tool slot tags. A reward of 1 is assigned when the output satisfies the formatting requirements and 0 otherwise.
    \item \textbf{Self-regulated Confidence calibration ($r_{\text{mc}}$)}. The self-regulated reward $r_{mc}$ aims to help the model establish clear cognitive boundaries, and its mathematical formulation is presented in Section~\ref{sec:3.3}. This reward penalizes ineffective reliance on external tools and encourages the model to invoke implicit tools only when its internal parametric knowledge is insufficient to solve the task, thereby reducing redundant computation.
\end{itemize}
The optimization objective of DAPO~\cite{Yu2025DAPOAO} is defined as
\begin{equation}
\begin{aligned}
\mathcal{J}_{\text{DAPO}}(\theta) = \quad&\mathbb{E}_{(Q,A)\sim\mathcal{D}, \{o_i\}_{i=1}^G\sim \pi_{\theta_\text{old}}(\cdot\mid Q)}\\&
\Bigg[\frac{1}{\sum_{i=1}^{G}|o_i|}\sum_{i=1}^{G}\sum_{t=1}^{|o_i|} 
\min \Big( r_{i,t}(\theta) \hat{A}_{i,t},  \\&
\ \text{clip} \Big( r_{i,t}(\theta), 1 - \varepsilon_{\text{low}}, 1 + \varepsilon_{\text{high}} \Big) \hat{A}_{i,t} \Big) \Bigg]
\label{eq:dapoloss_clip_higher}
\end{aligned}
\end{equation}
\subsection{MC-Reward: Self-regulated Confidence Calibration}
\label{sec:3.3}
MC-Reward is designed to discourage ineffective reliance on external tools and encourage the model to make informed decisions about knowledge routing, ensuring that implicit tools are invoked only when the model's internal parametric knowledge is insufficient to solve the current task.

\subsubsection{When to Use} 
A key challenge is how to characterize the model's current capability boundary. In this work, we use the group average accuracy during the rollout stage as a dynamic signal of the model's capability. Specifically, given a query $x$, we compute the average accuracy $p$ over all $G$ sampled trajectories within the same rollout group:
\[
p=\frac{1}{G}\sum_{k=1}^{G}\mathbb{I}(\text{correct}(y^{(k)},x)),
\]
where $G$ denotes the number of sampled trajectories in a rollout group and $\mathbb{I}(\cdot)$ is the indicator function. Intuitively, $p$ reflects how likely the model is to solve the query using only its internal knowledge. During training, $p$ is updated dynamically at each rollout, allowing BEE to adjust its tool usage behavior according to both its current capability and the difficulty of the query.
When $p$ falls below a predefined threshold $\tau$, the query is considered to be beyond the model's current capability. In this case, MC-Reward encourages the model to invoke implicit tools so that it can obtain fine-grained visual cues. Conversely, when $p \geq \tau$, the model is expected to solve the task using its internal knowledge alone, and unnecessary tool invocations are penalized.

Overall, MC-Reward encourages implicit tools to focus on challenging queries that exceed the model's current capability. This mechanism not only enables the model to develop an awareness of its own capability boundary, but also improves the efficiency of computational resource usage during inference.

\subsubsection{Propensity Shaping} 
The primary motivation for introducing MC-Reward is to enable BEE to decide whether to invoke external tools according to its current capability and the difficulty of the task. To quantify and regulate this behavior, we use the capability indicator obtained in the previous stage, denoted by $p \in [0,1]$, where $p > \tau$ indicates an easy task and $p < \tau$ indicates a difficult one. Our goal is to define a tool advantage function $\Delta \mathcal{A}(p)$:
\begin{equation}
\Delta \mathcal{A}(p)=\mathcal{R}_{\text{tool}}-\mathcal{R}_{\text{notool}}.
\end{equation}

This function is expected to satisfy the following boundary conditions:
\begin{equation}
\begin{cases}
\Delta \mathcal{A}(p) < 0, & \text{as } p > \tau \quad (\text{Easy tasks, discourage}), \\
\Delta \mathcal{A}(p) \geq 0, & \text{as } p \leq \tau \quad (\text{Difficult tasks, encourage}).
\end{cases}
\label{eq:eq_target}
\end{equation}

In our formulation, whether the model invokes a tool, represented by $\mathbb{I}_{\text{tool}} \in {0,1}$, and whether the prediction is correct, represented by $\mathbb{I}_{\text{acc}} \in {0,1}$, jointly define a $2 \times 2$ action-outcome space. Based on this formulation, we define the gating reward $R_{\text{gating}}(t)$ as
\begin{equation}
\Delta R_{\text{mc}}(t)=
\mathbb{I}_{\text{tool}}
\cdot
\mathcal{F}(p,\mathbb{I}_{\text{acc}})
+
(1-\mathbb{I}_{\text{tool}})
\cdot
\mathcal{G}(p,\mathbb{I}_{\text{acc}}).
\end{equation}

To avoid training instability caused by discontinuous rewards, we construct the reward using smooth and continuous functions. We first define the base reward for both correct and incorrect predictions as a linear function of the capability indicator:
$R_{base}=\alpha+\beta p$.
For tool usage, we introduce a cosine-based gating term:
$\Omega_{buffer}=r\cos \pi x+u$.
The reward functions for the tool and no-tool cases are then defined as
\begin{equation}
\begin{array}{l}
\mathcal{F}(p, \mathbb{I}_{\text{acc}}) = \mathbb{I}_{\text{acc}} \cdot \left[ R_{\text{base1}} + \Omega_{\text{buffer1}}(p) \right] + \\
(1 - \mathbb{I}_{\text{acc}}) \cdot \left[ R_{\text{base2}} + \Omega_{\text{buffer2}}(p)\right].
\end{array}
\end{equation}
\begin{equation}
\mathcal{G}(p, \mathbb{I}_{\text{acc}}) = \mathbb{I}_{\text{acc}} \cdot  R_{\text{base1}} + (1 - \mathbb{I}_{\text{acc}}) \cdot R_{\text{base2}}.
\end{equation}

This smooth design naturally satisfies the desired boundary conditions and allows the reward signal to vary continuously with model capability, leading to more stable policy updates during training. Detailed visualizations and analyses are provided in Section~\ref{sec:ana_reward}.
Overall, MC-Reward introduces a difficulty-aware intrinsic reward that discourages unnecessary tool usage on easy problems while encouraging tool-assisted reasoning on challenging ones.

%% file: table_figs/fig0_main.tex
\begin{figure*}[t]
    \centering
    \includegraphics[width=1.0\linewidth]{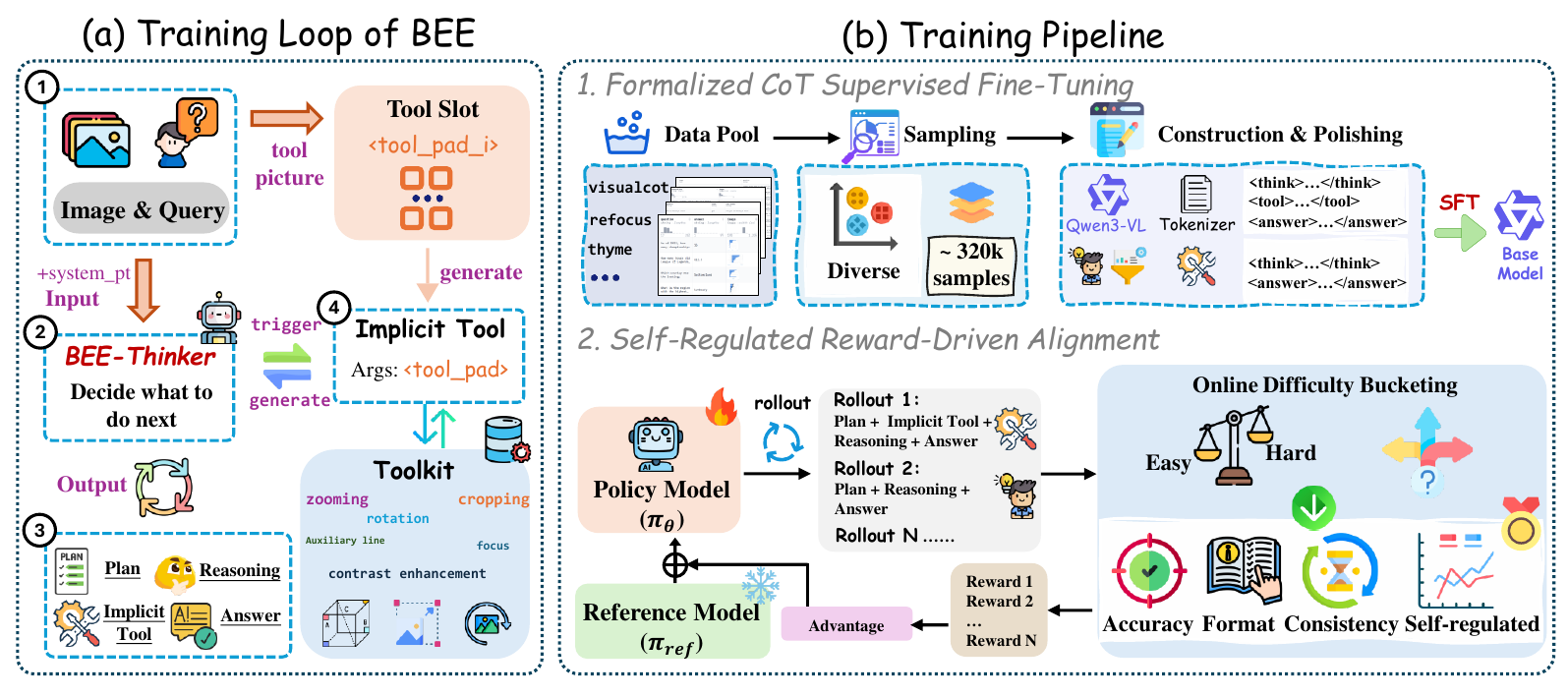}
    \caption{
        (a) \textbf{Overview of BEE}. BEE adaptively selects between implicit visual tool behavior and direct reasoning based on the model's current capability. (b) \textbf{Two-Stage Training Pipeline}. In {\color{gray}\textit{Stage 1 Formalized CoT Supervised Fine-Tuning}}, the model is trained using pre-constructed formalized CoT labels to activate the model's implicit tool representations, adaptive switching capability, and basic reasoning ability. In {\color{gray}\textit{Stage 2 Self-Regulated Reward-Driven Alignment}}, a self-regulated reward objective is introduced during training, allowing the model to learn when to rely on its own capability and when to invoke implicit visual tools.        
    }
    \label{fig:fig0_main}
\end{figure*}

%% file: table_figs/fig3_data2.tex
\begin{figure}[t]
    \centering
    \includegraphics[width=0.99\linewidth]{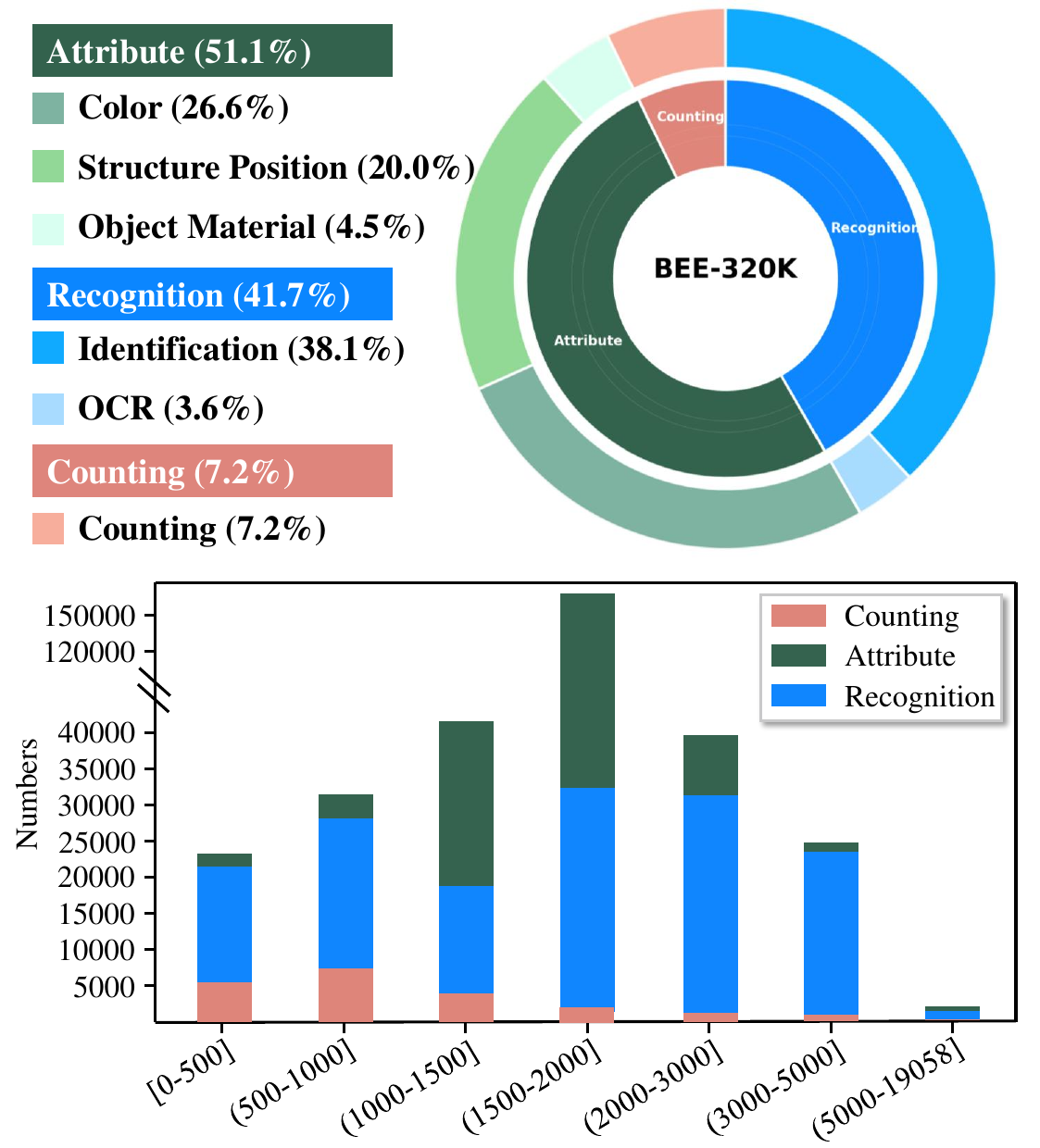}
    \caption{
        {\bf Statistics of SFT training data.} \textbf{Top}: The distribution of tasks across three main categories.  \textbf{Bottom}: The data length statistic.
    }
    \label{fig:fig3_data2}
\end{figure}

%% file: table_figs/fig3_motivation.tex
\begin{figure}[ht]
    \centering
    \includegraphics[width=0.8\linewidth]{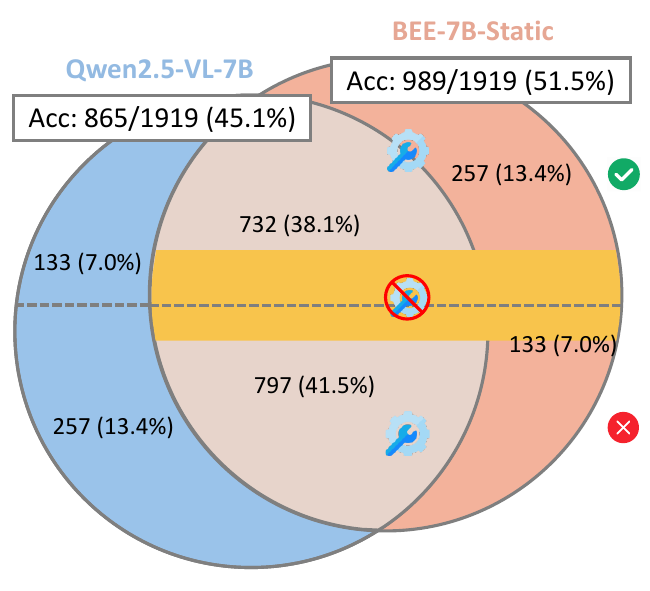}
    \caption{
        {\bf Data-driven comparison of performance and tool usage rate.} BEE-7B-Static still makes extensive use of tools even within the range of its own capabilities (\eg, problems solvable by the base model), leading to unnecessary resource consumption and indicating a lack of capability boundary awareness.
    }
    \label{fig:fig3_motivation}
\end{figure}

%% file: sec/4_exp.tex
\section{Experiments} \label{sec:4}
\input{table_figs/tab0_main}
\input{table_figs/tab3_fig_compare}
\subsection{Experimental Settings}
\noindent {\bf Models}.
We use Qwen2.5-VL-7B-Instruct~\cite{bai2025qwen2} and Qwen3-VL-4/8B-Instruct~\cite{bai2025qwen3} as base models. All these models first train on {\MethodName}-SFT-320K datasets, and then train on 55K RL data with DAPO~\cite{Yu2025DAPOAO}.


\noindent {\bf Evaluation metric}.
We evaluate on eight benchmarks in real-world understanding, general reasoning and out-of-distribution generalization.
\textit{\textbf{(i)} Real-world Understanding:} V\textsuperscript{*}~\cite{wu2024v}, HR-Bench~\cite{wang2025divide}, MME-RealWorld~\cite{zhang2024mme}.
\textit{\textbf{(ii)} Out-of-distribution Generalization:} 
General Reasoning: LogicVista~\cite{xiao2024logicvista},
Hallucination: POPE~\cite{li2023evaluating}.
\textit{V*}~\cite{wu2024v} is a benchmark comprising high-resolution, visually cluttered images designed to evaluate the fine-grained visual grounding and reasoning abilities of VLMs in visual search scenarios.
\textit{HR-Bench}~\cite{wang2025divide} is an ultra-high-resolution image benchmark designed to evaluate how well VLMs preserve fine visual details across both fine-grained single-instance perception and cross-instance reasoning tasks.
\textit{MME-RealWorld}~\cite{zhang2024mme} is a large-scale, high-resolution benchmark spanning five domains and 43 subtasks, designed to evaluate models on complex real-world scenarios that demand precise local detail perception and multi-step reasoning.
\textit{LogicVista}~\cite{xiao2024logicvista} is a benchmark featuring 448 samples across five distinct skills, designed to comprehensively evaluate the logical reasoning capabilities of multimodal LLMs.
\textit{POPE}~\cite{li2023evaluating} is a benchmark designed to evaluate object hallucination in MLLMs. We set the threshold $\tau$ to $2/3$ in our experiments.

\noindent {\bf Baselines}. 
To ensure a comprehensive evaluation, we benchmark our {\MethodName} against five distinct categories of baselines. \textit{\textbf{(i)} closed-source frontier models}, such as Gemini~\cite{comanici2025gemini}. \textit{\textbf{(ii)} strong open-source models}, including Kimi~\cite{team2026kimi}, GLM~\cite{hong2025glm}, and Qwen3~\cite{bai2025qwen3}. \textit{\textbf{(iii)} base models}, including the Qwen2.5-VL~\cite{bai2025qwen2}/Qwen3-VL~\cite{bai2025qwen3} series.  \textit{\textbf{(iv)} thinking with images agentic models}, including Deepeyes~\cite{zheng2025deepeyes}, DeepeyesV2~\cite{hong2025deepeyesv2}, Mini-o3~\cite{lai2025mini}, and Thyme~\cite{zhang2025thyme}. \textit{\textbf{(v)} models fine-tuned on open-source datasets}, including Pixel-Reasoner~\cite{su2025pixel}. \textit{\textbf{(vi)} implicit latent reasoning models}, such as Monet~\cite{wang2026monet} and EVA~\cite{chen2026latent}. Notably, other training-free methods are excluded from the main comparative results, as these techniques are fundamentally orthogonal and can be directly integrated to further enhance our proposed method. 

\noindent {\bf Efficiency comparisons setting}.
All efficiency comparisons are conducted on a single NVIDIA H800 GPU with a consistent NVIDIA driver version. During evaluation, no other processes interfere with the system. All methods are tested using BF16 precision with a batch size of 1, and are executed with VLLM for inference, except for Thyme, which is evaluated using the raw PyTorch implementation provided in its official codebase. Detailed environment configurations for the efficiency comparisons are provided in {\color{blue} \bf APPENDIX \ref{app:effi}}, and the training hyperparameters are detailed in {\color{blue} \bf APPENDIX \ref{app:hyper}}.

\subsection{Main Results}
\noindent {\bf Leading performance across various benchmarks}.
Table~\ref{tab:tab0_main} compares BEE with other leading multimodal models on perception, reasoning, and general-purpose tasks. On perception benchmarks, BEE outperforms current state-of-the-art open-source models, including GLM-4.5V, Qwen3-VL-235B, and Kimi-k2.5/2.6, whose model sizes range from 108B to 1T parameters. BEE achieves an average accuracy of 75.3, exceeding the strongest baseline by 1.2 points. This result suggests that test-time scaling remains an important factor for improving performance on perception tasks. On out-of-distribution benchmarks such as general reasoning, BEE consistently surpasses its backbone model, Qwen-VL-Base, indicating that the observed improvements are not simply the result of overfitting to specific perception benchmarks. The gains in both perception and reasoning also lead to a substantial reduction in hallucinations, resulting in more reliable model outputs.

\noindent {\bf Exceeding "thinking with image" models}.
We further compare BEE with agentic models. Table~\ref{tab:tab3_compare} shows that BEE-7B achieves performance that is comparable to or better than other models in the same family, and consistently outperforms them at both the 4B and 8B scales. These results suggest that BEE's implicit tool execution process can effectively replace explicit tools, eliminating the need for frequent tool invocation and re-encoding.

\input{table_figs/tab4_training_time}

\noindent {\bf Surpassing latent reasoning models}.
We further compare BEE with latent reasoning models. As shown in Table~\ref{tab:tab3_compare}, BEE consistently outperforms latent reasoning models across different model scales on all benchmarks except MME-Lite, improving the average accuracy from 69.5 to 76.8. These results indicate that BEE's self-regulation tool internalization mechanism is more effective than a purely latent reasoning paradigm.

\noindent {\bf Time-efficient training for tool competence}.
Table~\ref{tab:tab4_training_time} compares the training data size and training time of BEE with those of other models. BEE requires considerably less training time than DeepEyesV2 and Thyme while maintaining higher efficiency, which substantially reduces the reliance on GPU and storage resources.

\noindent {\bf Lower inference latency via internalizing tool-use benefits}.
Figure~\ref{fig:fig3_compare} presents the performance and inference speed of BEE compared with agentic models across different scales. BEE not only achieves higher performance but also runs faster, with BEE-7B improving from 0.07 sample/s to 1.61 sample/s (HRBench4K). 
Figure~\ref{fig:fig0_first} further presents a comparison of the BEE model across different resolutions. For example, at 8K resolution, the BEE method achieves a $\sim 86.2\%$ improvement in inference speed compared to Thyme.

\noindent {\bf Cost-effective deployment with performance advantages}.
We further evaluate the deployment efficiency and cost of different models in practical scenarios (\cf, Table~\ref{tab:tab5_cost}). Due to its autoregressive design, BEE can be deployed as easily as a standard Qwen-VL-Instruct model, without relying on external sandbox interfaces required by DeepEyesV2. Compared with the DeepEyes series, BEE achieves lower average latency, higher throughput, and lower operational cost, reducing the cost from 6.4\$ and 1.28\$ to 0.72\$.
\input{table_figs/tab5_system_metrics}

\noindent {\bf Additional results}.
The pseudocode for computing these system metrics is provided in {\color{blue} \bf APPENDIX ~\ref{app:metrics}}. The ablation study of $\tau$ is in {\color{blue} \bf APPENDIX ~\ref{app:add_tau}}. More comprehensive experimental results are reported in {\color{blue} \bf APPENDIX ~\ref{app:add_2}}, and the training curves of the model are shown in {\color{blue} \bf APPENDIX ~\ref{app:curves}}. Case studies, including both successful and failed examples across real-world scenarios, mathematics, general reasoning, autonomous driving, and medical applications, are presented in {\color{blue} \bf APPENDIX ~\ref{app:sucess}}, {\color{blue} \bf APPENDIX ~\ref{app:fail}}.

\subsection{Ablation Studies}
\input{table_figs/tab6_num_part}
\input{table_figs/tab2_split_dataset_mme_lite}
\input{table_figs/tab12_ntg}

\noindent {\bf Importance of tool slots}.
To verify the effectiveness of the implicit visual tools, we construct several variants. Specifically, we replace the target image used for implicit tool representation with the original image, substitute the implicit visual tokens with meaningless $\langle$unk$\rangle$ tokens of the same length that are added to the CodeBook, and remove the implicit tools entirely, relying instead on pure text reasoning. The detailed designs of these variants are provided in  {\color{blue} \bf APPENDIX ~\ref{app:alternative_form}}. We train all variants using SFT+DAPO with the same amount of training data. Table~\ref{tab:tab6_tool} presents their performance under this controlled setting.

\begin{itemize}
    \item \textbf{Replacing the target image with the original image.} This causes the largest drop on HRBench relative to BEE-7B, particularly at the 8K resolution. This result suggests that the target image, which provides a clearer view of the object, removes distracting background content, and includes auxiliary guidance cues, is important for accurate visual perception. 
    \item \textbf{Replacing the implicit visual tokens with meaningless $\langle \text{unk} \rangle$ tokens.} This leads to consistent declines across all evaluation metrics. This finding indicates that the performance gains come from the implicit tool design itself rather than from the additional reasoning time associated with longer output sequences. 
    \item \textbf{Pure text-based reasoning.} When the implicit tools are removed and the model relies solely on text-based reasoning, performance also drops consistently across all metrics. These results show that the implicit tools improve image understanding and, in turn, support more effective reasoning.
\end{itemize}

\noindent {\bf Token length comparison for tool slots}.
We further compare the effect of different numbers of implicit tool tokens on model performance. As shown in Figure~\ref{fig:fig6_token_length}, BEE shows only small performance variations across different token lengths, and even with as few as four discrete tokens, it already achieves strong results.

\noindent {\bf Delve deep into perception tasks}.
Taking MME-Real-Lite as an example, it includes a large number of high-resolution perception tasks in real-world scenarios. In Table~\ref{tab:tab2_splitdataset_lite}, we present the performance of BEE and the baseline model on various tasks. It can be observed that for tasks on which the baseline model already performs well, such as OCR, Diagram and Table tasks, with accuracies exceeding 65\% and in some cases approaching 90\%, the improvement provided by BEE is limited (\eg, some metrics show a slight decline). However, for more difficult tasks on which the perception ability of Qwen2.5-VL-7B is relatively weak, such as monitoring and autonomous driving, BEE achieves improvements of more than 25\% in both perception and reasoning tasks, with a more pronounced gain in reasoning tasks.

\noindent {\bf Comparison of NTG metrics}.
Table~\ref{tab:tab12_ntg} presents the NTG comparison among conventional methods, BEE-static, and BEE-7B. The results show that the MC-Reward design substantially improves the true gain from tool use. Moreover, BEE-7B achieves a higher NTG than conventional methods, indicating that it derives greater practical benefit from tool use.


\section{Deeper Analysis}
\label{sec:analysis}

\subsection{Analysis and Visualization of MC-Reward}
\label{sec:ana_reward}
\input{table_figs/fig5_reward_curve}
\input{table_figs/fig6_tool_change}
As model capabilities improve, some problems that were previously difficult become easier. In this context, the implicit tool should adjust dynamically to changes in the model's ability. We first describe the tendency analysis and the visualization analysis, with additional derivations provided in {\color{blue} \bf APPENDIX ~\ref{app:sec_analysis_mcreward}}.

In the tendency analysis, following the formulation presented in Section~\ref{sec:3.3}, cases are classified based on whether a tool is used and whether the final answer is correct. These cases are divided into four categories, as shown in Equation~\ref{eq:all_first}. 
    \begin{equation}
    \label{eq:all_first}
        \Delta R(t) = 
        \begin{cases} 
         \alpha_1+ \beta_1p + \gamma \cos(\pi p) + \mu, & \text{if } \mathbb{I}_{\text{tool}} = 1, \mathbb{I}_{\text{acc}} = 1 \\
        \alpha_2 + \beta_2p + \gamma \cos(\pi p) + \mu, & \text{if } \mathbb{I}_{\text{tool}} = 1, \mathbb{I}_{\text{acc}} = 0 \\
        \alpha_1 + \beta_1p, & \text{if } \mathbb{I}_{\text{tool}} = 0, \mathbb{I}_{\text{acc}} = 1 \\
        \alpha_2 + \beta_2p, & \text{if } \mathbb{I}_{\text{tool}} = 0, \mathbb{I_{\text{acc}}} = 0
        \end{cases}
    \end{equation}

The constraints and derivations are provided in {\color{blue} \bf APPENDIX ~\ref{app:sec_analysis_mcreward}}. Based on the tendency analysis, the final formulation applied in the main experiments is as follows.
    \begin{equation}
        \Delta R(t) = 
        \begin{cases} 
         0.45 - 0.2p + 0.3 \cos(\pi p) , & \text{if } \mathbb{I}_{\text{tool}} = 1, \mathbb{I}_{\text{acc}} = 1 \\
        -0.35 + 0.3p + 0.3 \cos(\pi p), & \text{if } \mathbb{I}_{\text{tool}} = 1, \mathbb{I}_{\text{acc}} = 0 \\
        0.3 - 0.2p, & \text{if } \mathbb{I}_{\text{tool}} = 0, \mathbb{I}_{\text{acc}} = 1 \\
        -0.5 + 0.3p, & \text{if } \mathbb{I}_{\text{tool}} = 0, \mathbb{I_{\text{acc}}} = 0
        \end{cases}
    \label{eq:eq_last_reward}
    \end{equation}

Figure~\ref{fig:fig5_reward} (Left) shows the reward values defined in Equation~\ref{eq:eq_last_reward} for easy and difficult problems. For easy problems, correct answers achieved without using external tools receive a higher reward, while using tools and still providing an incorrect answer results in a larger penalty. For difficult problems, correct answers obtained with tool assistance are rewarded more. If the answer remains incorrect, the reward encourages continued use of tools to support further reasoning and problem solving. This design aligns with the objective described in Equation~\ref{eq:eq_target}. We also provide visualizations for other parameter values in the right Figure. The final model exhibits a similar trend to the one shown in the left Figure.

\input{table_figs/fig11_mc_performance}

\subsection{Analysis of Implicit Tool Usage}
Fig~\ref{fig:fig6_tool_change} shows how the tool call rate of the BEE model changes throughout training. 
\begin{itemize}
    \item \textbf{For easy problems.} The initial model obtained from the SFT stage shows limited metacognitive awareness and often makes unnecessary tool calls on easy problems. As training proceeds, the MC-reward mechanism helps the model develop better metacognitive judgment in these cases, allowing it to solve them using its own capability without relying on external tools. This leads to a steady improvement in accuracy, reflected in an increasing number of easy problems being solved without tool usage.
    \item \textbf{For difficult problems.} The model adjusts its tool-use behavior in the early stage, and the call rate gradually increases as training continues. The model keeps attempting to solve these challenging cases, and the number of remaining difficult problems gradually decreases, with the tool call rate finally stabilizing at around 41\%. It is also observed that some problems exceed the upper bound of the model’s capability, where further tool usage does not lead to correct solutions. 
\end{itemize}
For more details on tool calling rates, changes across varying difficulty levels during training, etc., please refer to {\color{blue} \bf APPENDIX ~\ref{app:curves}}.

\subsection{The Impact of MC-Reward on Performance}
\input{table_figs/tab11_mc_metrics}
Both BEE-Static and BEE-MC significantly outperform the baseline model, as shown in Figure~\ref{fig:fig11_mc}. The introduction of MC-Reward enables the model to use tools more appropriately and gradually develop self-regulated ability, which improves reasoning efficiency. It also leads to better results on some metrics compared to BEE-Static, suggesting that enhanced self-regulated ability makes the model more capable.
Table~\ref{tab:tab11_mc_metrics} demonstrates that introducing MC-reward results in lower latency and optimized costs when compared to BEE-Static.

\subsection{Analysis of Implicit Tokens}
\input{table_figs/fig12_mask}
\input{table_figs/tab9_replace}
\input{table_figs/tab10_decouple}
Figure~\ref{fig:fig12_mask} illustrates the effectiveness of implicit tokens using two ablation strategies. 
\begin{itemize}
    \item \textbf{Masked.} During inference, these implicit tokens are masked so that the model does not rely on them for subsequent generation. We observe that the model tends to repeatedly output these implicit tokens, suggesting that later token generation still depends on their overall content.
    \item \textbf{Replaced.} We further replace all implicit tokens with a single shared token. Table~\ref{tab:tab9_replace} compares the performance under different implicit token settings, showing a degradation of about $2-4$ points. This indicates that the structure of the generated implicit tokens is important for subsequent token generation. Together with the impact of the \textless unk\textgreater{} token reported in Table~\ref{tab:tab6_tool}, these results suggest that implicit tokens are beneficial for assisting the model in reasoning.
\end{itemize}

\subsection{Analysis of Decoupled Perception and Reasoning.}
We further decouple the optimization of text and visual tokens by introducing an adapter structure. The optimization of visual tokens is performed within the adapter, while the optimization of text is handled by the MLP in the original attention block. The details is provided in {\color{blue} \bf APPENDIX ~\ref{app:add_6}}.
We observe that this decoupling leads to a clear separation between text and visual representations, as shown by the t-SNE visualization. Table~\ref{tab:tab10_decouple} reports the performance after decoupling, where BEE achieves better results.

%% file: table_figs/tab0_main.tex
\begin{table*}[t]
\centering
\caption{{\bf Performance comparison on various tasks.} For all open-source models, the best performance for each metric is {\bf bolded}, and the second best is {\ul underlined}. BEE consistently improves over the corresponding Qwen-VL baselines.
The result ($^{\dagger}$, $^{\ddagger}$) is collected from \cite{zhang2025skywork} and \cite{bai2025qwen3}, respectively.
}
\label{tab:tab0_main}
\scalebox{0.92}{

\begin{tabular}{lccccccccc}
\toprule
                                        &                                                                        & \multicolumn{6}{c}{Real-World Understanding}                                                                                                                                                                     & \multicolumn{2}{c}{OOD Generalization}                 \\ \cline{3-10} 
\multirow{-2}{*}{Models}                & \multirow{-2}{*}{\begin{tabular}[c]{@{}c@{}}Param\\ Size\end{tabular}} & \multicolumn{1}{l}{\cellcolor[HTML]{F7AA97}RW-Avg} & V*                                    & HR-4K                        & HR-8K                       & MME-Real                    & MME-Real-Lite               & LogicVista                  & \multicolumn{1}{l}{POPE} \\ \midrule
\multicolumn{10}{c}{\textit{Closed-Source Models}}                                                                                                                                                                                                                                                                                                                                           \\ \midrule
{\color[HTML]{656565} Gemini-2.5-Flash~\cite{comanici2025gemini}} & {\color[HTML]{656565} -}                                               & {\color[HTML]{656565} 66.9}                     & {\color[HTML]{656565} 72.3$^{\dagger}$}           & {\color[HTML]{656565} 77.5$^{\dagger}$}  & {\color[HTML]{656565} 73.7$^{\dagger}$} & {\color[HTML]{656565} 60.9$^{\dagger}$} & {\color[HTML]{656565} 50.2$^{\dagger}$} & {\color[HTML]{656565} 60.0$^{\ddagger}$}    & {\color[HTML]{656565} -} \\
{\color[HTML]{656565} Gemini-2.5-Pro~\cite{comanici2025gemini}}   & {\color[HTML]{656565} -}                                               & {\color[HTML]{656565} 74.8}                     & {\color[HTML]{656565} 79.1$^{\dagger}$}           & {\color[HTML]{656565} 83.9$^{\dagger}$}  & {\color[HTML]{656565} 81.5$^{\dagger}$} & {\color[HTML]{656565} 71.3$^{\dagger}$} & {\color[HTML]{656565} 58.3$^{\dagger}$} & {\color[HTML]{656565} 68.7$^{\ddagger}$} & {\color[HTML]{656565} -} \\ \midrule
\multicolumn{10}{c}{\textit{Open-Source Models}}                                                                                                                                                                                                                                                                                                                                             \\ \midrule
\rowcolor[HTML]{F4F7F7} 
Qwen3-VL-Instruct~\cite{bai2025qwen3}                       & 4B                                                                     & \cellcolor[HTML]{F4F7F7}68.3                    & \cellcolor[HTML]{F4F7F7}80.6          & \cellcolor[HTML]{F4F7F7}77.4 & 70.4                        & 63.4                        & 49.7                        & 53.2                        & 88.4                     \\
\rowcolor[HTML]{F4F7F7} 
Qwen2.5-VL-Instruct~\cite{bai2025qwen2}                     & 7B                                                                     & \cellcolor[HTML]{F4F7F7}62.6                    & \cellcolor[HTML]{F4F7F7}76.4          & \cellcolor[HTML]{F4F7F7}68.8 & 65.3                        & 58.3                        & 44.1                        & 39.8                        & 85.8                     \\
\rowcolor[HTML]{F4F7F7} 
Qwen3-VL-Instruct~\cite{bai2025qwen3}                        & 8B                                                                     & \cellcolor[HTML]{F4F7F7}69.1                    & \cellcolor[HTML]{F4F7F7}85.3          & \cellcolor[HTML]{F4F7F7}76.1 & 70.8                        & 64.0                        & 49.3                        & 55.3                        & 87.6                     \\
\rowcolor[HTML]{F4F7F7} 
Qwen2.5-VL-Instruct~\cite{bai2025qwen2}                      & 32B                                                                    & \cellcolor[HTML]{F4F7F7}66.4                    & \cellcolor[HTML]{F4F7F7}81.2          & \cellcolor[HTML]{F4F7F7}73.4 & 70.4                        & 61.0                        & 46.2                        & 54.4                        & 86.2                     \\
GLM-4.5V~\cite{hong2025glm}                                 & 108B                                                                   & 70.5                                            & 83.3                                  & 81.6                         & 74.9                        & 64.1                        & 48.5                        & 62.4                        & 87.5                     \\
Qwen3-VL-Instruct~\cite{bai2025qwen3}                        & 235B                                                                   & 74.1                                            & 80.6                                  & \textbf{83.0}                & \textbf{80.4}               & {\ul 69.4}                  & 55.0                        & \textbf{65.8}               & 88.6                     \\
Kimi-K2.5~\cite{team2025kimi}                                & 1T                                                                     & 70.8                                            & 85.9                                  & {\ul 81.9}                   & 75.4                        & 63.8                        & 47.1                        & 63.7                        & 88.7                     \\
Kimi-K2.6~\cite{team2026kimi}                                & 1T                                                                     & 71.2                                            & 84.8                                  & 79.3                         & 74.8                        & 66.4                        & 50.8                        & {\ul 64.8}                  & \textbf{89.4}            \\ \midrule
\multicolumn{10}{c}{\textit{Our Models}}                                                                                                                                                                                                                                                                                                                                                     \\ \midrule
\rowcolor[HTML]{DDEEFF} 
BEE-4B (Qwen3-VL)                           & 4B                                                                     & \cellcolor[HTML]{DDEEFF}{\ul 74.1}              & \cellcolor[HTML]{DDEEFF}\textbf{91.6} & \cellcolor[HTML]{DDEEFF}78.4 & {\ul 77.0}                  & 68.3                        & {\ul 55.4}                  & 53.4                        & 88.6                     \\
\rowcolor[HTML]{DDEEFF} 
BEE-7B (Qwen2.5-VL)                           & 7B                                                                     & \cellcolor[HTML]{DDEEFF}70.0                    & \cellcolor[HTML]{DDEEFF}83.8          & \cellcolor[HTML]{DDEEFF}75.4 & 73.9                        & 64.4                        & 52.3                        & 43.0                        & {\ul 89.3}               \\
\rowcolor[HTML]{DDEEFF} 
BEE-8B (Qwen3-VL)                           & 8B                                                                     & \cellcolor[HTML]{DDEEFF}\textbf{75.3}           & \cellcolor[HTML]{DDEEFF}{\ul 90.6}    & \cellcolor[HTML]{DDEEFF}80.2 & 76.8                        & \textbf{69.6}               & \textbf{59.4}               & 56.3                        & 87.7                     \\ \bottomrule
\end{tabular}
}
\end{table*}



%% file: table_figs/tab3_fig_compare.tex
\begin{figure*}[t]
    \centering
    \begin{minipage}{0.48\textwidth}
        \centering
        \captionof{table}{{\bf Performance comparison on various tasks.} For all open-source models, the best performance for each metric is {\bf bolded}, and the second best is {\ul underlined}.}
        \label{tab:tab3_compare}
        \resizebox{\linewidth}{!}{%
        \renewcommand{\arraystretch}{1.1}
        \setlength{\tabcolsep}{6pt} 
        \begin{tabular}{lccccc}
        \toprule
                                              & \multicolumn{5}{l}{Real-World Understanding}                                                                                         \\
        \multirow{-2}{*}{Models}              & V*                                    & HR-4K         & HR-8K                              & MME-Real-Lite & \multicolumn{1}{l}{\cellcolor[HTML]{F7AA97}Avg} \\ \midrule
        \multicolumn{6}{c}{\textit{Base Models}}                                                                                                                                     \\ \midrule
        \rowcolor[HTML]{F4F7F7} 
        Qwen3-VL-4B~\cite{bai2025qwen3}                           & 80.6                                  & 77.4          & 70.4                               & 49.7          & 69.5                    \\
        \rowcolor[HTML]{F4F7F7} 
        Qwen2.5-VL-7B~\cite{bai2025qwen2}                         & 76.4                                  & 68.8          & 65.3                               & 44.1          & 63.7                    \\
        \rowcolor[HTML]{F4F7F7} 
        Qwen3-VL-8B~\cite{bai2025qwen3}                           & \cellcolor[HTML]{F4F7F7}85.3          & 76.1          & \cellcolor[HTML]{F4F7F7}70.8       & 49.3          & 70.4                    \\
        \rowcolor[HTML]{F4F7F7} 
        Qwen2.5-VL-32B~\cite{bai2025qwen2}                        & \cellcolor[HTML]{F4F7F7}81.2          & 73.4          & \cellcolor[HTML]{F4F7F7}70.4       & 46.2          & 67.8                    \\ \midrule
        \multicolumn{6}{c}{\textit{Agentic Models}}                                                                                                                                  \\ \midrule
        Pixel Reasoner-7B~\cite{su2025pixel}                     & 84.3                                  & 74.0          & 66.9                               & -             & -                       \\
        DeepEyes-7B~\cite{zheng2025deepeyes}                           & 84.3                                  & 74.0          & 66.9                               & 53.2          & 69.6                    \\
        DeepEyesV2-7B~\cite{hong2025deepeyesv2}                         & 81.8                                  & {\ul 77.9}    & 73.8                               & 51.6          & 71.3                    \\
        Thyme-VL-7B~\cite{zhang2025thyme}                           & 82.2                                  & 77.0          & 72.0                               & 55.2          & 71.6                    \\
        Mini-o3-7B~\cite{lai2025mini}                            & 88.2                                  & 77.5          & 73.3                               & -             & -                       \\ \midrule
        \multicolumn{6}{c}{\textit{Latent Models}}                                                                                                                                   \\ \midrule
        Monet-7B~\cite{wang2026monet}                              & 83.3                                  & 71.0          & 68.0                               & {\ul 55.5}    & 69.5                    \\
        EVA-7B~\cite{chen2026latent}                                & 80.2                                  & 74.0          & 69.5                               & 49.8          & 68.4                    \\ \midrule
        \multicolumn{6}{c}{\textit{Our Models}}                                                                                                                                      \\ \midrule
        \rowcolor[HTML]{DDEEFF} 
        BEE-4B (Qwen3-VL)                         & \cellcolor[HTML]{DDEEFF}\textbf{91.6} & {\ul 78.4} & \cellcolor[HTML]{DDEEFF}{ \textbf{77.0}} & 55.4          & {\ul 75.6}              \\
        \rowcolor[HTML]{DDEEFF} 
        BEE-7B (Qwen2.5-VL)                         & \cellcolor[HTML]{DDEEFF}83.8          & 75.4          & \cellcolor[HTML]{DDEEFF}73.9       & 52.3          & 71.4                    \\
        \rowcolor[HTML]{DDEEFF} 
        \cellcolor[HTML]{DDEEFF}BEE-8B (Qwen3-VL) & {\ul 90.6}                            & \textbf{80.2}          &   {\ul 76.8}                    & \textbf{59.4} & \textbf{76.8}           \\ \bottomrule
        \end{tabular}
        }
    \end{minipage}
    \hfill %
    \begin{minipage}{0.5\textwidth}
        \centering
        \includegraphics[width=\linewidth]{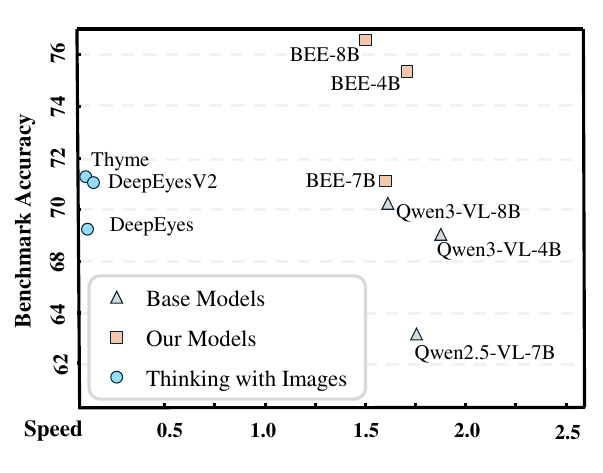}
        \captionof{figure}{\textbf{Benchmark accuracy versus inference speed}. Our models achieve higher accuracy than the base models and agentic baselines while maintaining the efficiency of single-pass inference.} 
        \label{fig:fig3_compare}
    \end{minipage}
\end{figure*}

%% file: table_figs/tab4_training_time.tex
\begin{table}[t]
\centering
\caption{{\bf Comparison of training resources, data scale, and training time.} Note: Some models used different training resources (\eg, 32 GPUs or 8 GPUs).
}
\label{tab:tab4_training_time}
\scalebox{0.8}{

\begin{tabular}{lcccccc} 
\toprule
\multirow{2}{*}{Method} & \multicolumn{2}{c}{Hardware} & \multicolumn{2}{c}{Data Size} & \multicolumn{2}{c}{Time ($\downarrow$)} \\ 
\cmidrule(lr){2-3} \cmidrule(lr){4-5} \cmidrule(lr){6-7} 
& SFT & RL & SFT & RL & SFT & RL \\ 
\midrule
DeepEyes~\cite{zheng2025deepeyes} & - & 4 Nodes & - & 47K & - & $\sim$50h \\
DeepEyesV2~\cite{hong2025deepeyesv2} & 4 Nodes & 4 Nodes & 268K & 83K &$\sim$28h & $\sim$54h \\
Thyme~\cite{zhang2025thyme} & 1 Node & 1 Node & 500K & 55K &$\sim$25h & $\sim$150h \\
\rowcolor{mycustomcolor}
BEE-7B & \textbf{1 Node} & \textbf{1 Node} & 320K & 55K & $\sim$4h & \textbf{$\sim$49h} \\ 
\bottomrule
\end{tabular}
}
\end{table}



%% file: table_figs/tab5_system_metrics.tex
\begin{table}[t]
\centering
\caption{{\bf System-level metrics}. Note: Cost is \$1.5/hour node price.}
\footnotesize
\scalebox{0.9}{
\begin{tabular}{l
>{\columncolor[HTML]{DDEEFF}}c cc}
\hline
Metrics                                                         & \multicolumn{1}{l}{\cellcolor[HTML]{DDEEFF}BEE-7B} & \multicolumn{1}{l}{DeepEyes~\cite{zheng2025deepeyes}} & \multicolumn{1}{l}{DeepEyesV2~\cite{hong2025deepeyesv2}} \\ \hline
Memory Usage (GB)                                               & 112.02                                             & 112.90                          & 120.40                            \\
\begin{tabular}[c]{@{}l@{}}Throughput\\ (sample/s)\end{tabular} & 1.61                                               & 0.07                            & 0.33                              \\
\begin{tabular}[c]{@{}l@{}}Throughput\\ (Token/s)\end{tabular}  & 80.64                                             & 186.30                          & 0.90                              \\
Avg. Latency (s)                                                & 1.37                                               & 8.82                            & 3.07                              \\
Latency Variance                                                & 1.04                                               & 18.85                           & 3.75                              \\
Cost Per 1K sample (\$)                                         & 0.72                                               & 6.40                            & 1.28                              \\ \hline
\end{tabular}}
\label{tab:tab5_cost}
\end{table}

%% file: table_figs/tab6_num_part.tex
\begin{figure*}[t]
    \centering
    \begin{minipage}{0.55\textwidth}
        \centering
        \captionof{table}{{\bf Performance comparison of different tool slot formats.} For all open-source models, the best performance for each metric is {\bf bolded}, and the second best is {\ul underlined}.}
        \label{tab:tab6_tool}
        \resizebox{\linewidth}{!}{%
        \renewcommand{\arraystretch}{1.1}
        \setlength{\tabcolsep}{6pt} 
        \begin{tabular}{lcccc}
            \toprule
                                          & \multicolumn{1}{l}{}                               & \multicolumn{3}{l}{Real-World Understanding}                                   \\ \cline{3-5} 
            \multirow{-2}{*}{Type}        & \multicolumn{1}{l}{\multirow{-2}{*}{Size(SFT+RL)}} & \multicolumn{1}{l}{HR-4K} & \multicolumn{1}{l}{HR-8K} & \multicolumn{1}{l}{V*} \\ \midrule
            \rowcolor[HTML]{F4F4F4} 
            Qwen2.5-VL-7B~\cite{bai2025qwen2}                 & -                                                  & 68.8                      & 65.3                      & 76.4                   \\
            replace the original image & 320K+55K                                           & 71.3                      & 66.8                      & 80.6                   \\
            replace the unk token         & 320K+55K                                           & 73.5                      & 68.3                      & 79.1                   \\
            pure text-based reasoning               & 320K+55K                                           & {\ul 74.5}                & {\ul 72.8}                & {\ul 82.7}             \\
            \rowcolor[HTML]{DDEEFF} 
            BEE-7B                & 320K+55K                                           & \textbf{75.4}             & \textbf{73.8}             & \textbf{83.8}          \\ \bottomrule
            \end{tabular}
        }
    \end{minipage}
    \hfill %
    \begin{minipage}{0.42\textwidth}
        \centering
        \includegraphics[width=\linewidth]{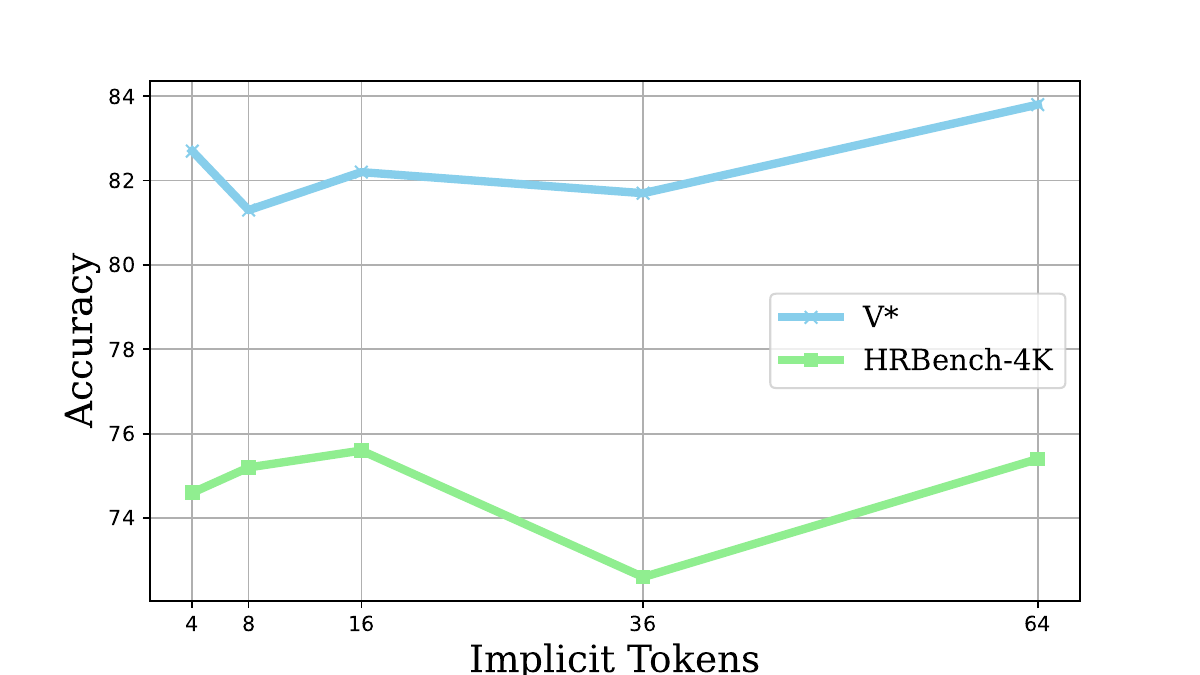}
        \captionof{figure}{\textbf{Effect of token length}.}
        \label{fig:fig6_token_length}
    \end{minipage}
\end{figure*}

%% file: table_figs/tab2_split_dataset_mme_lite.tex
\begin{table*}[t]
\centering
\caption{{\bf Performance comparison on the MME-Real-Lite.} BEE shows larger improvements on more challenging perception and reasoning tasks, as well as monitoring and autonomous driving tasks, where the baseline performs relatively poorly.
}
\label{tab:tab2_splitdataset_lite}
\scalebox{0.99}{

\begin{tabular}{lcccccc}
\hline
\multicolumn{7}{c}{\cellcolor[HTML]{D6ECFA}Perception}                                                                                \\
Model                & Monitoring      & Autonomous Driving & OCR             & Diagram and Table & Remote Sensing  & Overall         \\
Qwen2.5-VL-7B~\cite{bai2025qwen2}        & 29.1            & 31.7               & 90.4            & 85.0              & 44.0            & 49.7            \\
BEE-7B               & 41.0            & 39.7               & 90.0            & 88.0              & 52.0            & 56.5            \\
\textit{Improvement} & \textit{40.9\%} & \textit{25.2\%}    & \textit{-0.4\%} & \textit{3.5\%}    & \textit{18.2\%} & \textit{13.7\%} \\ \hline
\multicolumn{7}{c}{\cellcolor[HTML]{F0E5DE}Reasoning}                                                                                 \\
Model                & Monitoring      & Autonomous Driving & OCR             & Diagram and Table & Remote Sensing  & Overall         \\
Qwen2.5-VL-7B~\cite{bai2025qwen2}        & 28.7            & 25.2               & 73.0            & 67.0              & -               & 37.8            \\
BEE-7B               & 47.3            & 33.0               & 76.0            & 64.0              & -               & 45.7            \\
\textit{Improvement} & \textit{64.8\%} & \textit{31.0\%}    & \textit{4.1\%}  & \textit{-4.5\%}   & \textit{-}      & \textit{20.9\%} \\ \hline
\end{tabular}

}
\end{table*}



%% file: table_figs/tab12_ntg.tex
\begin{table}[t]
\centering
\caption{
        {\bf NTG metric comparison.} The BEE-7B model achieves the maximum actual marginal benefit of tool invocation.
}
\label{tab:tab12_ntg}
\scalebox{0.9}{

\begin{tabular}{lcccc}
\hline
Model & Thyme~\cite{zhang2025thyme}                 & DeepEyes~\cite{zheng2025deepeyes}              & BEE-Static & BEE-7B \\ \hline
NTG   & \multicolumn{1}{c}{39.6} & \multicolumn{1}{c}{2.3} &   6.2         &  66.0   \\ \hline
\end{tabular}
}
\end{table}

%% file: table_figs/fig5_reward_curve.tex
\begin{figure*}[ht]
    \centering
    \includegraphics[width=1.0\linewidth]{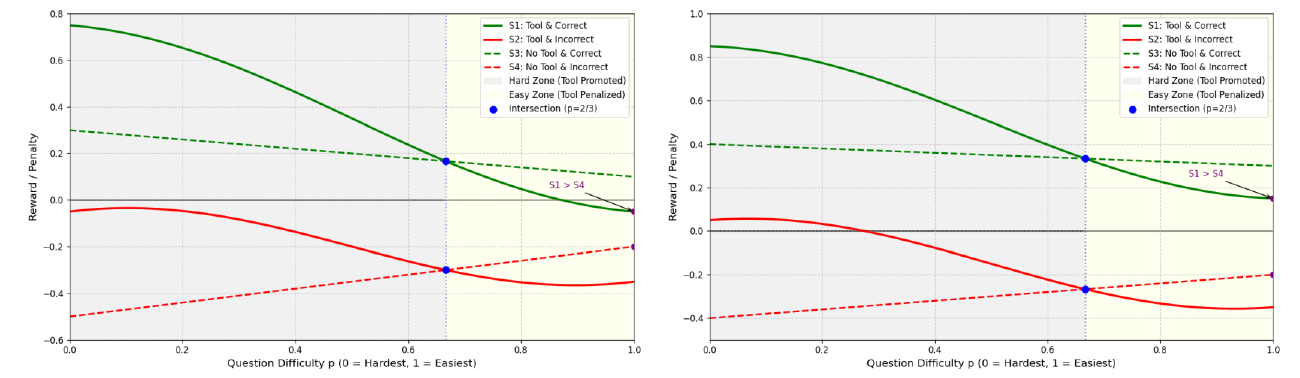}
    \caption{
        {\bf Visualization of the MC-Reward curve.} MC-Reward is defined over two continuous regions. In the easy-problem region, using tools always yields a lower reward than direct reasoning. In the difficult-problem region, using tools consistently yields a higher reward than direct reasoning.
    }
    \label{fig:fig5_reward}
\end{figure*}

%% file: table_figs/fig6_tool_change.tex
\begin{figure*}[ht]
    \centering
    \includegraphics[width=1.0\linewidth]{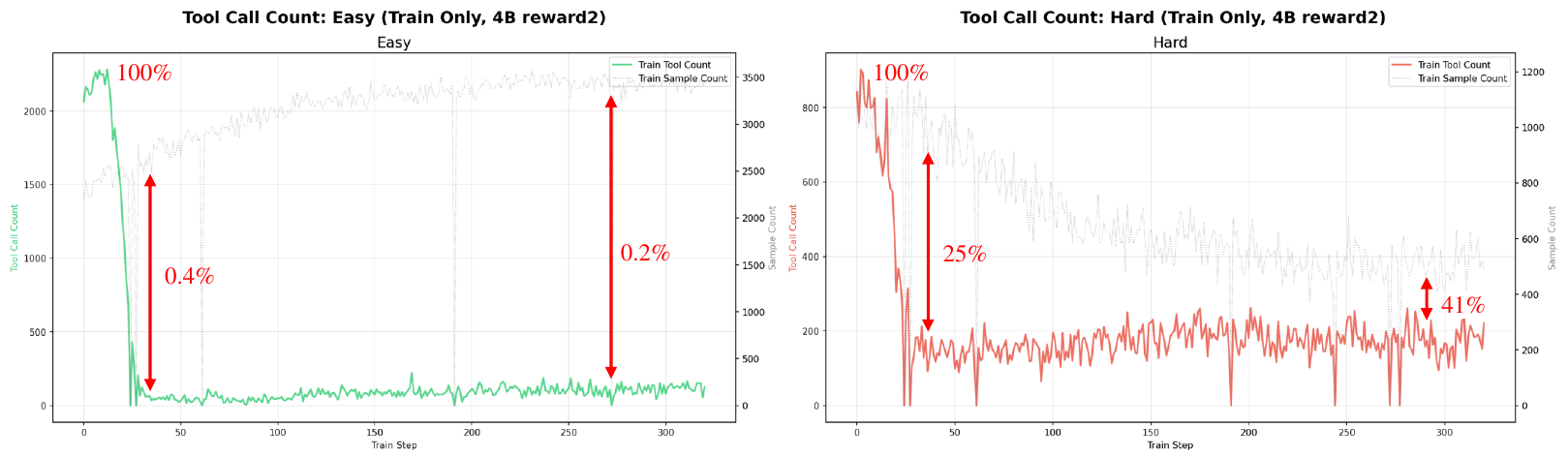}
    \caption{
        {\bf Comparison of BEE tool calling rate across tasks during training.} During the early stage of SFT training, the model makes many redundant tool calls, with the call rate approaching 100\% on both easy and difficult problems. As training progresses, the model quickly adapts its tool-use behavior and gradually reduces unnecessary calls on easy problems, eventually reaching a call rate of around 0.2\%. On difficult problems, the tool call rate first decreases to about 25\% in the middle stage and then increases as the model continues to explore strategies for solving challenging cases, finally stabilizing at approximately 41\%.
    }
    \label{fig:fig6_tool_change}
\end{figure*}

%% file: table_figs/fig11_mc_performance.tex
\begin{figure}[t]
    \centering
    \includegraphics[width=1.0\linewidth]{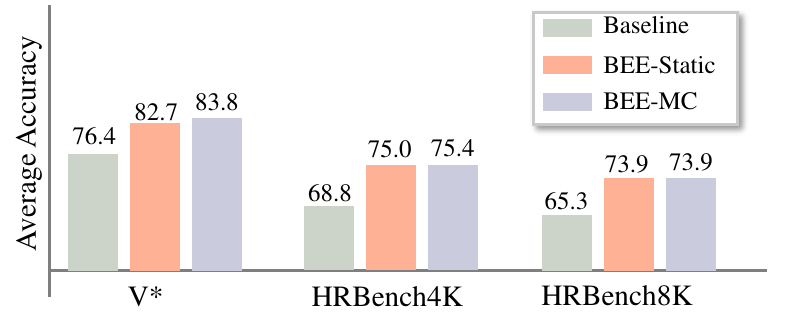}
    \caption{
        {\bf Performance comparison of different components across tasks.} The introduction of MC-Reward  enables the model to use tools more appropriately and improves its self-regulated ability.
    }
    \label{fig:fig11_mc}
\end{figure}

%% file: table_figs/tab11_mc_metrics.tex
\begin{table}[t]
\centering
\caption{{\bf System-level metrics}. Note: Cost is \$1.5/hour node price.}
\label{tab:tab11_mc_metrics}
\scalebox{0.99}{

\begin{tabular}{lcc}
\toprule
Metrics                                                         & \multicolumn{1}{l}{BEE-7B-Static} & \multicolumn{1}{l}{BEE-7B-MC} \\ \midrule
Memory Usage (GB)                                               & 112.03                            & 112.02                        \\
\begin{tabular}[c]{@{}l@{}}Throughput\\ (sample/s)\end{tabular} & 0.88                              & 1.61                          \\
\begin{tabular}[c]{@{}l@{}}Throughput\\ (Token/s)\end{tabular}  & 181.62                            & 80.64                         \\
Avg. Latency (s)                                                & 2.53                              & 1.37                          \\
Latency Variance                                                & 3.1                               & 1.04                          \\
Cost Per 1K sample (\$)                                         & 1.05                              & 0.72                          \\ \bottomrule
\end{tabular}
}
\end{table}

%% file: table_figs/fig12_mask.tex
\begin{figure}[ht]
    \centering
    \includegraphics[width=1.0\linewidth]{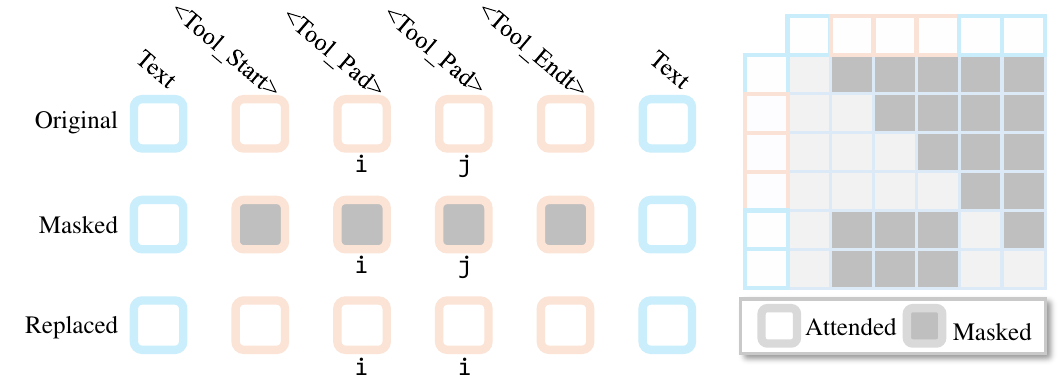}
    \caption{
        {\bf Behavioral analysis of tool slot during the testing phase.} \textit{Masked:} During inference, the generated tool tokens are masked out, ensuring that the generation of subsequent tokens does not depend on them.
        \textit{Replaced:} Different tool tokens are replaced with a single, uniform tool token, making all tool tokens identical (\eg, \textless Tool\_Pad\_i\textgreater{} or \textless Tool\_Pad\_j\textgreater{}).
    }
    \label{fig:fig12_mask}
\end{figure}

%% file: table_figs/tab9_replace.tex
\begin{table}[t]
\centering
\caption{
        {\bf Performance comparison of replacing different implicit tokens.}  For all open-source models, the best performance for each metric is {\bf bolded}, and the second best is {\ul underlined}.
}
\label{tab:tab9_replace}
\scalebox{0.75}{

\begin{tabular}{lcccccc}
\hline
\multirow{2}{*}{Type} & \multicolumn{3}{c}{V*}                                                                    & \multicolumn{3}{l}{HRBench-4K}                                                  \\ \cline{2-7} 
                      & \multicolumn{1}{l}{Attribute} & \multicolumn{1}{l}{Spatial} & \multicolumn{1}{l}{Overall} & \multicolumn{1}{l}{FSP} & \multicolumn{1}{l}{FCP} & \multicolumn{1}{l}{Overall} \\ \hline
Qwen2.5-VL-7B~\cite{bai2025qwen2}         & 78.2                          & 73.6                        & 76.4                        & 85.2                    & 52.2                    & 68.8                        \\
\rowcolor[HTML]{DDEEFF}  BEE-7B                & \textbf{80.3}                 & \textbf{86.1}               & \textbf{83.8}               & \textbf{92.5}           & \textbf{58.3}           & \textbf{75.4}               \\
Replace-1             & {\ul 80.0}                    & 82.8                        & 81.1                        & {\ul 86.5}              & 52.7                    & {\ul 69.6}                  \\
Replace-2             & 79.1                          & {\ul 84.5}                  & {\ul 81.2}                  & 86.0                    & {\ul 53.2}              & 69.6                        \\ \hline
\end{tabular}
}
\end{table}

%% file: table_figs/tab10_decouple.tex
\begin{table}[t]
\centering
\caption{
        {\bf Performance comparison of decoupling perception and reasoning.}  
}
\label{tab:tab10_decouple}
\scalebox{0.75}{

\begin{tabular}{lcccccc}
\toprule
\multirow{2}{*}{Method} & \multicolumn{3}{c}{V*}        & \multicolumn{3}{c}{HRBench-4K} \\ \cline{2-7} 
                        & Attribute & Spatial & Overall & FSP     & FCP     & Overall    \\ \midrule
Qwen2.5-VL-7B~\cite{bai2025qwen2}           & 78.2      & 73.6    & 76.4    & 85.2    & 52.2    & 68.8       \\
BEE-7B                  & 80.3      & 86.1    & 83.8    & 92.5    & 58.3    & 75.4       \\
BEE-7B-adapter          & 81.5      & 86.0    & 84.2    & 92.7    & 59.7    & 76.2       \\ \bottomrule
\end{tabular}
}
\end{table}

%% file: sec/5_conclusions.tex
\section{Conclusion}
\label{sec:conclusion}
In this work, we propose BEE, a novel self-regulated reasoning framework for implicit visual tools that establishes a cognitive boundary, allowing the model to adaptively balance internal parametric knowledge and implicit tool invocation.
By integrating formalized chain-of-thought supervised fine-tuning and self-regulated reward-driven alignment strategy, BEE enhances the perception and reasoning capabilities of multimodal large language models while substantially improving inference efficiency. Extensive experiments demonstrate that our method improves overall performance and successfully achieves adaptive tool routing. We believe that our BEE can provide meaningful references for designing low-latency, self-regulated MLLMs.

%% file: sec/6_appendix.tex

\clearpage
\appendices
\onecolumn
\input{appendix/table_of_content}
\twocolumn

\clearpage
\section{Method Design.}
\label{app:method_design}

\subsection{Theoretical Analysis of MC-Reward}
\label{app:sec_analysis_mcreward}
Based on the value ranges of the accuracy, format, and consistency rewards and the computation in Equation 1, the range of MC-reward under the four scenarios is

\begin{equation}
\label{eq:eq_11}
\left.
\begin{aligned}
 \alpha_1 + \beta_1p + \gamma \cos (\pi p) + u \leq 1, \\
 \alpha_1 + \beta_1p \leq 1, \\
-1 \le \alpha_2 + \beta_2p + \gamma \cos (\pi p) + u, \\
-1 \le \alpha_2 + \beta_2p,
\end{aligned}
\right\}
\quad \text{for } p \in [0, 1].
\end{equation}

\begin{equation}
\label{eq:eq_12}
\gamma \cos (\pi p) + u = 0, \quad \text{at } p = \frac{2}{3}.
\end{equation}

\begin{equation}
\label{eq:eq_13}
\left.
\begin{aligned}
\alpha_1 + \beta_1p + \gamma \cos (\pi p) + u > \alpha_1 + \beta_1p, \\
\alpha_2 + \beta_2p + \gamma \cos (\pi p) + u > \alpha_2 + \beta_2p,
\end{aligned}
\right\}
\quad \text{for } p \in [0, \frac{2}{3}).
\end{equation}

\begin{equation}
\label{eq:eq_14}
\left.
\begin{aligned}
\alpha_1 + \beta_1p + \gamma \cos (\pi p) + u < \alpha_1 + \beta_1p, \\
\alpha_2 + \beta_2p + \gamma \cos (\pi p) + u < \alpha_2 + \beta_2p,
\end{aligned}
\right\}
\quad \text{for } p \in (\frac{2}{3}, 1].
\end{equation}

By integrating the global boundary inequality constraints in Equation~\ref{eq:eq_11}, the critical point equality in Equation~\ref{eq:eq_12}, and the action-advantage interval inequalities in Equations~\ref{eq:eq_13} and \ref{eq:eq_14}, we perform a rigorous mathematical analysis of the admissible parameter space for the internal system parameters $\alpha_1, \beta_1, \alpha_2, \beta_2, \gamma, u$.

1. Analytical relationship and sign determination for $u$ and $\gamma$

Following the seamless junction design at the cognitive threshold $p = \frac{2}{3}$ and Equation~\ref{eq:eq_12}:

\begin{equation}
\gamma \cos\left(\pi \cdot \frac{2}{3}\right) + u = 0.
\end{equation}

Substituting $\cos\left(\frac{2\pi}{3}\right) = -0.5$ yields:

\begin{equation}
u = 0.5 \gamma.
\end{equation}

Inserting $u = 0.5\gamma$ into Equation~\ref{eq:eq_13} eliminates the common linear term, giving the condition for the difficult exploration region $p \in [0, \frac{2}{3})$:
\begin{equation}
\gamma \left( \cos(\pi p) + 0.5 \right) > 0.
\end{equation}
Since $\cos(\pi p) > -0.5$ for all $p \in [0, \frac{2}{3})$, the inequality strictly holds if and only if $\gamma > 0$. This condition also ensures that Equation~\ref{eq:eq_14} is automatically satisfied.

To simplify the nonlinear boundary, we set $\gamma = 0.3$, resulting in $u = 0.15$. The nonlinear patch term then becomes:
\begin{equation}
P(p) = 0.3 \cos(\pi p) + 0.15.
\end{equation}
2. Derivative and extremal properties of the core function

The composite reward function has the unified algebraic form $H(p) = \alpha + \beta p + 0.3 \cos(\pi p) + 0.15$. Its derivative with respect to $p$ is:
\begin{equation}
H'(p) = \beta - 0.3 \pi \sin(\pi p).
\end{equation}
Based on the critical slope $k = 0.3 \pi \approx 0.942$, $H(p)$ exhibits three monotonicity patterns:

Pattern A ($\beta \le 0$) is strictly decreasing, with maximum at $p = 0$ and minimum at $p = 1$.

Pattern B ($\beta \ge 0.3 \pi$) is strictly increasing, with maximum at $p = 1$ and minimum at $p = 0$.

Pattern C ($0 < \beta < 0.3 \pi$) has internal extrema. Let $\sin(\theta) = \frac{\beta}{0.3\pi}$; the local maximum occurs at $p_1 = \frac{\theta}{\pi}$, and the local minimum at $p_2 = 1 - \frac{\theta}{\pi}$.

3. Upper bounds for positive-reward region parameters $(\alpha_1, \beta_1)$

According to Equation~\ref{eq:eq_11}, it suffices to ensure that the maximum does not exceed the upper limit $1$:

$\alpha_1 + \beta_1 p \le 1$ and $H_1(p) \le 1$ for all $p \in [0, 1]$.

Case 3.1: $\beta_1 \le 0$ (strictly decreasing)

The global maximum is $H_1(0)$. Enforcing the endpoint:
\begin{equation}
H_1(0) = \alpha_1 + 0.45 \le 1 \implies \alpha_1 \le 0.55.
\end{equation}
This automatically satisfies the basic constraint $\alpha_1 \le 1$. No lower bound check is required. The admissible region is $\alpha_1 \le 0.55$.

Case 3.2: $\beta_1 \ge 0.3 \pi$ (strictly increasing)

The global maximum is $H_1(1)$. Enforcing:
\begin{equation}
H_1(1) = \alpha_1 + \beta_1 - 0.15 \le 1 \implies \alpha_1 + \beta_1 \le 1.15.
\end{equation}
Taking the intersection with the basic constraint $\alpha_1 + \beta_1 \le 1$, the admissible region is $\alpha_1 + \beta_1 \le 1$.

Case 3.3: $0 < \beta_1 < 0.3 \pi$ (internal maximum)

The internal peak must satisfy:
\begin{equation}
\alpha_1 + \beta_1 \left(\frac{\theta}{\pi}\right) + 0.3 \sqrt{1 - \left(\frac{\beta_1}{0.3\pi}\right)^2} + 0.15 \le 1.
\end{equation}
Along with the endpoint constraint $\alpha_1 + \beta_1 \le 1$.

4. Lower bounds for negative-penalty region parameters $(\alpha_2, \beta_2)$

According to Equation~\ref{eq:eq_11}, it suffices to ensure that the minimum does not fall below $-1$:

$\alpha_2 + \beta_2 p \ge -1$ and $H_2(p) \ge -1$ for all $p \in [0, 1]$.

Case 4.1: $\beta_2 \le 0$ (strictly decreasing)

The global minimum is $H_2(1)$. Enforcing the endpoint:
\begin{equation}
H_2(1) = \alpha_2 + \beta_2 - 0.15 \ge -1 \implies \alpha_2 + \beta_2 \ge -0.85.
\end{equation}
Intersection with the basic lower bound $\alpha_2 + \beta_2 \ge -1$ gives the admissible region $\alpha_2 + \beta_2 \ge -0.85$.

Case 4.2: $\beta_2 \ge 0.3 \pi$ (strictly increasing)

The global minimum is $H_2(0)$. Enforcing:
\begin{equation}
H_2(0) = \alpha_2 + 0.45 \ge -1 \implies \alpha_2 \ge -1.45.
\end{equation}
Intersection with the basic lower bound $\alpha_2 \ge -1$ gives $\alpha_2 \ge -1$.

Case 4.3: $0 < \beta_2 < 0.3 \pi$ (internal minimum)

The internal trough must satisfy:
\begin{equation}
\alpha_2 + \beta_2 \left(1 - \frac{\theta}{\pi}\right) - 0.3 \sqrt{1 - \left(\frac{\beta_2}{0.3\pi}\right)^2} + 0.15 \ge -1.
\end{equation}
Along with the endpoint constraint $\alpha_2 \ge -1$.

Based on these conditions, we select $\alpha_1 = 0.3$, $\beta_1 = -0.2$, $\alpha_2 = -0.5$, and $\beta_2 = 0.3$. Alternative values are also feasible, and as long as the MC-reward exhibits a similar trend, the trained model converges to comparable outcomes.

\noindent
\textbf{Why are $R_{base}$ and $\Omega_{buffer}$ selected in Equation 7?}
In the initial stage of our exploration, we investigate a discrete reward mechanism that assigns fixed constant rewards to different behavioral phases. However, experimental results demonstrate that under this configuration, the tool utilization capability of the model remains stagnant and fluctuates significantly during training, occasionally even degenerating to zero calls. This sub-optimal performance primarily stems from the sparse feedback signals inherent in discrete rewards, which induces a reward cliff effect and consequently deprives the model of clear gradient guidance. To alleviate this issue, we reformulate the reward mechanism into a continuous functional form. Specifically, we implement a linear continuous function for the baseline reward, complemented by a cosine-based continuous reward specifically designed for tool invocation. This dual formulation provides a continuous optimization landscape for the model. More importantly, the cosine model maintains excellent smoothness at the critical boundaries of 0 and 1, which successfully eliminates gradient discontinuities, prevents training variance, and ultimately enables the model to stably acquire tool invocation capabilities.

\subsection{Algorithm of BEE.}
\label{app:appendix_a2}
\input{table_figs/Algorighm}
The algorithmic procedure of the BEE method is illustrated in Algorithm~\ref{alg:training_process}.
The BEE algorithm has two main stages. In the first stage, the BEE model generates implicit tools, and we fine-tune it on the BEE-SFT-320K dataset to establish an initial ability for this task. In the second stage, the model learns to control its own behavior and adjust tool usage based on its capabilities.

\subsection{Semantic Tool Tokenizer}
\input{table_figs/fig14_tokenizer}
\label{app:appendix_a3}

\begin{itemize}
    \item \textbf{Data.} We trained the tokenizer using the complete SFT dataset alongside the pre-training data from the LLaVA~\cite{liu2023visual} paper.

    \item \textbf{Training.}
    Inspired by autoregressive image generation paradigms~\cite{vqgan,vqvae} and unified vision-language models (VLMs)~\cite{illume+, semhitok, liquid}, we develop a semantic tool tokenizer that discretizes tool targets into code tokens, thereby enabling unified autoregressive reasoning and planning.
    
    Specifically, given a tool representation ${B} \in \mathbb{R}^{H \times W \times C}$, we first extract semantic features using a frozen encoder $E_{\theta}(\cdot)$: ${F} = E_{\theta}({B})$.
    The features are projected via an MLP $g_{\phi}(\cdot)$ into a quantization space: ${H} = g_{\phi}({F})$.
    We then apply vector quantization with a learnable codebook ${Z} = \{ {z}_k \}_{k=1}^K$, where feature is mapped to its nearest code:
    \begin{equation}
    \mathbf{Z_q}, {I_q} = \mathop{\arg\min}\limits_{k \in \{1,\dots,K\}} \| {H} - \mathbf{Z[k]} \|_2^2
    \end{equation}
    Where $K$ is the codebook size, $\mathbf{Z_q}$ is the quantized feature, and ${I_q}$ is the quantized index.
    The quantized features are then decoded by $D_{\psi}(\cdot)$ to reconstruct the semantic representation: $\hat{F} = D_{\psi}({Z}_q)$.
    The tokenizer is trained with a standard VQ objective:
    \begin{equation}
        \mathcal{L} = \mathcal{L}_{sem} (F,\hat{F}) + \| \text{sg}[H] - Z_q \|_2^2 + \beta \| \mathbf{H} - \text{sg}[\mathbf{Z}_q] \|_2^2
    \end{equation}
    where $\mathcal{L}_{sem}$ represents the mean squared error loss, $\text{sg}[\cdot]$ denotes the stop-gradient operator.
    
    The semantic tool tokenizer converts tool representations into discrete codes, thereby enabling seamless integration with existing vlm frameworks.
    
    As shown in Fig.~\ref{fig:fig14_tokenizer}, we demonstrate how to train a semantic tool tokenizer and how to use it to convert tool representations into discrete special tokens compatible with the autoregressive VLM interface.
\end{itemize}







\clearpage
\section{Experiment Details}
\label{app:experiment_details}

\subsection{Curating {\MethodName} Training Data.}
\label{app:detail_bee_data}
\input{table_figs/Algorithm_data}
\input{table_figs/tab7_data_source}
\input{table_figs/Algorithm_system}
BEE training data consists of two parts.: BEE-SFT-320K and BEE-RL-55K. 
\begin{itemize}
    \item \textbf{SFT dataset.} For the supervised fine-tuning phase, we construct the BEE-SFT dataset, comprising approximately 320,000 samples integrated from Vision-CoT~\cite{shao2024visual}, ReFocus~\cite{fu2025refocus}, Thyme~\cite{zhang2025thyme}, and Zebra-CoT~\cite{li2025zebra}. The data format for implicit token understanding is shown as follows:
    
    \texttt{\textbf{Question}: <START\_OF\_GEN><IMG\_OF\_0920>\\<IMG\_OF\_0646><IMG\_OF\_0018><IMG\_OF\_0990>\\<END\_OF\_GEN>Share a concise interpretation of the image provided.}

    \texttt{\textbf{Answer}: radar image for severe weather, as shown by the national weather service.}
    
    \item \textbf{RL dataset.} For the subsequent reinforcement learning stage, we compile the BEE-RL dataset, consisting of roughly 55,000 samples derived from Thyme~\cite{zhang2025thyme}.
\end{itemize}
Table~\ref{tab:tab_aba_data} presents the tasks included in each dataset, the types of visual operations, and their corresponding proportions.
The algorithmic procedure of the BEE dataset is illustrated in Algorithm~\ref{alg:data_process}.
Please note that our SFT and RL datasets do not contain any data from additional scenarios such as logical reasoning, autonomous driving, or medical applications, and these can therefore serve as out-of-distribution (OOD) test sets.

\subsection{Hyperparameters}
\label{app:hyper}
\input{table_figs/tab_supp0_sft_hyper}
\input{table_figs/tab_supp1_rl_hyper}
\begin{itemize}
    \item \textbf{SFT Stage.} During the SFT stage, we use the LLaMA-Factory framework\footnote{\url{https://github.com/hiyouga/LlamaFactory}} and train our models based on Qwen2.5-VL-7B-Instruct and Qwen3-VL-4B/8B-Instruct. The detailed hyperparameter settings are summarized in Table~\ref{tab:tab_supp0_sft}. All variant models in this paper use the same training framework and hyperparameter settings as BEE during the SFT stage.
    \item \textbf{RL Stage.} During the RL stage, we use the Easy-R1 framework\footnote{\url{https://github.com/hiyouga/easyr1}} and keep the backbone models the same as those used in the SFT stage. The detailed hyperparameter settings are summarized in Table~\ref{tab:tab_supp1_rl}. All variant models also use the same training framework and hyperparameter settings as BEE during the RL stage.
\end{itemize}

\subsection{Alternative Forms of Implicit Tools}
\label{app:alternative_form}
\input{table_figs/fig10_implicit_tool_compare}
Figure~\ref{fig:fig10_compare} shows examples of different alternatives to implicit tool usage. Figure~\ref{fig:fig10_compare} (a) corresponds to the BEE-SFT data format, Figure~\ref{fig:fig10_compare} (b) replaces the implicit tool target with the original image, Figure~\ref{fig:fig10_compare} (c) replaces the implicit tool token with a meaningless \texttt{<unk>} token, and Figure~\ref{fig:fig10_compare} (d) removes the implicit tool entirely.

\subsection{Efficiency comparisons setting}
\label{app:effi}
\input{table_figs/tab_supp2_efficiency}
The parameters for comparing the efficiency of each model are in Table~\ref{tab:tab_supp2_efficiency}.


\subsection{SFT Training Examples.}
\label{app:sft_rl_example}
\input{table_figs/fig9_sft_example}
Figure~\ref{fig:fig9_sft} shows examples from our SFT dataset, including cases that require implicit tool usage and cases that can be solved without it.

\subsection{Pseudocode for system-level metrics.}
\label{app:metrics}
Because different methods are built on different codebases, and some tool-based methods require additional processing with external tools, we add the computation procedure described in Algorithm~\ref{alg:metrics} to the implementation of each method. This allows us to measure throughput, latency, and cost consistently across all compared models.


\clearpage
\section{Additional Results.}
\label{app:additional_results}
\input{table_figs/tab8_all_results}
\input{table_figs/tab2_splitdataset}
\input{table_figs/fig13_decouple_tsne}
\input{table_figs/Algorithm_zwz}
\input{table_figs/Algorithm_bee}


\subsection{Substantial Gain over Base Models.}
\label{app:add_2}
Table~\ref{tab:tab8_main} presents a comparison of BEE-7B with the baseline model Qwen2-VL-7B, the larger model Qwen2.5-VL-32B, and the recent method ZwZ-7B. We also report the performance of the closed-source model Gemini-2.5 Pro for reference. The evaluation protocol used in the official ZwZ repository differs from ours. For a fair comparison, we evaluate the publicly released ZwZ model within our unified evaluation framework (\eg, VLMEValKit\footnote{\url{https://github.com/open-compass/vlmevalkit}}) and use the same system prompt as in the original implementation. In addition, all model outputs are evaluated by Qwen2.5-VL-235B, which is also the judge model used for BEE. Consequently, some results differ from those reported in the original ZwZ paper. A comparison of the two evaluation protocols is shown in Algorithm~\ref{alg:evaluation_pipeline} and Algorithm~\ref{alg:hrbench4k_matching}, respectively.
\begin{itemize}
    \item \textbf{Compared with the baseline models.} BEE consistently performs better on \textbf{perception tasks}. The improvement is particularly evident on MME-Real-Lite, where BEE achieves a score of 52.3, surpassing the 44.1 obtained by Qwen2.5-VL-7B by more than 18.6\%. On \textbf{challenging reasoning benchmarks}, BEE is not trained on datasets that are similar to these tasks, which makes them out-of-distribution evaluations. Even in this setting, BEE achieves 43.0 on LogicVista, outperforming Qwen2.5-VL-7B, which scores 39.8. However, BEE still shows relatively weak performance on some benchmarks, such as VisuLogic and MathVerse.
    On \textbf{more general benchmarks}, BEE brings limited gains on tasks where the baseline performance is already strong, including SEEDBench and AI2D. In contrast, it consistently outperforms Qwen2.5-VL-7B on benchmarks with relatively lower baseline scores, such as CMMMU\_val, MME, and OCRBench. These results suggest that the performance gains of BEE do not primarily come from fine-tuning on in-domain data, but instead reflect improvements in more general capabilities. BEE also further reduces hallucinations on the POPE benchmark.

    \item \textbf{Compared with ZwZ-7B.} BEE outperforms it on almost all perception benchmarks except MME-Real. This result suggests that the implicit tool mechanism in BEE effectively benefits multimodal reasoning. On complex reasoning tasks, ZwZ exhibits a noticeable drop in performance on VisuLogic and MathVerse. On more general benchmarks, BEE and ZwZ show comparable performance.
\end{itemize}

\subsection{Analysis of Specialized Benchmarks.}
\label{app:add_3}
Taking MME-RealWorld as an example, it contains a large number of high-resolution perception tasks in real-world scenarios. Table~\ref{tab:tab2_splitdataset_2} reports the performance of BEE and the baseline model across different task categories. We find that BEE brings only limited gains on tasks where the baseline model already performs strongly, such as OCR, Diagram, and Table, where the accuracy exceeds 60\% and even approaches 85\% on some tasks. BEE even suffers a drop of around 5 percentage points on the reasoning subsets of Diagram and Table. In contrast, on more challenging tasks where Qwen2.5-VL-7B shows weaker perception ability, such as monitoring and autonomous driving, BEE improves performance by more than 15\% on both perception and reasoning tasks, with larger gains on reasoning tasks.


\input{table_figs/tab13_tau}
\subsection{Ablation study on $\tau$.}
\label{app:add_tau}
Table~\ref{tab:tab13_tau} shows the effect of the threshold $\tau$ across different tasks and models. The results indicate that performance is stable across models, with the best results obtained when $\tau$ is between $1/2$ and $3/4$. The same trend is observed across different tasks.

\input{table_figs/tab14_part}
\subsection{Ablation study on each component.}
\label{app:add_part}
Table~\ref{tab:tab14_part} presents the results of the two stages of model training.


\subsection{Analysis of decoupled training.}
\label{app:add_6}
We utilize t-SNE to project pure textual tokens and visual tokens into a shared low-dimensional space (e.g., dimension 2, perplexity 30).


\subsection{Training curves}
\label{app:curves}
\input{table_figs/fig4_curve}
We further provide an analysis of the training dynamics of BEE-7B, including learning curves of different reward signals, changes in tool usage and accuracy across different difficulty levels, the overall tool usage rate during training, and the evolution of the number of problems at each difficulty level.
\begin{itemize}
    \item \textbf{Training reward curves.} Figure~\ref{fig:fig4_curve} presents the curves of the accuracy reward, consistency reward, and tool reward. All rewards increase steadily during training, indicating that BEE gradually improves and eventually reaches an accuracy of around 80\% on the training set.
    \item \textbf{Tool usage rate \& accuracy curves.} Figure~\ref{fig:fig7_aba1} shows the changes in tool usage rate and accuracy across different difficulty levels. For easy problems, accuracy continuously improves, while the tool usage rate gradually decreases and eventually approaches zero, suggesting that the model can solve these problems without external tools. For hard problems, some relatively easier instances are gradually solved during training, leaving a subset of more difficult problems that exceed the model's capability. As a result, the number of hard problems decreases over time, which also leads to a drop in the average accuracy of this subset. This behavior is consistent with the trend shown in Figure~\ref{fig:fig7_aba2}. Despite this change in the problem distribution, the tool usage rate on hard problems eventually stabilizes at around 40\%. The tool usage rate on medium-difficulty problems follows a similar trend to that of hard problems.
    \item \textbf{Difficulty curves.} Figure~\ref{fig:fig7_aba2} illustrates the changes in the number of problems at different difficulty levels during training. The number of hard and medium-difficulty problems gradually decreases, while the number of easy problems steadily increases. This suggests that BEE progressively solves more challenging problems, resulting in a shift of the overall distribution toward easier cases.
\end{itemize}

\input{table_figs/fig7_aba1_curve}
\input{table_figs/fig7_aba2_curve}


\clearpage
\section{Discussion}
\label{app:discussion}


\subsection{Limitations}
\label{app:dis_2}
Our error case analysis reveals two main limitations of BEE. 
The first is its limited ability to handle complex logical reasoning tasks. In some cases, the model skips detailed reasoning steps and produces answers without sufficient justification. Such problems are also difficult for humans without prior exposure to similar tasks. We think this limitation mainly stems from the lack of complex reasoning examples in the current training data. In future work, the data construction strategy introduced in Section 3.2 can be extended by using recent foundation models such as Gemini3-Pro to generate high-quality reasoning chains, which may further improve the logical reasoning ability of BEE.

The second limitation arises when multiple similar objects appear in the same image. BEE sometimes fails to identify all relevant targets. In our autonomous driving case studies, we observe that the model occasionally misses some vehicles. This issue is likely caused by the lack of autonomous driving scenarios and dense multi-object perception examples in the current training data. Incorporating more such scenarios into the training set is a natural direction for improving BEE's ability to perceive and recognize a large number of objects.

\clearpage
\section{Case Studies.}
\label{app:case_studies}
\subsection{Success Cases.}
\label{app:sucess}
Figures~\ref{fig:fig8_o1} - ~\ref{fig:fig8_o11} showcase representative outputs of BEE on a broad range of tasks, including real-world image understanding (\cf, Figure~\ref{fig:fig8_o4}, ~\ref{fig:fig8_o5}, ~\ref{fig:fig8_o6}, ~\ref{fig:fig8_o7}), mathematical (\cf, Figure~\ref{fig:fig8_o1}) and general reasoning~\ref{fig:fig8_o2}, logical reasoning (\cf, Figure~\ref{fig:fig8_o13}), autonomous driving (\cf, Figure~\ref{fig:fig8_o11}), and medical analysis (\cf, Figure~\ref{fig:fig8_o8}, \ref{fig:fig8_o9}). The results reported here are obtained based on the BEE-4B model.
\begin{itemize}
    \item \textbf{Real-world image understanding.}
    Figures~\ref{fig:fig8_o4}, ~\ref{fig:fig8_o5}, ~\ref{fig:fig8_o6}, ~\ref{fig:fig8_o7} present evaluations on scenarios involving small objects, rotated objects, and blurry objects. We observe that the BEE model can directly produce correct answers for these relatively simple cases without requiring complex reasoning. These scenarios are also straightforward for human perception, suggesting that BEE is capable of selecting an appropriate reasoning strategy based on the input difficulty. This indicates that BEE exhibits a certain level of metacognitive ability. 
    \item \textbf{Chart interpretation, Mathematical reasoning and General reasoning.}
    Figure~\ref{fig:fig8_o2} presents questions involving flowcharts. The results show that the BEE model can accurately extract information from the images and perform correct reasoning, demonstrating its robustness.
    Figure~\ref{fig:fig8_o1} presents a spatial geometry problem. The BEE model provides a reasonable solution strategy but does not include detailed computational steps.
    \item \textbf{Logical reasoning.} Figure~\ref{fig:fig8_o13} presents an example of complex logical reasoning. The model produces the correct final answer but does not provide detailed intermediate reasoning steps, although humans also tend to lack a strictly formal reasoning process for this type of problem.
    \item \textbf{Autonomous driving.} Figure~\ref{fig:fig8_o11} presents a relatively complex autonomous driving example. The BEE model adapts its tool use according to the complexity of the scene and identifies a small vehicle within it, suggesting that the BEE model has strong metacognitive capabilities.
    \item \textbf{Medical analysis.} Figures~\ref{fig:fig8_o8}, ~\ref{fig:fig8_o9} present two medical-related examples. Without relevant domain knowledge, humans also often lack reliable judgment on such problems, while the BEE model still provides accurate answers, which indicates its robustness.
\end{itemize}
The results highlight the model's ability to perform consistently across diverse multimodal scenarios. 

\subsection{Failure Cases.}
\label{app:fail}
Figures~\ref{fig:fig8_o13} - ~\ref{fig:fig8_o10} present several failure cases that reveal the current limitations of the model. 
\begin{itemize}
    \item \textbf{Missing relevant information.} In scenes containing a large number of visual elements, BEE may miss part of the relevant information (\cf, Figure~\ref{fig:fig8_o10}). We examine the outputs of the BEE model and compare them with the corresponding images. Our inspection suggests that this behavior is likely driven by the model’s overconfidence. It may also reflect the limited presence of similar multi-object autonomous driving scenarios in the BEE training data. These limitations could be mitigated by incorporating more such data into future training.
    \item \textbf{Imperfect intermediate reasoning steps.} The model also occasionally produces imperfect intermediate reasoning steps when handling complex logical reasoning problems (\cf, Figure~\ref{fig:fig8_o12}). We further compare BEE with several existing methods, including visual grounding and display-based approaches, and find that similar issues persist across these systems. While BEE demonstrates strong overall performance, these observations indicate that there is still room for improvement. Complex logical reasoning remains challenging even for humans. A possible direction is to use a stronger model such as Gemini3-Pro to generate high-quality reasoning chains, which may further improve the BEE model’s performance on complex reasoning tasks.
\end{itemize}

\input{table_figs/fig8_output1}
\input{table_figs/fig8_output2}
\input{table_figs/fig8_output4}
\input{table_figs/fig8_output5}
\input{table_figs/fig8_output6}
\input{table_figs/fig8_output7}
\input{table_figs/fig8_output8}
\input{table_figs/fig8_output9}
\input{table_figs/fig8_output11}
\input{table_figs/fig8_output13}
\input{table_figs/fig8_output12}
\input{table_figs/fig8_output10}

%% file: appendix/table_of_content.tex
\section*{Appendix Content}
\vspace{0.5em}
\noindent
\hyperref[app:method_design]{A \hspace*{0.5em} Method Design} \dotfill \pageref{app:method_design} \\[2pt]
\hyperref[app:sec_analysis_mcreward]{\hspace*{1.7em} A.1 \hspace*{0.5em} Theoretical Analysis of MC-Reward} \dotfill \pageref{app:sec_analysis_mcreward} \\[2pt]
\hyperref[app:appendix_a2]{\hspace*{1.7em} A.2 \hspace*{0.5em} Algorithm of BEE} \dotfill \pageref{app:appendix_a2} \\[2pt]
\hyperref[app:appendix_a3]{\hspace*{1.7em} A.3 \hspace*{0.5em} Semantic Tool Tokenizer} \dotfill \pageref{app:appendix_a3} \\[6pt]
\noindent
\hyperref[app:experiment_details]{B \hspace*{0.5em} Experiment Details} \dotfill \pageref{app:experiment_details} \\[2pt]
\hyperref[app:detail_bee_data]{\hspace*{1.7em} B.1 \hspace*{0.5em} Curating {\MethodName} Training Data} \dotfill \pageref{app:detail_bee_data} \\[2pt]
\hyperref[app:hyper]{\hspace*{1.7em} B.2 \hspace*{0.5em} Hyperparameters} \dotfill \pageref{app:hyper} \\[2pt]
\hyperref[app:alternative_form]{\hspace*{1.7em} B.3 \hspace*{0.5em} Alternative Forms of Implicit Tools} \dotfill \pageref{app:alternative_form} \\[2pt]
\hyperref[app:effi]{\hspace*{1.7em} B.4 \hspace*{0.5em} Efficiency comparisons setting} \dotfill \pageref{app:effi} \\[2pt]
\hyperref[app:sft_rl_example]{\hspace*{1.7em} B.5 \hspace*{0.5em} SFT Training Examples} \dotfill \pageref{app:sft_rl_example} \\[2pt]
\hyperref[app:metrics]{\hspace*{1.7em} B.6 \hspace*{0.5em} Pseudocode for System-Level Metrics} \dotfill \pageref{app:metrics} \\[6pt]
\noindent
\hyperref[app:additional_results]{C \hspace*{0.5em} Additional Results} \dotfill \pageref{app:additional_results} \\[2pt]
\hyperref[app:add_2]{\hspace*{1.7em} C.1 \hspace*{0.5em} Substantial Gain over Base Models} \dotfill \pageref{app:add_2} \\[2pt]
\hyperref[app:add_3]{\hspace*{1.7em} C.2 \hspace*{0.5em} Analysis of Specialized Benchmarks} \dotfill \pageref{app:add_3} \\[2pt]
\hyperref[app:add_tau]{\hspace*{1.7em} C.3 \hspace*{0.5em} Ablation study on $\tau$} \dotfill \pageref{app:add_tau} \\[2pt]
\hyperref[app:add_part]{\hspace*{1.7em} C.4 \hspace*{0.5em} Ablation study on each component} \dotfill \pageref{app:add_part} \\[2pt]
\hyperref[app:add_6]{\hspace*{1.7em} C.5 \hspace*{0.5em} Analysis of decoupled training} \dotfill \pageref{app:add_6} \\[2pt]
\hyperref[app:curves]{\hspace*{1.7em} C.6 \hspace*{0.5em} Training Curves} \dotfill \pageref{app:curves} \\[6pt]

\noindent
\hyperref[app:discussion]{D \hspace*{0.5em} Discussion} \dotfill \pageref{app:discussion} \\[2pt]
\hyperref[app:dis_2]{\hspace*{1.7em} D.1 \hspace*{0.5em} Limitations} \dotfill \pageref{app:dis_2} \\

%
\noindent
\hyperref[app:case_studies]{E \hspace*{0.5em} Case Studies} \dotfill \pageref{app:case_studies} \\[2pt]
\hyperref[app:sucess]{\hspace*{1.7em} E.1 \hspace*{0.5em} Success Cases} \dotfill \pageref{app:sucess} \\[2pt]
\hyperref[app:fail]{\hspace*{1.7em} E.2 \hspace*{0.5em} Failure Cases} \dotfill \pageref{app:fail} \\

%% file: table_figs/Algorighm.tex
\begin{algorithm}[t]
   \caption{{BEE} Algorithm}
   \label{alg:training_process}
    \textbf{Input:} Cold-Start Dataset $\mathcal{D}_{BEE,SFT}$, DAPO Dataset $\mathcal{D}_{BEE,RL}$, Base Model $\pi_{\theta}$
     
    \textbf{Output:} Improved Model $\pi_{\theta, \text{BEE}}$
     
   \begin{algorithmic}[1]

    \State \textcolor{gray}{$\triangleright$  \textbf{Formalized CoT Supervised Fine-tuning}} \Comment{see §3.2}
    \State $\pi_{\theta, SFT}  \leftarrow$ update $\pi_{\theta}$ with SFT on $\mathcal{D}_{BEE, SFT}$ 
    


    \State \textcolor{gray}{$\triangleright$  \textbf{Self-regulated Reward-driven Alignment}} \Comment{see §3.2} 
    \For{each batch $B$ in $D_{BEE, RL}$}
    \State $T_i, \mathcal{A}_i \leftarrow \pi_{\theta, SFT}$(B)
    \State $r_i \leftarrow \operatorname{Self\_Regulated\_Reward}(T_i, \mathcal{A}_i)$ 
    \State $\pi_{\theta, \text{RL}} \leftarrow$ DAPO($\pi_{\theta, SFT},T_i,\mathcal{A}_i,r_i$) \Comment{Using Eq 3} 
    \EndFor
    \State \Return $\pi_{\theta, \text{RL}}$

   \end{algorithmic}

    
    



\end{algorithm}

%% file: table_figs/fig14_tokenizer.tex
\begin{figure}[t]
    \centering
    \includegraphics[width=1.0\linewidth]{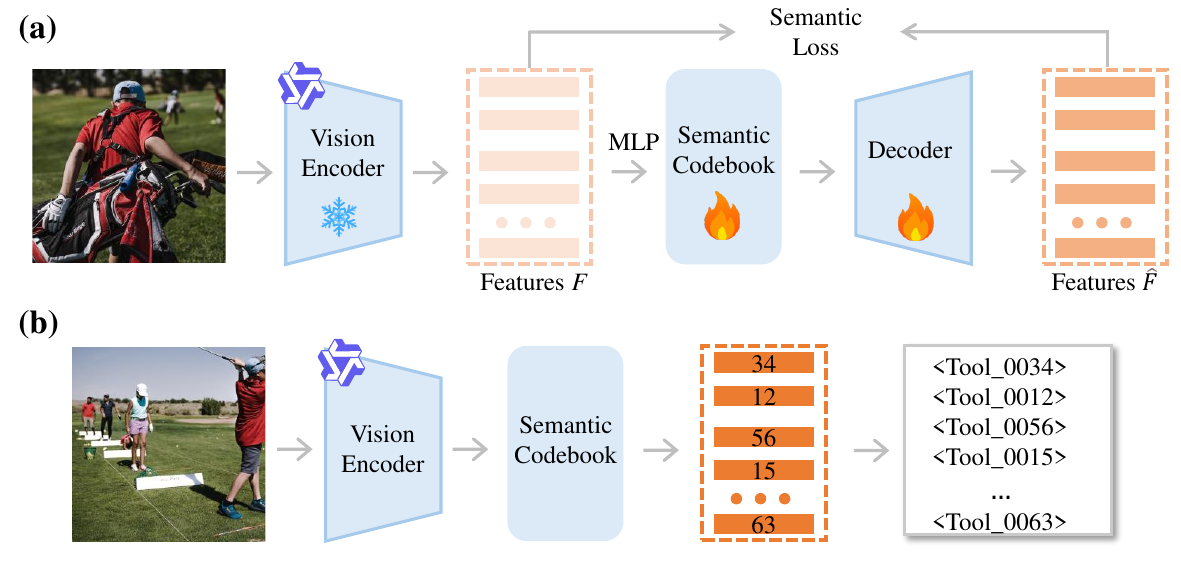}
    \caption{
        {\bf Training framework of the semantic tokenizer.}
    }
    \label{fig:fig14_tokenizer}
\end{figure}

%% file: table_figs/Algorithm_data.tex
\begin{algorithm}[t]
   \caption{{BEE-SFT Training Data} Construction Pipeline}
   \label{alg:data_process}
    \textbf{Input:} Base Dataset $\mathcal{D}_{base}$, a tool-call action $f(\cdot)$, a tool-tokenizer $T(\cdot)$, CoT generator $G$
     
    \textbf{Output:} BEE-SFT-320K Dataset $\mathcal{D}_{BEE,SFT}$
     
   \begin{algorithmic}[1]
    \State $\mathcal{D}_{BEE,SFT} \leftarrow$ $\varnothing$

    \For{each ${I,Q,A}$ $I$ in $D_{base}$}
    \State $C_{text}, C_{code} \leftarrow G(Q,A,I)$
    \State $I_{tool} \leftarrow f(C_{code})$ 
    \State $\hat{I} \leftarrow T(I_{tool})$
    \State $C \leftarrow C_{text} \cup Replace(C_{code}, \hat{I})$
    \State $\hat{C} \leftarrow$ Quality-check(C)
    \State $\mathcal{D}_{BEE,SFT} \leftarrow \mathcal{D}_{BEE,SFT} \cup \hat{C}$
    
    \EndFor
    \State \Return $\mathcal{D}_{BEE,SFT}$

   \end{algorithmic}

    
    



\end{algorithm}

%% file: table_figs/tab7_data_source.tex
\begin{table*}[t]
\centering
\caption{\textbf{Statistics of the BEE-SFT-320K dataset}. 
}
\label{tab:tab_aba_data}
\scalebox{1.0}{

\begin{tabular}{llll}
\hline
Source     & Domain                       & Visual Operation Type                        & Samples \\ \hline
Visual-CoT & Real-world, documents, chart & Cropping, drawing bounding box               & 20K    \\
Thyme      & Real-world, chart, OCR      & Cropping (\eg, zooming), rotation, low-contrast enhancement & 297.6K  \\
Zebra-CoT    & Real-world, chart            & Cropping, drawing auxiliary lines                                     & 2K      \\
ReFocus    & Chart                        & Drawing bounding box, highlighting, focus          & 0.4K    \\ \hline
\end{tabular}
}
\end{table*}

%% file: table_figs/Algorithm_system.tex
\begin{algorithm}[t]
    \caption{System-Level Metric Evaluation for MLLM Inference}
    \label{alg:metrics}
        \begin{algorithmic}[1]
            \Require Dataset $\mathcal{D}$, Node hourly cost $C_h$
            \State Initialize $N_{samples} \leftarrow 0$, $N_{tokens} \leftarrow 0$, $\mathcal{L} \leftarrow []$
            \State $T_{start} \leftarrow \text{CurrentTime}()$
            \For{each sample $x_i \in \mathcal{D}$}
                \State $t_{start}^i \leftarrow \text{CurrentTime}()$
                \State $y_i, n_{tokens}^i \leftarrow \text{ModelInference}(x_i)$ \Comment{Execute inference loop}
                \State $t_{end}^i \leftarrow \text{CurrentTime}()$
                \State $l_i \leftarrow t_{end}^i - t_{start}^i$ \Comment{Calculate sample latency}
                \State $\mathcal{L}\text{.append}(l_i)$
                \State $N_{samples} \leftarrow N_{samples} + 1$
                \State $N_{tokens} \leftarrow N_{tokens} + n_{tokens}^i$
            \EndFor
            \State $T_{end} \leftarrow \text{CurrentTime}()$
            \State $T_{total} \leftarrow T_{end} - T_{start}$
            
            \State \Comment{\textbf{Compute Throughput Metrics}}
            \State $\text{Throughput}_{samples} \leftarrow N_{samples} / T_{total}$
            \State $\text{Throughput}_{tokens} \leftarrow N_{tokens} / T_{total}$
            \State $\text{TPOT} \leftarrow (T_{total} / N_{tokens}) \times 1000$ \Comment{Time per Output Token (ms)}
            
            \State \Comment{\textbf{Compute Latency Statistics}}
            \State $L_{avg} \leftarrow \text{Mean}(\mathcal{L})$
            \State $L_{std} \leftarrow \text{StandardDeviation}(\mathcal{L})$
            \State $L_{P50}, L_{P90}, L_{P99} \leftarrow \text{Percentiles}(\mathcal{L}, \{50, 90, 99\})$
            
            \State \Comment{\textbf{Compute Cost Analysis}}
            \State $C_{sec} \leftarrow C_h / 3600$ \Comment{Cost per second}
            \State $\text{Cost}_{1k} \leftarrow (T_{total} / N_{samples}) \times 1000 \times C_{sec}$ \Comment{Cost per 1K samples}
            
            \State \Return $\text{Throughput}_{samples}, \text{Throughput}_{tokens}, \text{TPOT},$ \\
\hspace*{1.5cm} $L_{avg}, L_{std}, \{L_{P50}, L_{P90}, L_{P99}\}, \text{Cost}_{1k}$
        \end{algorithmic}
\end{algorithm}

%% file: table_figs/tab_supp0_sft_hyper.tex

\begin{table*}[t]
\centering
\caption{{\bf Hyperparameters used for SFT on Qwen2.5-VL-7B-Instruct and Qwen3-VL-2/4/8B-Instruct.}}
\label{tab:tab_supp0_sft}
\begin{tabularx}{\linewidth}{>{\hsize=0.8\hsize}X >{\hsize=1.2\hsize}X}
\toprule
\textbf{Parameters}              & \textbf{SFT}                                                                                                                                                              \\ \midrule
\textbf{General}                 &                                                                                                                                                                           \\
Model                            & \begin{tabular}[c]{@{}l@{}}Qwen/Qwen2.5-VL-7B-Instruct\\ Qwen/Qwen3-VL-2B-Instruct\\ Qwen/Qwen3-VL-4B-Instruct\\ Qwen/Qwen3-VL-8B-Instruct\end{tabular} \\
Thinking                         & False                                                                                                                                                                     \\
Train vision tower               & False                                                                                                                                                                     \\
Train projector                  & False                                                                                                                                                                     \\
Train language model             & True                                                                                                                                                                      \\ \midrule
\textbf{Data}                    &                                                                                                                                                                           \\
Prompt format                    & Chat template (system + user)                                                                                                                                             \\
Completion-only loss             & True                                                                                                                                                                      \\
Max. sequence length             & 8196                                                                                                                                                                      \\ \midrule
\textbf{Batching}                &                                                                                                                                                                           \\
Per-device train batch size      & 8                                                                                                                                                                         \\
Per-device eval batch size       & 1                                                                                                                                                                         \\
Gradient accumulation steps      & 1                                                                                                                                                                         \\
Effective global batch size      & 64                                                                                                                                                                        \\ \midrule
\textbf{Optimization / Training} &                                                                                                                                                                           \\
Optimizer                        & Adamw\_torch\_fused                                                                                                                                                       \\
Learning rate                    & 1e-5                                                                                                                                                                      \\
Scheduler                        & Cosine                                                                                                                                                                    \\
Warmup ratio                     & 0.03                                                                                                                                                                      \\
Num. epochs                      & 2                                                                                                                                                                         \\
Weight decay                     & 0.0                                                                                                                                                                       \\
Adam ${\beta}_{1}$                          & 0.9                                                                                                                                                                       \\
Adam ${\beta}_{2}$                           & 0.999                                                                                                                                                                     \\
Precision                        & bfloat16                                                                                                                                                                  \\
Gradient checkpointing           & True                                                                                                                                                                      \\
Deepspeed                        & Zero3                                                                                                                                                                     \\ \bottomrule
\end{tabularx}
\end{table*}

%% file: table_figs/tab_supp1_rl_hyper.tex

\begin{table*}[t]
\centering
\caption{{\bf Hyperparameters used for RL on Qwen2.5-VL-7B-Instruct and Qwen3-VL-2/4/8B-Instruct.}}
\label{tab:tab_supp1_rl}
\begin{tabularx}{\linewidth}{>{\hsize=0.8\hsize}X >{\hsize=1.2\hsize}X}
\toprule
\textbf{Parameters}                  & \textbf{DAPO}                                                                                                                                           \\ \midrule
\textbf{General}                     &                                                                                                                                                         \\
Model                                & \begin{tabular}[c]{@{}l@{}}Qwen/Qwen2.5-VL-7B-Instruct\\ Qwen/Qwen3-VL-2B-Instruct\\ Qwen/Qwen3-VL-4B-Instruct\\ Qwen/Qwen3-VL-8B-Instruct\end{tabular} \\
Thinking                             & False                                                                                                                                                   \\ \midrule
\textbf{Data}                        &                                                                                                                                                         \\
Max. prompt length                   & 4096                                                                                                                                                    \\
Max. response length                 & 4096                                                                                                                                                    \\
Train batch size                     & 512                                                                                                                                                     \\
Validation batch size                & 1024                                                                                                                                                    \\
Shuffle                              & True                                                                                                                                                    \\
Seed                                 & 1                                                                                                                                                       \\ \midrule
\textbf{Batching}                    &                                                                                                                                                         \\
Rollouts per prompt (n)              & 8                                                                                                                                                       \\
PPO mini-batch size                  & 128                                                                                                                                                     \\
PPO micro-batch size per GPU         & 2                                                                                                                                                       \\
Logprob micro-batch size per GPU     & 8                                                                                                                                                       \\
Max grad norm                        & 1.0                                                                                                                                                     \\ \midrule
\textbf{Rollout / Generation}        &                                                                                                                                                         \\
Inference engine                     & VLLM                                                                                                                                                    \\
Temperature                          & 1.0                                                                                                                                                     \\
Top-p                                & 1.0                                                                                                                                                     \\
Top-k                                & -1                                                                                                                                                      \\
Max new tokens                       & 8192                                                                                                                                                    \\
Tensor parallel size                 & 1                                                                                                                                                       \\
Enable chunked prefill               & True                                                                                                                                                    \\ \midrule
\textbf{Actor / Policy Optimization} &                                                                                                                                                         \\
PPO epochs                           & 3                                                                                                                                                       \\
Clip ratio low                       & 0.2                                                                                                                                                     \\
Clip ratio high                      & 0.3                                                                                                                                                     \\
Online filtering                     & False                                                                                                                                                   \\
Disable KL                           & True                                                                                                                                                    \\
Dynamic batching                     & True                                                                                                                                                    \\ \midrule
\textbf{Reference Model}             &                                                                                                                                                         \\
Strategy                             & FSDP                                                                                                                                                    \\
CPU offload                          & True                                                                                                                                                    \\
Offload params                       & False                                                                                                                                                   \\ \midrule
\textbf{Parallelism}                 &                                                                                                                                                         \\
Number of GPUs                       & 8                                                                                                                                                       \\
Number of nodes                      & 1                                                                                                                                                       \\
Sequence parallel                    & 1                                                                                                                                                       \\
Gradient checkpointing               & True                                                                                                                                                    \\ \midrule
\textbf{Training}                    &                                                                                                                                                         \\
Optimizer                            & Adamw                                                                                                                                                   \\
Learning rate                        & 2e-6                                                                                                                                                \\
Scheduler                            & Constant                                                                                                                                                \\
Adam ${\beta}$                                 & (0.9, 0.999)                                                                                                                                            \\
Warmup ratio                         & 0.0                                                                                                                                                     \\
Weight decay                         & 1e-2                                                                                                                                                    \\
Precision                            & Bfloat16                                                                                                                                                \\ \bottomrule
\end{tabularx}
\end{table*}

%% file: table_figs/fig10_implicit_tool_compare.tex
\begin{figure*}[ht]
    \centering
    \includegraphics[width=1.0\linewidth]{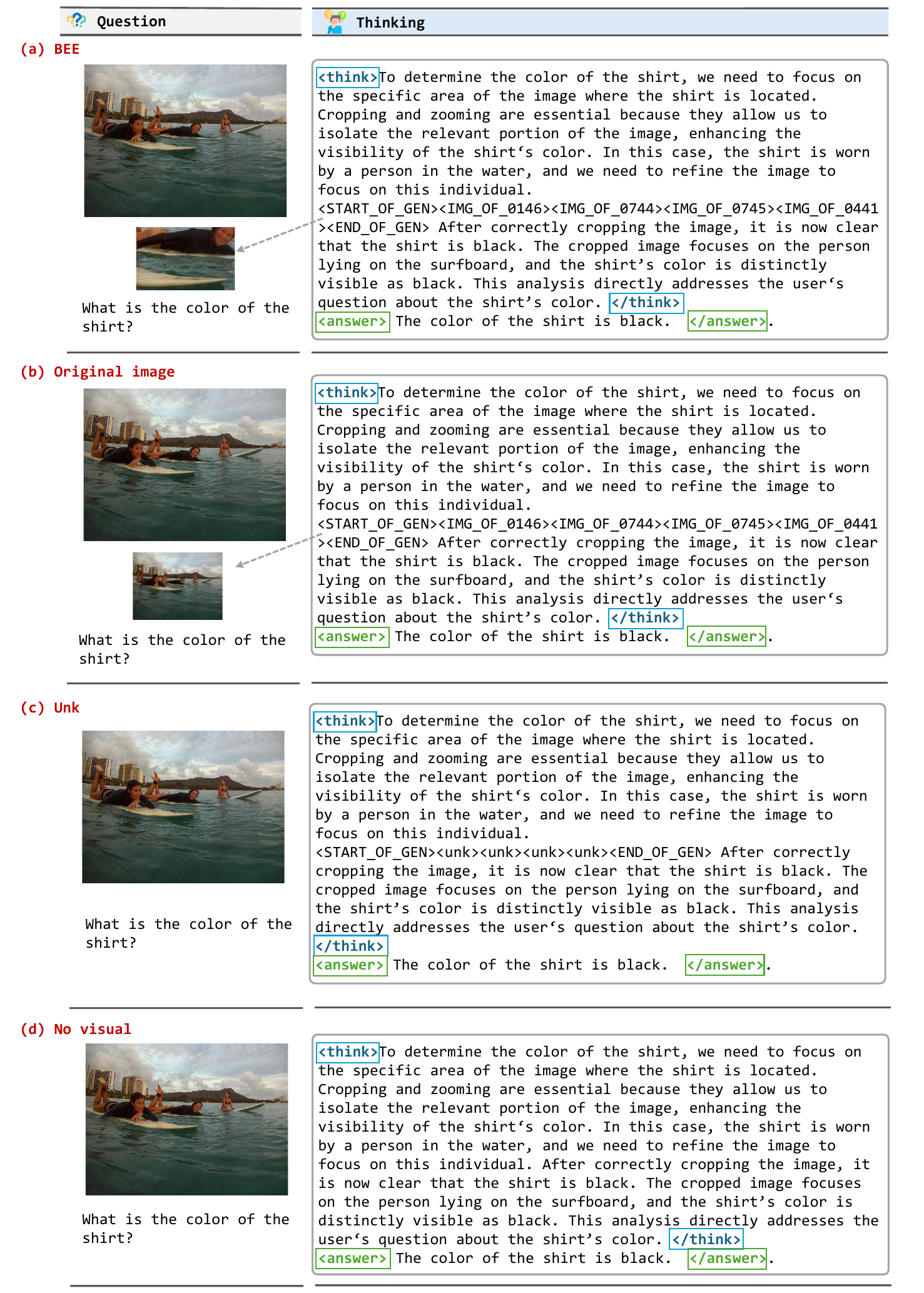}
    \caption{
        {\bf Variant forms of tool slot in the training phase.}
    }
    \label{fig:fig10_compare}
\end{figure*}

%% file: table_figs/tab_supp2_efficiency.tex

\begin{table*}[t]
\centering
\caption{{\bf Efficiency comparison settings used for Thyme, Mini-o3, Qwen2.5/3-VL, DeepEyes, and {\MethodName}.}}
\label{tab:tab_supp2_efficiency}
\begin{tabularx}{\linewidth}{>{\hsize=0.8\hsize}X >{\hsize=1.2\hsize}X}
\toprule
Parameters                   & Details                     \\ \midrule
\textbf{Hardware / Software} &                             \\
Platform                     & H800                        \\
GPUs                         & 1                           \\
CUDA                         & 12.4                        \\
NVIDIA Driver                & 570.124.06                  \\
FlashAttention               & 2.8.3                       \\
VLLM                         & 0.8.5                       \\
Transformer                  & 4.49.0                      \\
Torch                        & 2.6.0                       \\
Python                       & 3.10.0                      \\ \midrule
\textbf{Evaluation}          &                             \\
Precision                    & Bfloat16                    \\
Max. input length            & -                           \\
Max. generation length       & 2048                        \\
Temperature                  & 0.01                        \\
Top-p                        & 0.001                       \\
Top-k                        & 1                           \\
Prompt format                & Chat template (system+user) \\ \midrule
\textbf{Benchmark}           &                             \\
Dataset                      & HRBench4K                   \\
Nums                         & 800                         \\ \bottomrule
\end{tabularx}
\end{table*}

%% file: table_figs/fig9_sft_example.tex
\begin{figure*}[ht]
    \centering
    \includegraphics[width=1.0\linewidth]{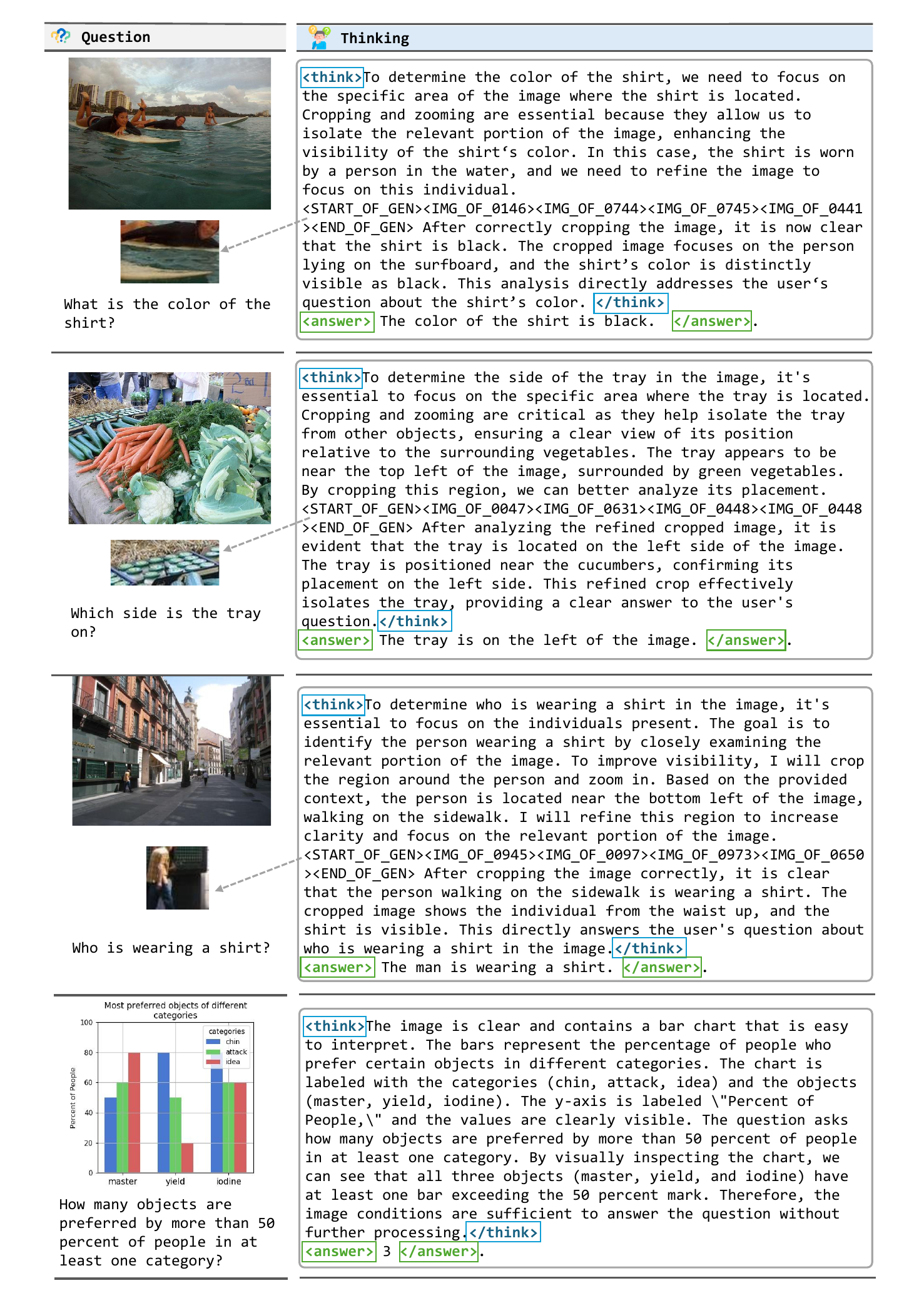}
    \caption{
        {\bf Data format of the final SFT training set.} 
    }
    \label{fig:fig9_sft}
\end{figure*}

%% file: table_figs/tab8_all_results.tex
\begin{table*}[t]
\centering
\caption{{\bf Performance comparison on various tasks.} For all open-source models, the best performance for each metric is {\bf bolded}, and the second best is {\ul underlined}. 
The result ($^{\dagger}$, $^{\ddagger}$) is collected from \cite{zhang2025skywork} and \cite{bai2025qwen3}, respectively.
}
\label{tab:tab8_main}
\scalebox{0.95}{

\begin{tabular}{lc
>{\columncolor[HTML]{DDEEFF}}c cccc}
\hline
Benchmark                       & Split      & BEE-7B        & Qwen2.5-VL-7B & Qwen2.5-VL-32B & \multicolumn{1}{l}{ZwZ-7B} & {\color[HTML]{656565} Gemini 2.5 Pro} \\ \hline
\multicolumn{7}{c}{\cellcolor[HTML]{F6D8D6}Perception}                                                                                                             \\
V*                              & Attribute  & {\ul 80.3}    & 78.2          & 77.4           & \textbf{82.6}              & {\color[HTML]{656565} -}              \\
                                & Spatial    & {\ul 86.1}    & 73.6          & \textbf{86.8}  & 82.9                       & {\color[HTML]{656565} -}              \\
                                & Overall    & \textbf{83.8} & 76.4          & 81.2           & {\ul 82.7}                 & {\color[HTML]{656565} 79.1}           \\ \cline{2-7} 
                                & FSP        & \textbf{92.5} & 85.2          & {\ul 87.5}     & 90.5                       & {\color[HTML]{656565} -}              \\
                                & FCP        & {\ul 58.3}    & 52.2          & \textbf{59.3}  & 56.5                       & {\color[HTML]{656565} -}              \\
\multirow{-3}{*}{HRBench-4K}    & Overall    & \textbf{75.4} & 68.8          & 73.4           & {\ul 73.5}                 & {\color[HTML]{656565} 83.9}           \\ \cline{2-7} 
                                & FSP        & \textbf{87.0} & 78.8          & {\ul 82.3}     & 59.7                       & {\color[HTML]{656565} -}              \\
                                & FCP        & {\ul 60.5}    & 51.8          & 58.5           & \textbf{87.7}              & {\color[HTML]{656565} -}              \\
\multirow{-3}{*}{HRBench-8K}    & Overall    & \textbf{73.9} & 65.3          & 70.4           & {\ul 73.7}                 & {\color[HTML]{656565} 81.5}           \\ \cline{2-7} 
                                & Perception & {\ul 67.2}    & 60.6          & 63.8           & \textbf{67.9}              & {\color[HTML]{656565} -}              \\
                                & Reasoning  & \textbf{43.7} & 38.6          & 40.4           & {\ul 43.8}                 & {\color[HTML]{656565} -}              \\
\multirow{-3}{*}{MME-Real}      & Overall    & {\ul 64.4}    & 58.3          & 61.0           & \textbf{65.0}              & {\color[HTML]{656565} 71.3}           \\ \cline{2-7} 
                                & Perception & {\ul 56.5}    & 48.8          & {\ul 50.6}     & \textbf{57.7}              & {\color[HTML]{656565} -}              \\
                                & Reasoning  & \textbf{45.7} & 37.7          & {\ul 39.3}     & 43.7                       & {\color[HTML]{656565} -}              \\
\multirow{-3}{*}{MME-Real-Lite} & Overall    & \textbf{52.3} & 44.1          & 46.2           & {\ul 52.2}                 & {\color[HTML]{656565} 58.3}           \\ \cline{2-7} 
CVBench-2D                      & Overall    & \textbf{75.0} & 74.2          & 72.2           & {\ul 74.7}                 & {\color[HTML]{656565} -}              \\ \hline
\multicolumn{7}{c}{\cellcolor[HTML]{E8ECF2}Reasoning}                                                                                                              \\
LogicVista                      & Overall    & 43.0          & 39.8          & \textbf{54.4}  & {\ul 47.7}                 & {\color[HTML]{656565} 66.4}           \\ \cline{2-7} 
VisuLogic                       & Overall    & {\ul 23.4}    & 20.0          & \textbf{24.7}  & 3.9                        & {\color[HTML]{656565} 26.9}           \\ \cline{2-7} 
WeMath                          & Overall    & {\ul 38.6}    & 34.3          & \textbf{47.1}  & 37.9                       & {\color[HTML]{656565} -}              \\ \hline
\multicolumn{7}{c}{\cellcolor[HTML]{EFE8E8}General}                                                                                                                \\
MMStar                          & Overall    & 63.5          & {\ul 64.7}    & \textbf{69.6}  & 62.9                       & {\color[HTML]{656565} 78.5}           \\ \cline{2-7} 
POPE                            & Overall    & \textbf{89.3} & 85.8          & {\ul 86.2}     & 87.7                       & {\color[HTML]{656565} -}              \\ \cline{2-7} 
SeedBench-Image                 & Overall    & 75.2          & {\ul 77.1}    & \textbf{77.5}  & 71.8                       & {\color[HTML]{656565} -}              \\ \cline{2-7} 
OCRBench\_mini                  & Overall    & {\ul 16.3}    & 15.8          & 15.1           & \textbf{17.5}              & {\color[HTML]{656565} -}              \\ \cline{2-7} 
MME                             & Overall    & {\ul 2307}    & 2290          & \textbf{2388}  & 2301                       & {\color[HTML]{656565} -}              \\ \cline{2-7} 
CMMMU\_val                      & Overall    & {\ul 73.9}    & 73.5          & \textbf{79.4}  & 70.9                       & {\color[HTML]{656565} -}              \\ \hline
\end{tabular}
}
\end{table*}



%% file: table_figs/tab2_splitdataset.tex
\begin{table*}[t]
\centering
\caption{{\bf Performance comparison on the MME-RealWorld.} BEE shows larger improvements on more challenging perception and reasoning tasks, as well as monitoring and autonomous driving tasks, where the baseline performs relatively poorly.
}
\label{tab:tab2_splitdataset_2}
\scalebox{0.95}{

\begin{tabular}{lcccccc}
\hline
\multicolumn{7}{c}{\cellcolor[HTML]{D6ECFA}Perception}                                                                               \\
Model                & Monitoring      & Autonomous Driving & OCR            & Diagram and Table & Remote Sensing  & Overall         \\
Qwen2.5-VL-7B        & 38.8            & 22.7               & 87.0           & 78.8              & 45.4            & 60.9            \\
BEE-7B               & 44.8            & 42.7               & 88.4           & 79.5              & 53.7            & 67.2            \\
\textit{Improvement} & \textit{15.5\%} & \textit{88.1\%}    & \textit{1.6\%} & \textit{0.9\%}    & \textit{18.3\%} & \textit{10.3\%} \\ \hline
\multicolumn{7}{c}{\cellcolor[HTML]{F0E5DE}Reasoning}                                                                                \\
Model                & Monitoring      & Autonomous Driving & OCR            & Diagram and Table & Remote Sensing  & Overall         \\
Qwen2.5-VL-7B        & 26.1            & 24.3               & 64.8           & 63.4              & -               & 38.6            \\
BEE-7B               & 39.4            & 29.7               & 69.4           & 60.2              & -               & 43.7            \\
\textit{Improvement} & \textit{51.0\%} & \textit{22.2\%}    & \textit{7.1\%} & \textit{-5.0\%}   & \textit{-}      & \textit{13.2\%} \\ \hline
\end{tabular}

}
\end{table*}



%% file: table_figs/fig13_decouple_tsne.tex
\begin{figure*}[ht]
    \centering
    \includegraphics[width=1.0\linewidth]{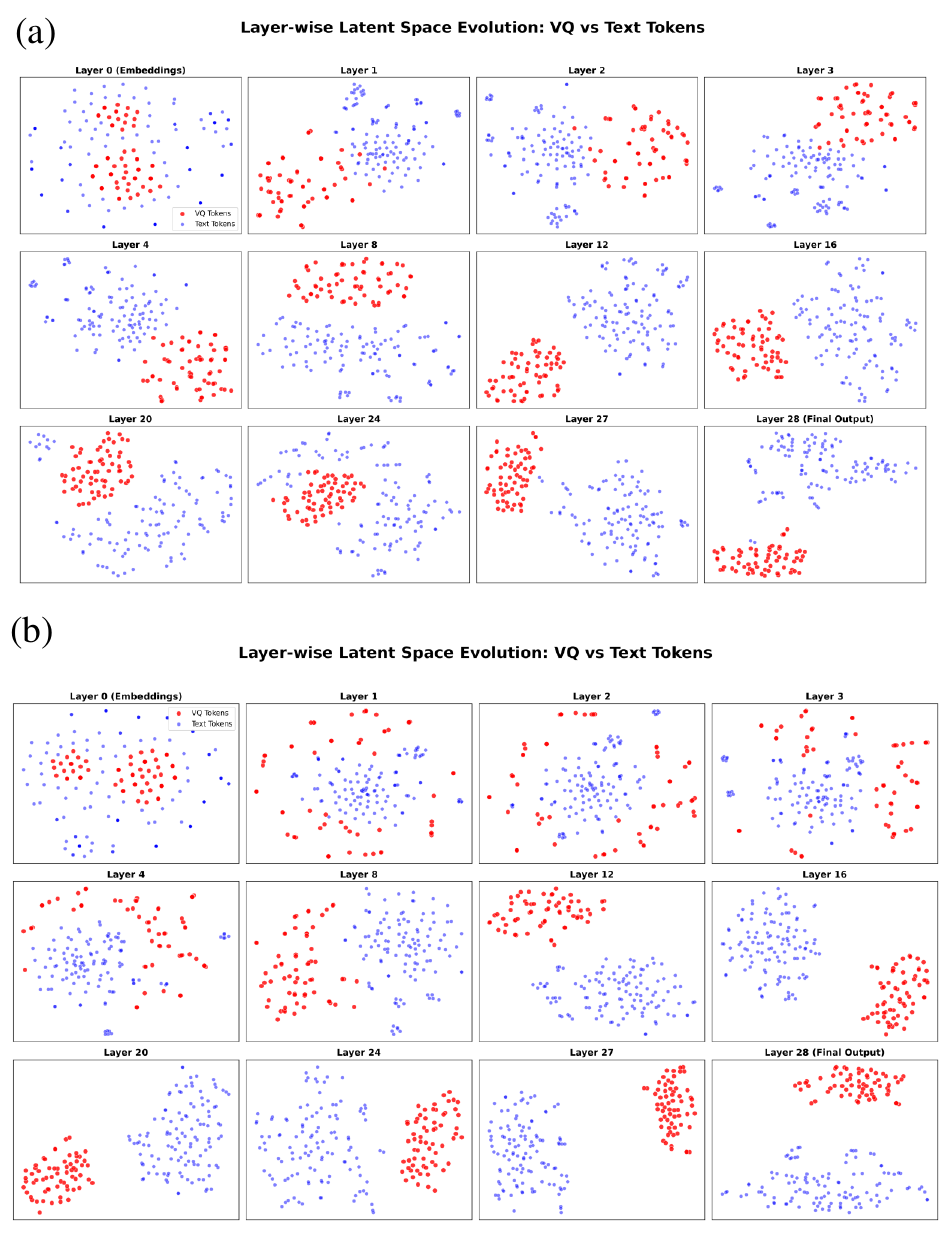}
    \caption{
        {\bf Visualization of perception and reasoning tokens.} (a): T-SNE visualization without text-vision decoupling. (b) T-SNE visualization with text-vision decoupling.
    }
    \label{fig:fig13_decouple}
\end{figure*}

%% file: table_figs/Algorithm_zwz.tex
\begin{algorithm}[t]
    \caption{Cascading Evaluation Pipeline for ZwZ Answer Matching}
    \label{alg:evaluation_pipeline}
    \begin{algorithmic}[1]
        \Require Dataset $\mathcal{D}$
        \State $is\_mcq \leftarrow \text{True}$ if $\mathcal{D}$ is an MCQ benchmark, else $\text{False}$
        \State Initialize $Q_{llm} \leftarrow []$ \Comment{Queue for samples requiring LLM intervention}
        
        \For{each sample $x_i \in \mathcal{D}$}
            \State \Comment{\textbf{Step 1: Extract the core answer from the model's raw output}}
            \If{$x_i.output$ contains ``$<answer>$''}
                \State $ans_{ext} \leftarrow \text{ExtractTextInsideTags}(x_i.output)$
            \ElsIf{$x_i.output$ contains ``Answer:''}
                \State $ans_{ext} \leftarrow \text{ExtractTextAfterColon}(x_i.output)$
            \Else
                \State $ans_{ext} \leftarrow \text{ExtractLastThreeLines}(x_i.output)$
            \EndIf
            
            \State \Comment{\textbf{Step 2: Strict rule-based matching}}
            \State $is\_correct \leftarrow \text{MathRulerGrade}(x_i.gt, ans_{ext})$
            
            \If{$is\_correct == \text{True}$}
                \State $x_i.status \leftarrow \text{``Yes''}, \ x_i.source \leftarrow \text{``mathruler''}$
                \State \textbf{continue} \Comment{Proceed to the next sample}
            \EndIf
            
            \State \Comment{\textbf{Step 3: Heuristic option matching (MCQ datasets only)}}
            \If{$is\_correct == \text{False}$ \textbf{and} $is\_mcq == \text{True}$}
                \State $L_{gt} \leftarrow \text{RegexExtractOption}(x_i.gt)$
                \State $L_{pred} \leftarrow \text{RegexExtractFirstOptionLetter}(ans_{ext})$
                
                \If{$L_{gt} \neq \emptyset$ \textbf{and} $L_{gt} == L_{pred}$}
                    \State $x_i.status \leftarrow \text{``Yes''}, \ x_i.source \leftarrow \text{``first\_letter\_match''}$
                    \State \textbf{continue}
                \EndIf
            \EndIf
            
            \State \Comment{\textbf{Step 4: Enqueue failed or ambiguous samples for LLM Evaluation}}
            \State $x_i.status \leftarrow \text{``Pending''}$
            \State $p_i \leftarrow \text{BuildJudgePrompt}(x_i.question, x_i.gt, ans_{ext})$
            \State $Q_{llm}\text{.append}((i, p_i))$
        \EndFor
        
        \State \Comment{\textbf{Step 5: Batch LLM Generation for final adjudication}}
        \If{$Q_{llm} \neq \emptyset$}
            \State $R_{llm} \leftarrow \text{VLLM\_Generate}(Q_{llm}.prompts)$
            
            \For{each $(result, i)$ \textbf{in} $\text{Zip}(R_{llm}, Q_{llm}.indices)$}
                \If{$result$ contains ``Yes'' \textbf{or} ``yes''}
                    \State $x_i.status \leftarrow \text{``Yes''}$
                \Else
                    \State $x_i.status \leftarrow \text{``No''}$
                \EndIf
                \State $x_i.source \leftarrow \text{``llm\_judge''}$
            \EndFor
        \EndIf
        
        \State \Comment{\textbf{Final Statistics}}
        \State $Accuracy \leftarrow \text{Count}(x_i.status == \text{``Yes''}) / |\mathcal{D}|$
        \State \Return $Accuracy$
    \end{algorithmic}
\end{algorithm}

%% file: table_figs/Algorithm_bee.tex
\begin{algorithm}[t]
      \caption{Cascading Answer Matching Pipeline for VLMEvalKit}                                                              
      \label{alg:hrbench4k_matching}                                                                                                                    
      \begin{algorithmic}[1]                                                                                                                            
          \Require Dataset $\mathcal{D}$, Judge LLM $\mathcal{M}$, Choices $\mathcal{C}=\{A, B, C, D\}$                                                 
                                                                                                                                                        
          \For{each sample $x_i \in \mathcal{D}$}                                                                                                       
              \State \Comment\textbf{Step 1: Rule-based option letter matching}                                                                             
              \State $ans \leftarrow x_i.\text{prediction}$                                                                                             
              \If{$ans$ contains rejection phrases (e.g., ``Sorry, I can't help'')}                                                                     
                  \State $opt \leftarrow \text{`Z'}$, \textbf{goto} \textsc{Verdict} \Comment{Refused to answer}                                                                                                                                                
              \EndIf                                                                                                                                    
                                                                                                                                                        
              \State Strip punctuation $\{.\ ()\ [\ ],:;!*\#\{\}\}$ from $ans$, $tokens \leftarrow \text{Split}(ans)$ by whitespace                                                                        
              \State $count \leftarrow$ number of tokens $\in \mathcal{C}$ appearing in $tokens$                                                        
                                                                                                                                                        
              \If{$count == 1$}                                                                                                                         
                  \If{matched letter is `A' \textbf{and} $|tokens| > 3$}                                                                                
                      \State \textbf{goto} \textsc{Step2} \Comment{`A' likely a quantifier, skip}                                                       
                  \EndIf                                                                                                                                
                  \State $opt \leftarrow$ the matched letter, \textbf{goto} \textsc{Verdict}                                                                                                                                        
              \ElsIf{$count == 0$ \textbf{and} `Z' $\in tokens$}                                                                                        
                  \State $opt \leftarrow \text{`Z'}$, \textbf{goto} \textsc{Verdict}                                                                                                                                         
              \EndIf                                                                                                                                    
                                                                        
              \State \Comment\textbf{Step 2: Rule-based option text matching}                                                                               
              \State \textsc{Step2}: $ans_{lower} \leftarrow \text{Lowercase}(ans)$                                                                                                                                          
              \If{$|ans_{lower}| > 2 \times \sum_{c \in \mathcal{C}} |c_{\text{text}}|$}
                  \State \textbf{goto} \textsc{Step3} \Comment{Answer too long, prone to false match}                                                   
              \EndIf                                                                                                                                    
              \State $cands \leftarrow []$                                                                                                              
              \For{$c \in \mathcal{C}$}                                                                                                                 
                  \If{$\text{Lowercase}(c_{\text{text}}) \subset ans_{lower}$}                                                                          
                      \State $cands$.append$(c)$                                                                                                        
                  \EndIf                                                                                                                                
              \EndFor                                                                                                                                   
              \If{$|cands| == 1$}                                                                                                                       
                  \State $opt \leftarrow cands[0]$, \textbf{goto} \textsc{Verdict}
              \EndIf                                                                                                                                    
                                                                        
              \State \Comment\textbf{Step 3: LLM-based answer extraction}                                                                                   
              \State \textsc{Step3}:                               
              \If{$\mathcal{M}$ is not available}                                                                                                       
                  \State $opt \leftarrow \text{`Z'}$, \textbf{goto} \textsc{Verdict} \Comment{No LLM under exact\_matching policy}                                                                                                                                                    
              \EndIf                                                                                                                                    
                                                                                                                                                        
              \For{$retry = 1$ \textbf{to} $3$}                                                                                                         
                  \State $prompt \leftarrow \text{BuildPrompt}(x_i.\text{question}, \mathcal{C}, ans)$, $llm\_ans \leftarrow \mathcal{M}.\text{generate}(prompt)$                                                                      
                  \If{$llm\_ans$ contains API failure message} \textbf{continue} \EndIf                                                                                                                                                                                                                                  
                  \State $opt \leftarrow \text{CanInfer}(llm\_ans, \mathcal{C})$ \Comment{Repeat Step 1 \& 2 on LLM output}                                         
                  \If{$opt \neq \text{False}$} \textbf{goto} \textsc{Verdict} \EndIf
              \EndFor

              \State \Comment\textbf{Step 4: Random fallback}
              \State $opt \leftarrow \text{RandomChoice}(\mathcal{C} \cup \{\text{`Z'}\})$

              \State \textbf{// Verdict}
              \State \textsc{Verdict}: $x_i.\text{hit} \leftarrow (opt == x_i.\text{GT}) \ ? \ 1 : 0$
          \EndFor

          \State \Return $\text{Accuracy} \leftarrow \frac{1}{|\mathcal{D}|} \sum_{x_i \in \mathcal{D}} x_i.\text{hit}$
      \end{algorithmic}
  \end{algorithm}

%% file: table_figs/tab13_tau.tex
\begin{table}[t]
\centering
\caption{
        {\bf Ablation study on $\tau$.} For all open-source models, the best performance for each metric is \textbf{bolded}.
}
\label{tab:tab13_tau}

\begin{tabular}{lccc}
\hline
Settings of $\tau$ & \multicolumn{1}{l}{V*} & \multicolumn{1}{l}{HRBench4K} & \multicolumn{1}{l}{HRBench8K} \\ \hline
\multicolumn{4}{c}{\textit{BEE-7B (Qwen2.5-VL)}}                                                             \\
1/3                 & 81.6                   & 73.7                          & 71.8                          \\
1/2                 & 82.8                   & 73.1                          & 72.2                          \\
2/3                 & \textbf{83.8}          & \textbf{75.4}                 & \textbf{73.9}                 \\
3/4                 & 83.7                   & 74.3                          & 73.5                          \\ \hline
\multicolumn{4}{c}{\textit{BEE-4B (Qwen3-VL)}}                                                               \\
1/3                 & 89.5                   & 76.6                          & 75.5                          \\
1/2                 & 91.2                   & 78.0                          & 76.3                          \\
2/3                 & \textbf{91.6}          & 78.4                          & \textbf{77.0}                 \\
3/4                 & 89.5                   & \textbf{78.6}                 & 76.5                          \\ \hline
\end{tabular}
\end{table}

%% file: table_figs/tab14_part.tex
\begin{table}[t]
\centering
\caption{
        {\bf Ablation study on each component.} For all open-source models, the best performance for each metric is \textbf{bolded}.
}
\label{tab:tab14_part}

\begin{tabular}{lccc}
\hline
Models        & \multicolumn{1}{l}{V*} & \multicolumn{1}{l}{HRBench4K} & \multicolumn{1}{l}{HRBench8K} \\ \hline
Qwen2.5-VL-7B & 76.4                   & 68.8                          & 65.3                          \\
SFT           & 75.4                   & 71.1                          & 65.8                          \\
BEE-7B        & \textbf{83.8}          & \textbf{75.4}                 & \textbf{73.9}                 \\ \hline
\end{tabular}
\end{table}

%% file: table_figs/fig4_curve.tex
\begin{figure*}[ht]
    \centering
    \includegraphics[width=1.0\linewidth]{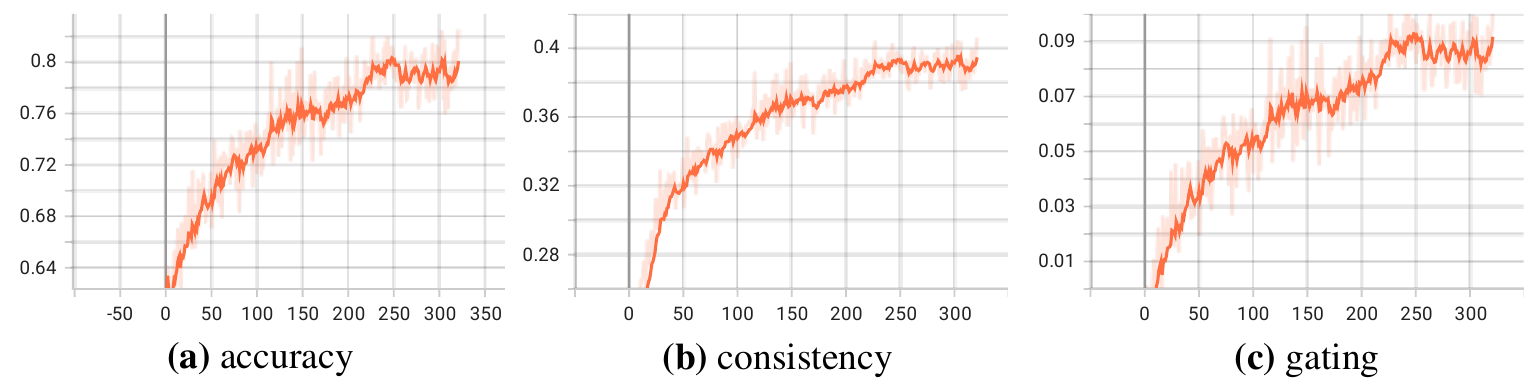}
    \caption{
        { \textbf{The training reward of Qwen2.5-VL-7B}.}
    }
    \label{fig:fig4_curve}
\end{figure*}

%% file: table_figs/fig7_aba1_curve.tex
\begin{figure*}[ht]
    \centering
    \includegraphics[width=1.0\linewidth]{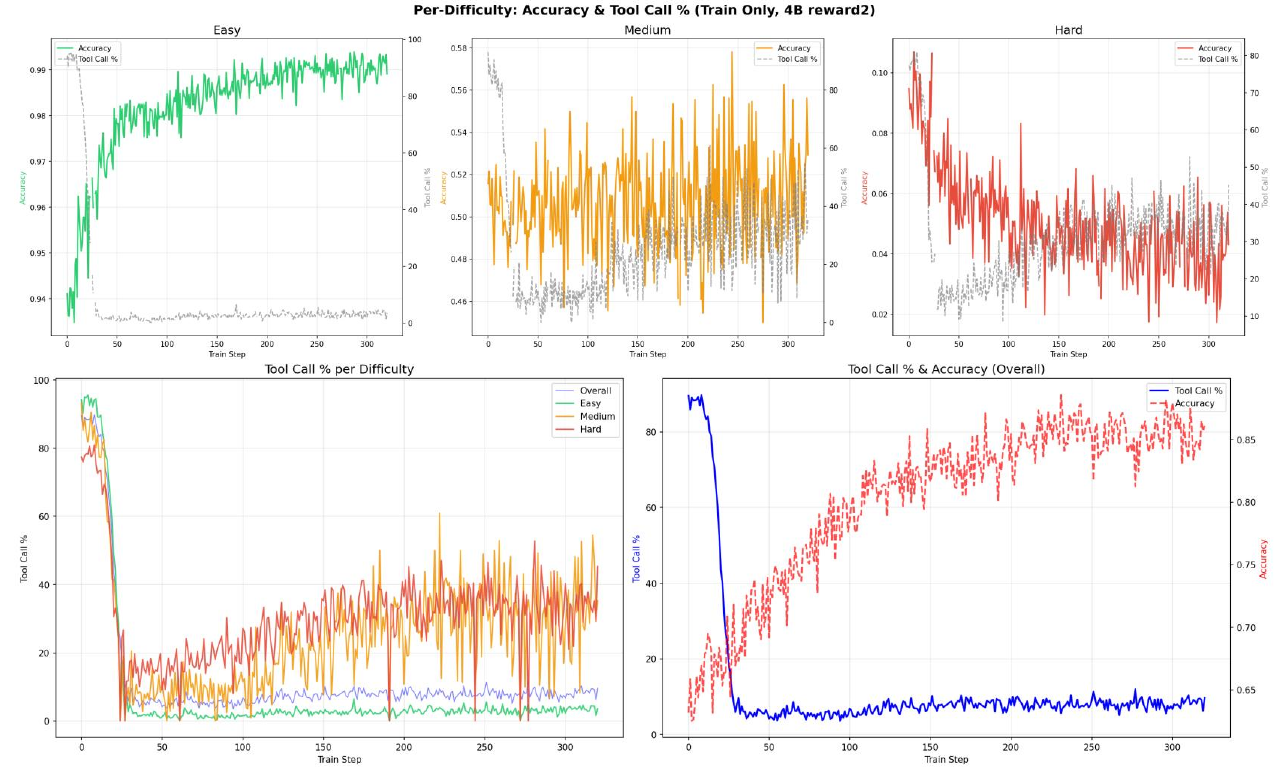}
    \caption{
        { \textbf{Tool usage rate and accuracy over time}.}
    }
    \label{fig:fig7_aba1}
\end{figure*}

%% file: table_figs/fig7_aba2_curve.tex
\begin{figure*}[ht]
    \centering
    \includegraphics[width=1.0\linewidth]{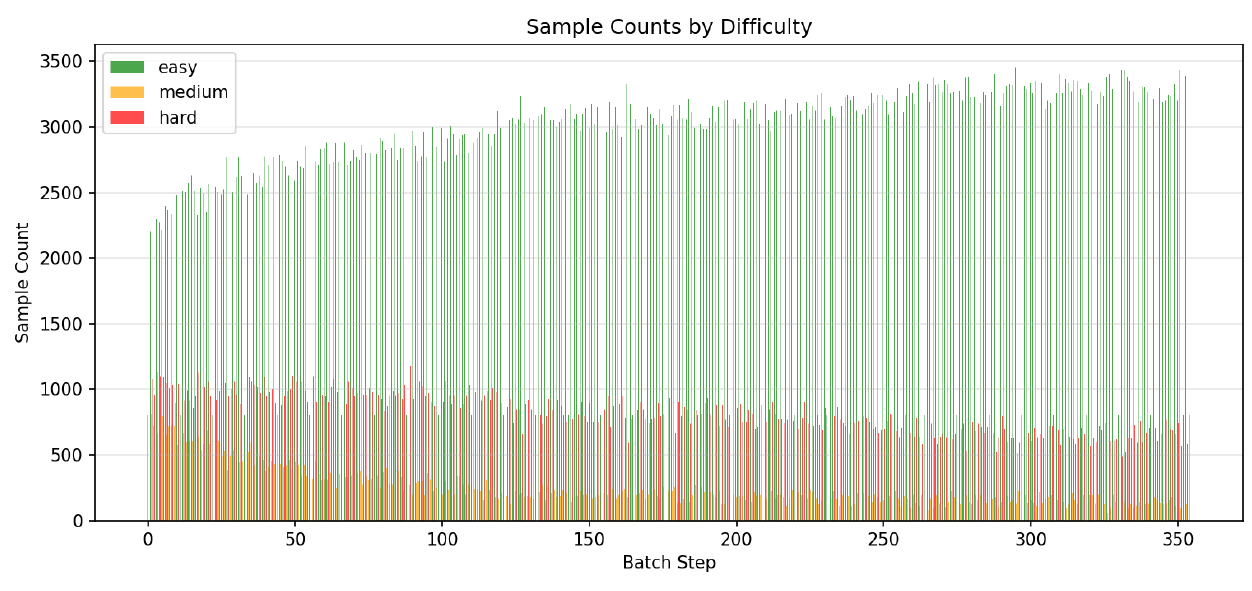}
    \caption{
        \textbf{Number of problems of different difficulties during training.}
    }
    \label{fig:fig7_aba2}
\end{figure*}

%% file: table_figs/fig8_output1.tex
\begin{figure*}[ht]
    \centering
    \includegraphics[width=1.0\linewidth]{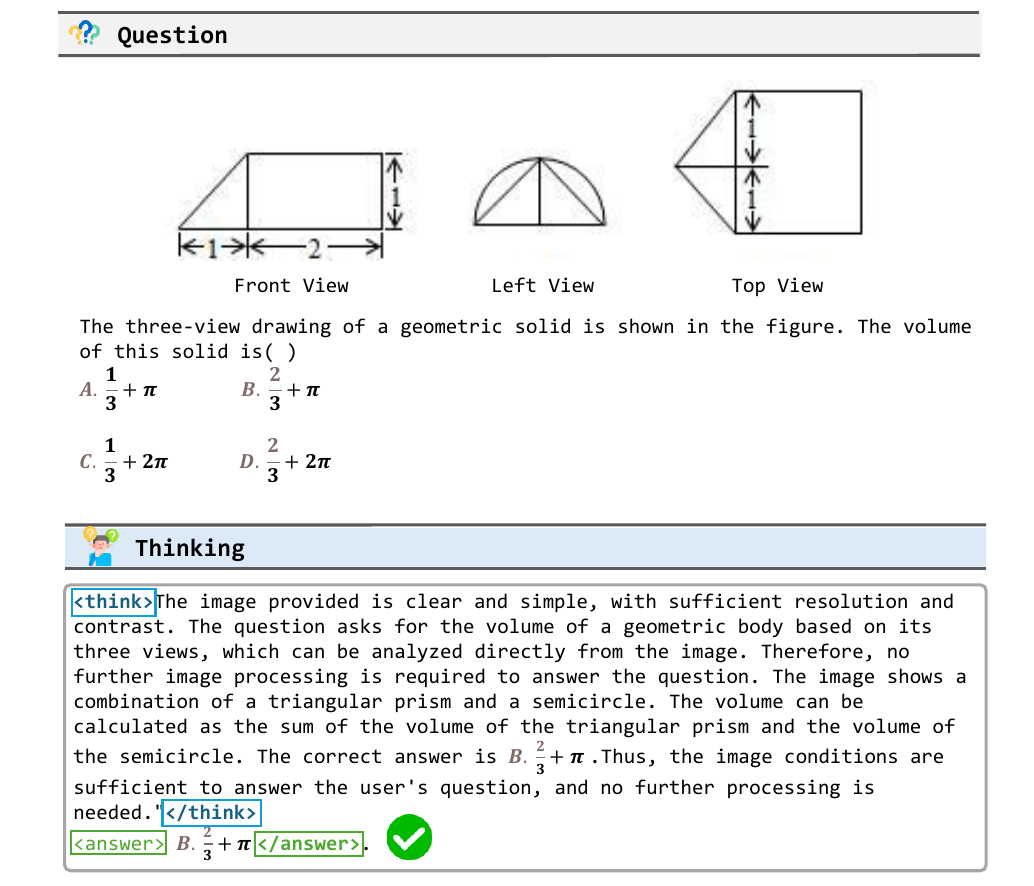}
    \caption{
        Mathematical reasoning case.
    }
    \label{fig:fig8_o1}
\end{figure*}

%% file: table_figs/fig8_output2.tex
\begin{figure*}[ht]
    \centering
    \includegraphics[width=1.0\linewidth]{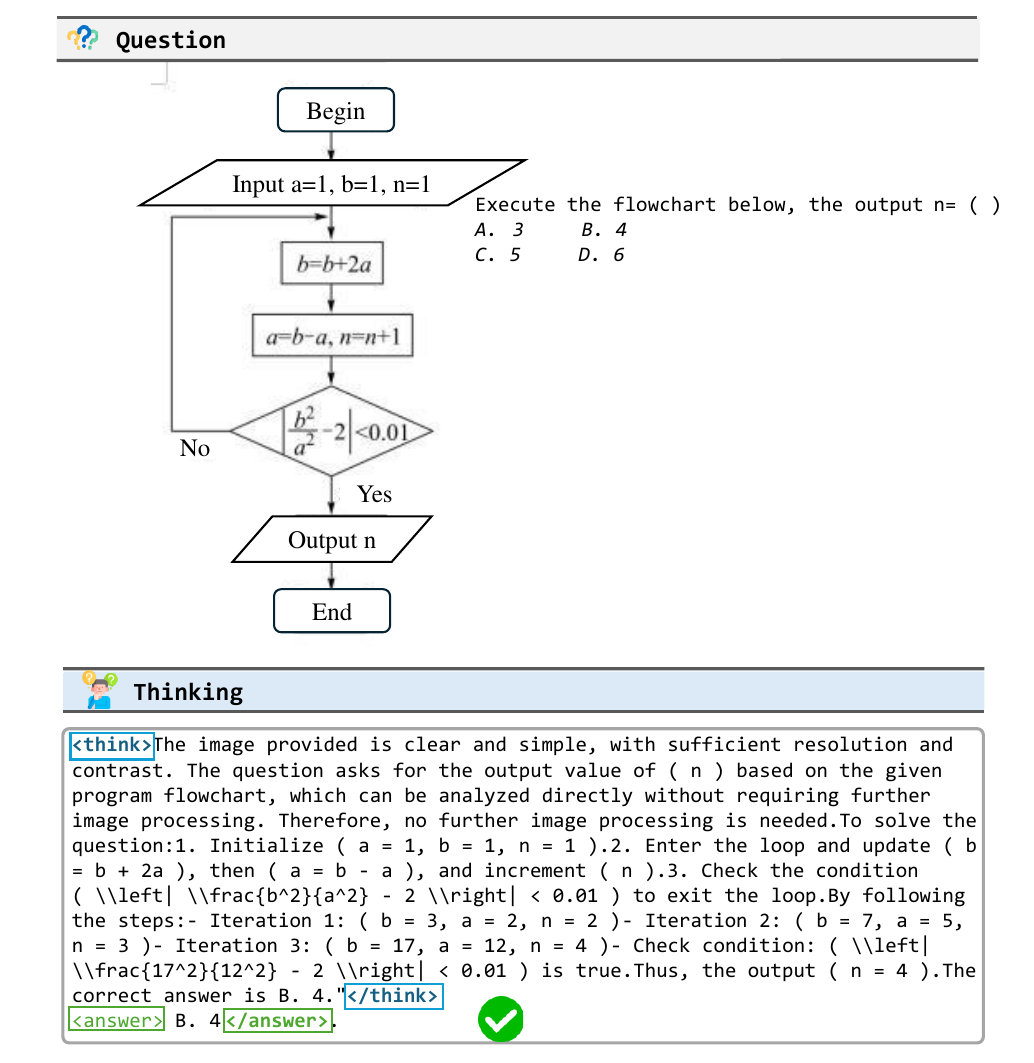}
    \caption{
        General reasoning case.
    }
    \label{fig:fig8_o2}
\end{figure*}

%% file: table_figs/fig8_output4.tex
\begin{figure*}[ht]
    \centering
    \includegraphics[width=1.0\linewidth]{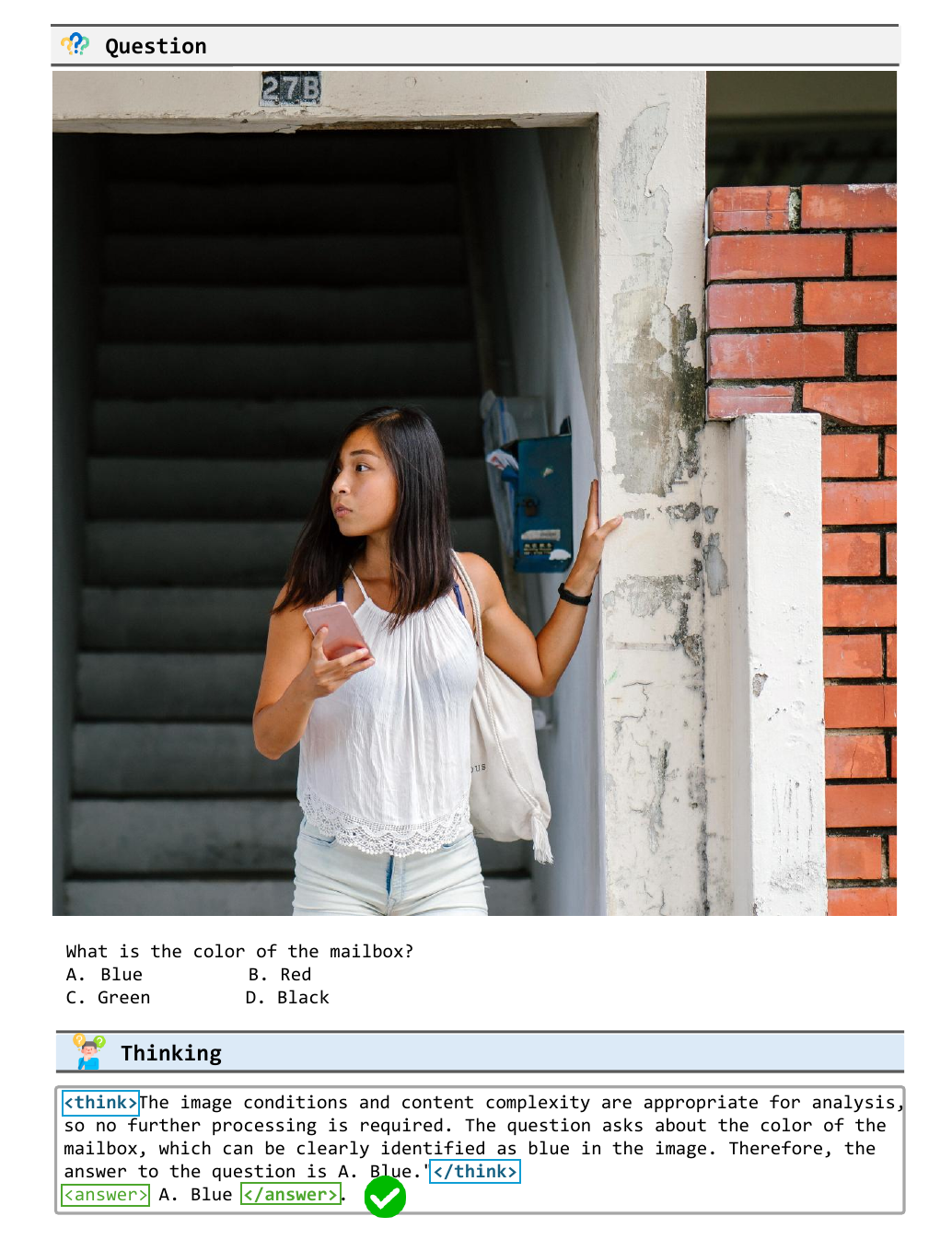}
    \caption{
        Real-world understanding case.
    }
    \label{fig:fig8_o4}
\end{figure*}

%% file: table_figs/fig8_output5.tex
\begin{figure*}[ht]
    \centering
    \includegraphics[width=1.0\linewidth]{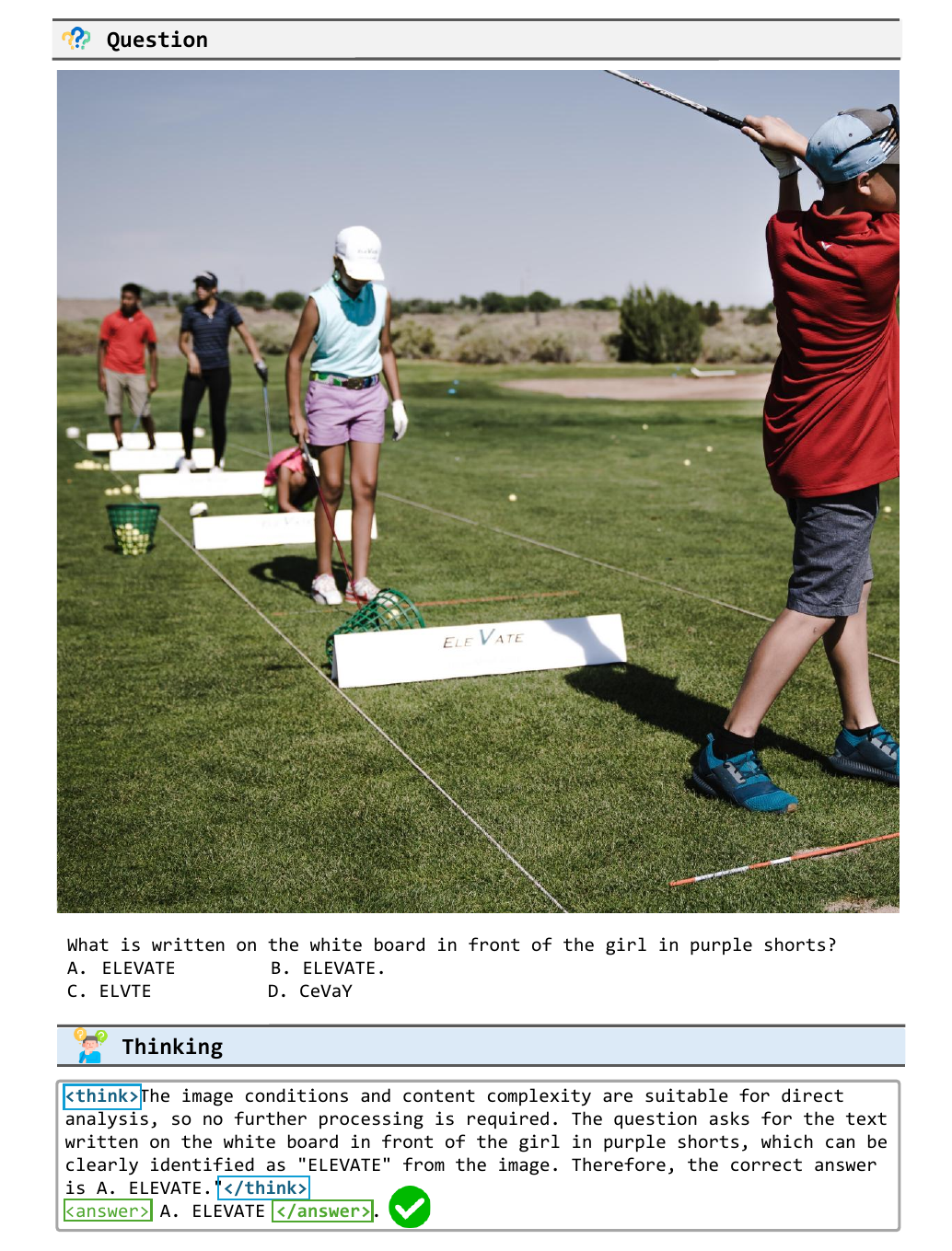}
    \caption{
        Real-world understanding case.
    }
    \label{fig:fig8_o5}
\end{figure*}

%% file: table_figs/fig8_output6.tex
\begin{figure*}[ht]
    \centering
    \includegraphics[width=1.0\linewidth]{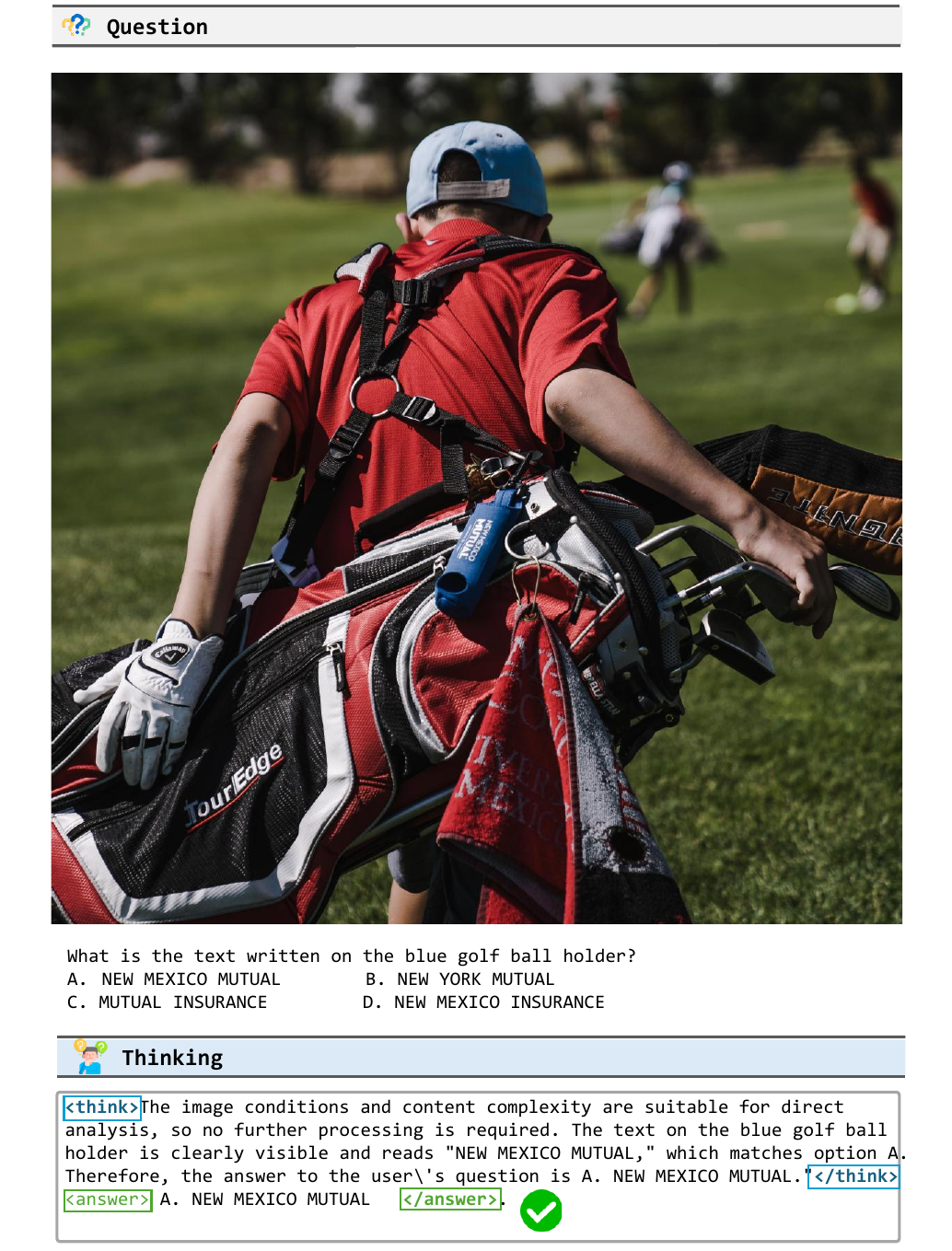}
    \caption{
        Real-world understanding case.
    }
    \label{fig:fig8_o6}
\end{figure*}

%% file: table_figs/fig8_output7.tex
\begin{figure*}[ht]
    \centering
    \includegraphics[width=1.0\linewidth]{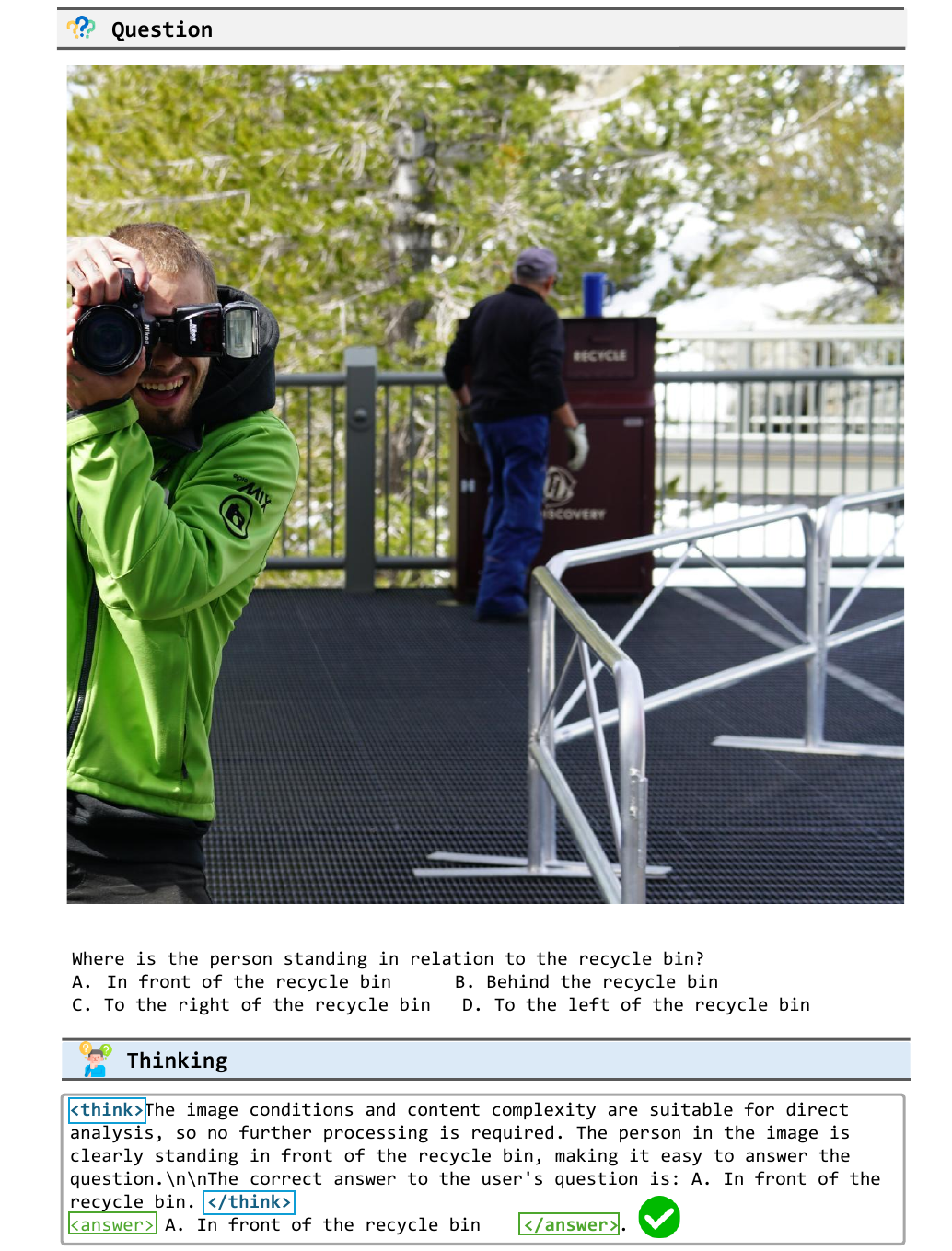}
    \caption{
        Real-world understanding case.
    }
    \label{fig:fig8_o7}
\end{figure*}

%% file: table_figs/fig8_output8.tex
\begin{figure*}[ht]
    \centering
    \includegraphics[width=1.0\linewidth]{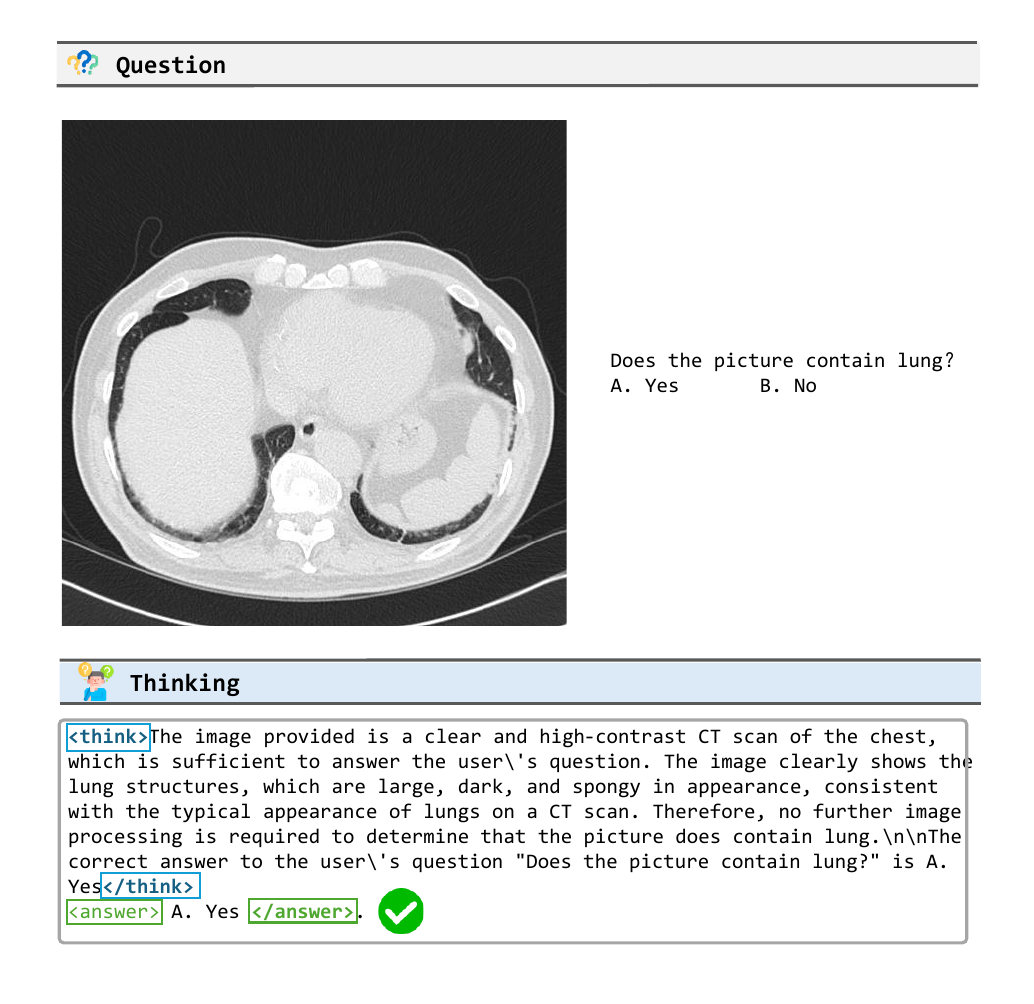}
    \caption{
        Medical case.
    }
    \label{fig:fig8_o8}
\end{figure*}

%% file: table_figs/fig8_output9.tex
\begin{figure*}[ht]
    \centering
    \includegraphics[width=1.0\linewidth]{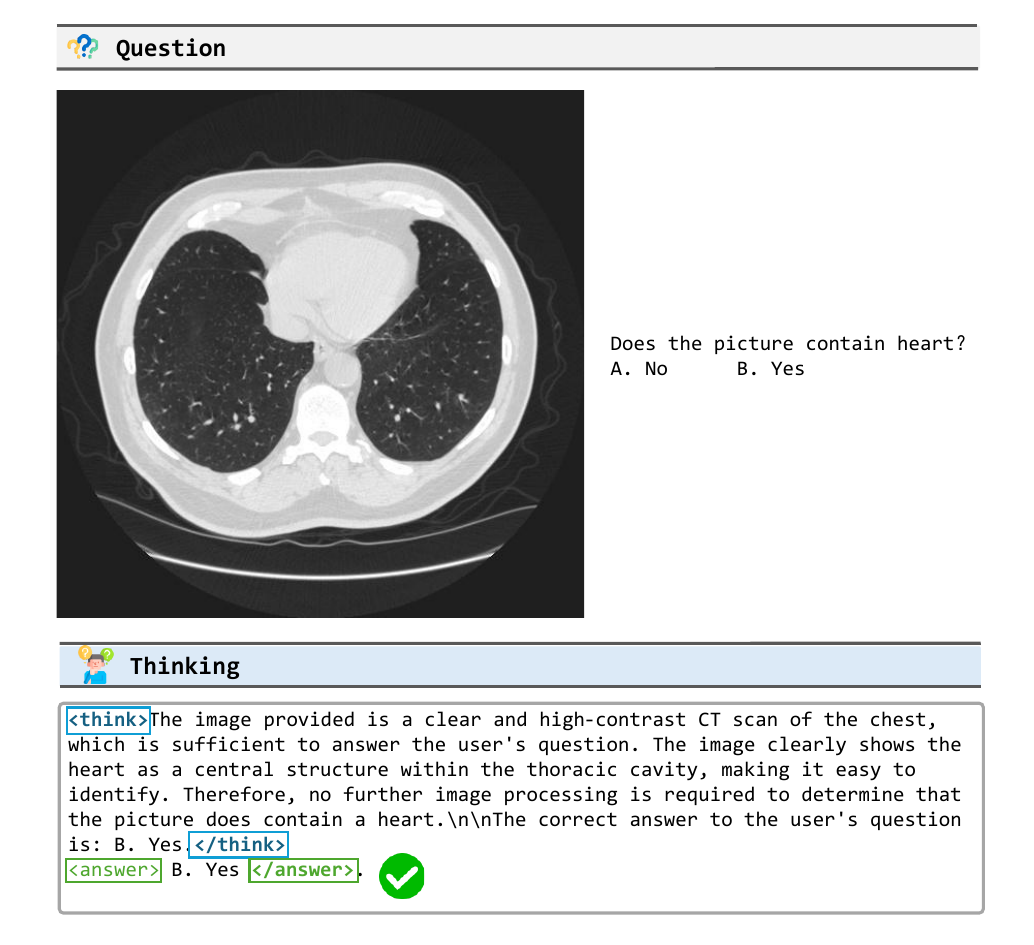}
    \caption{
        Medical case.
    }
    \label{fig:fig8_o9}
\end{figure*}

%% file: table_figs/fig8_output11.tex
\begin{figure*}[ht]
    \centering
    \includegraphics[width=1.0\linewidth]{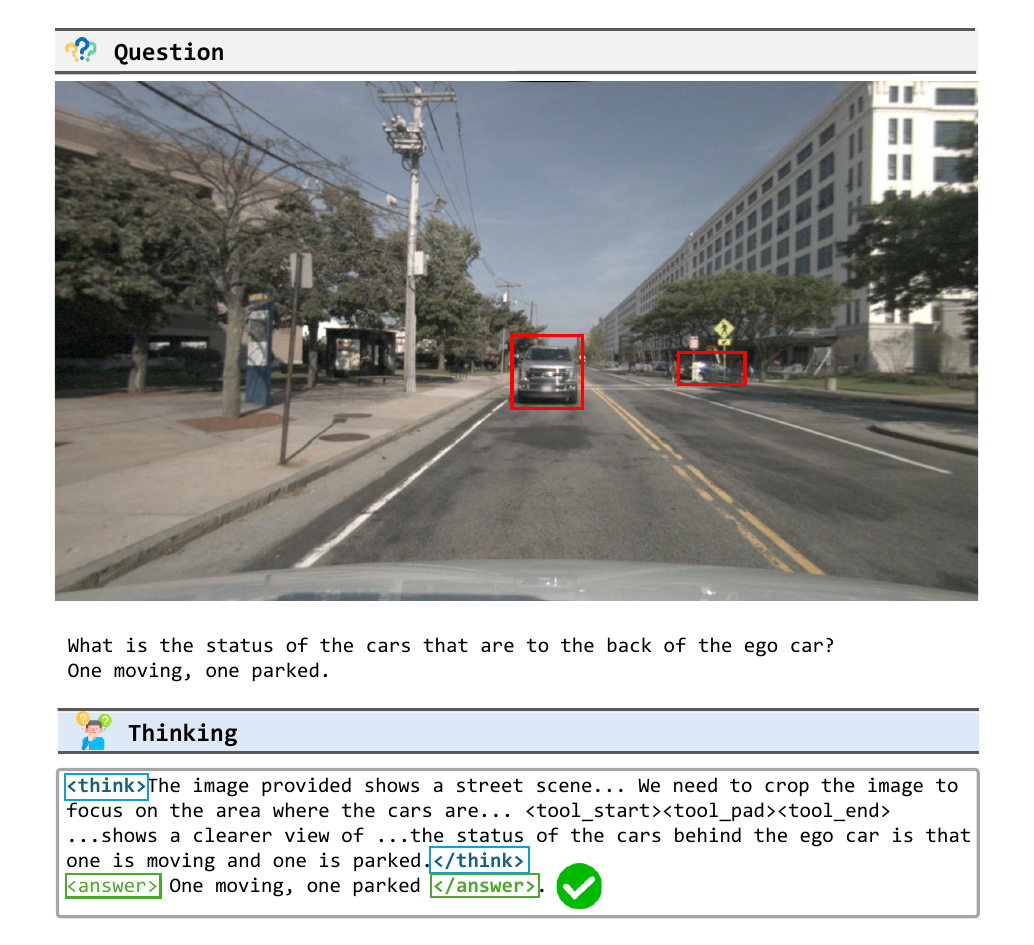}
    \caption{
        Autonomous driving case.
    }
    \label{fig:fig8_o11}
\end{figure*}

%% file: table_figs/fig8_output13.tex
\begin{figure*}[ht]
    \centering
    \includegraphics[width=1.0\linewidth]{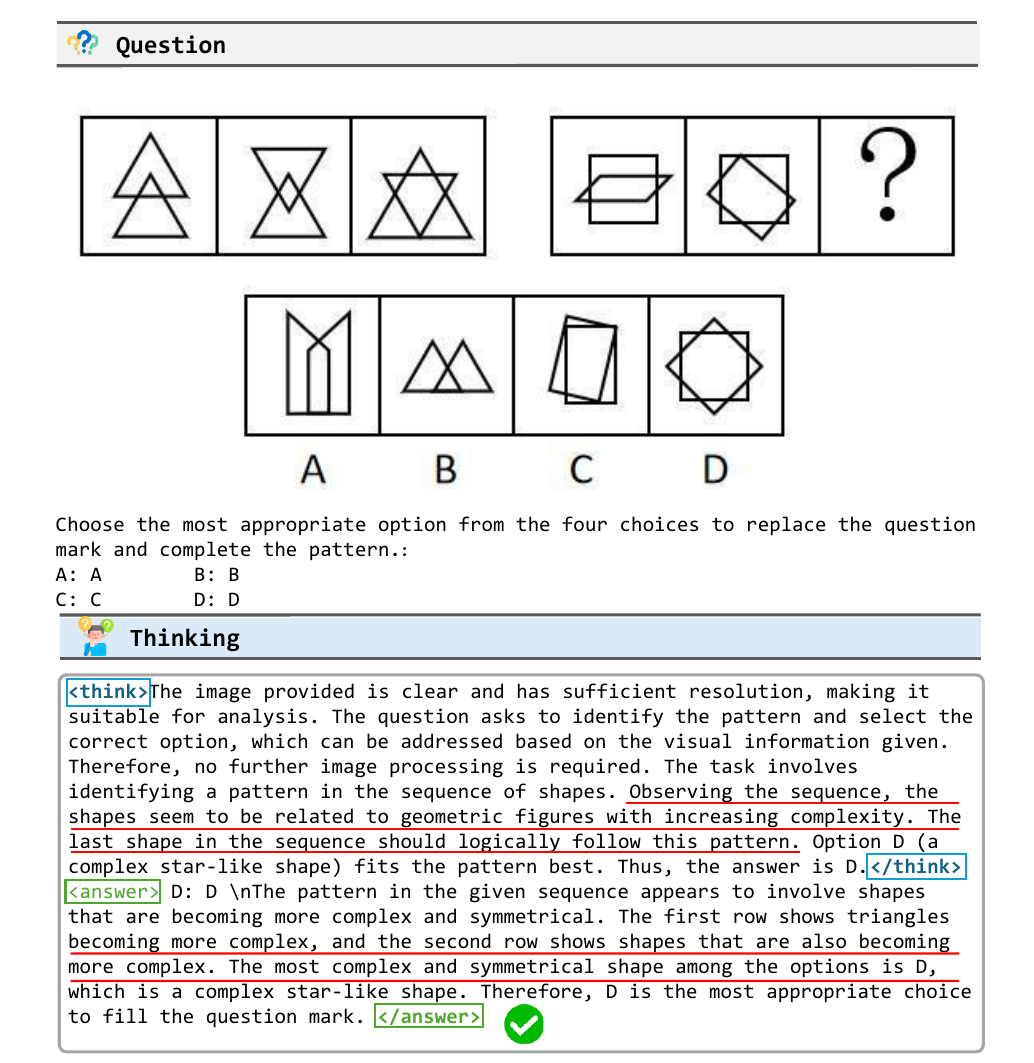}
    \caption{
        Logical reasoning case.
    }
    \label{fig:fig8_o13}
\end{figure*}

%% file: table_figs/fig8_output12.tex
\begin{figure*}[ht]
    \centering
    \includegraphics[width=1.0\linewidth]{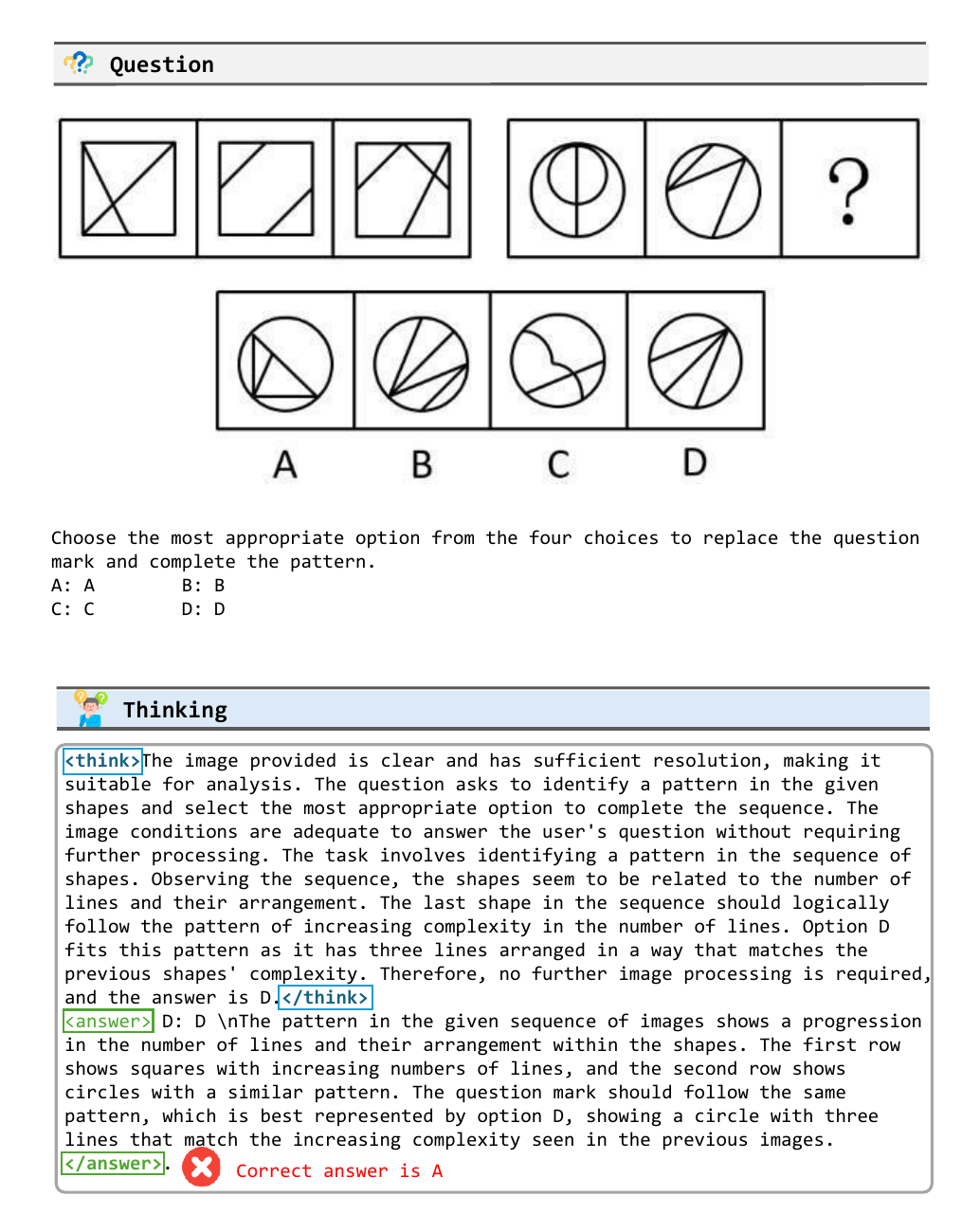}
    \caption{
        Logical reasoning case.
    }
    \label{fig:fig8_o12}
\end{figure*}

%% file: table_figs/fig8_output10.tex
\begin{figure*}[ht]
    \centering
    \includegraphics[width=1.0\linewidth]{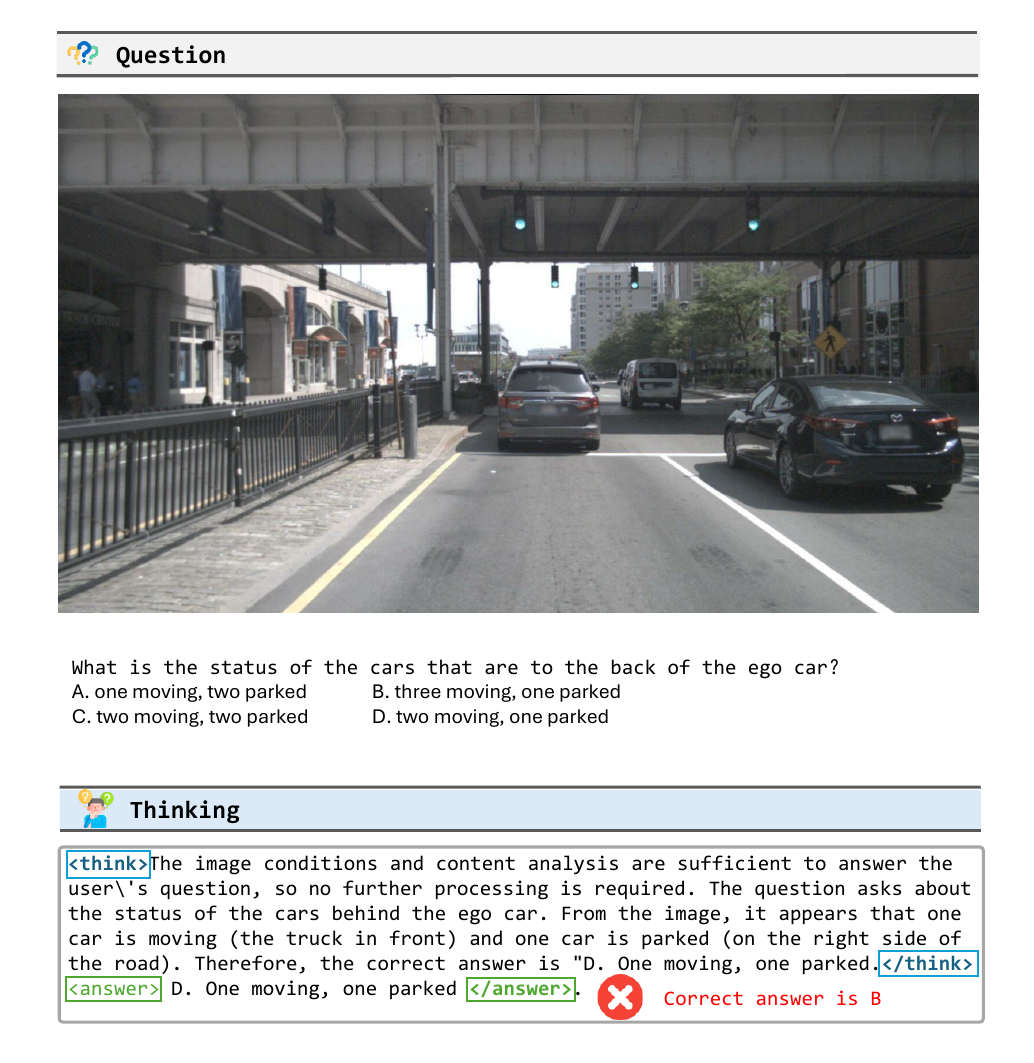}
    \caption{
        Autonomous driving case.
    }
    \label{fig:fig8_o10}
\end{figure*}